%% file: main.tex
\pdfoutput=1
\documentclass{article}
\usepackage{PRIMEarxiv} % provides a nice arXiv preprint style + headers

\usepackage[utf8]{inputenc}
\usepackage[T1]{fontenc}
\usepackage[english]{babel}

\usepackage{amsmath,amssymb,mathtools}
\usepackage{bm}
\usepackage{siunitx}
\usepackage{physics}

\usepackage{graphicx}
\usepackage{booktabs}
\usepackage{array}
\usepackage{tabularx}
\newcolumntype{L}{>{\raggedright\arraybackslash}X}

\usepackage{natbib} % keep this: formats \cite commands
\usepackage{etoolbox}
\AtBeginEnvironment{thebibliography}{\footnotesize}

\usepackage{hyperref}
\addto\extrasenglish{%
}
\usepackage[nameinlink]{cleveref}

\usepackage{microtype}
\usepackage{nicefrac}
\usepackage{float}
\usepackage{placeins}
\usepackage{comment}
\usepackage{wrapfig}
\usepackage{multicol}
\usepackage{listings}

\usepackage{xcolor}
\newif\ifdraft
\drafttrue % set \draftfalse before submission

\usepackage{algorithm}
\usepackage{algpseudocode}
\usepackage{caption}
\DeclareCaptionLabelFormat{alglabel}{Algorithm~#2}

\algdef{SE}[SUBALG]{Indent}{EndIndent}{}{\algorithmicend\ }%
\algtext*{EndIf}
\algtext*{EndFor}
\algtext*{EndLoop}
\algtext*{EndWhile}
\algtext*{Indent}
\algtext*{EndIndent}

\usepackage[toc,title,page]{appendix}

\begin{document}

\title{Spiking Neural Networks for Continuous Control via End-to-End Model-Based Learning}

\author{
  Justus Huebotter\textsuperscript{1}, 
  Pablo Lanillos\textsuperscript{1,2}, 
  Marcel van Gerven\textsuperscript{1}, 
  Serge Thill\textsuperscript{1}\\
  \textsuperscript{1} Donders Centre for Cognition, Radboud University, Nijmegen, The Netherlands\\
  \textsuperscript{2} Cajal Neuroscience Center, Spanish National Research Council, Madrid, Spain\\
  \texttt{justus.huebotter@donders.ru.nl}
}
\maketitle

\begin{abstract}
Despite recent progress in training spiking neural networks (SNNs) for classification, their application to continuous motor control remains limited. 
Here, we demonstrate that fully spiking architectures can be trained end-to-end to control robotic arms with multiple degrees of freedom in continuous environments.  
Our predictive-control framework combines Leaky Integrate-and-Fire dynamics with surrogate gradients, jointly optimizing a forward model for dynamics prediction and a policy network for goal-directed action.  
We evaluate this approach on both a planar 2D reaching task and a simulated 6-DOF Franka Emika Panda robot with torque control.  
In direct comparison to non-spiking recurrent baselines trained under the same predictive-control pipeline, the proposed SNN achieves comparable task performance while using substantially fewer parameters.
An extensive ablation study highlights the role of initialization, learnable time constants, adaptive thresholds, and latent-space compression as key contributors to stable training and effective control.  
Together, these findings establish spiking neural networks as a viable and scalable substrate for high-dimensional continuous control, while emphasizing the importance of principled architectural and training design.
%We conclude that while stable and effective control can be achieved, recurrent spiking networks remain highly sensitive to hyperparameter settings, underscoring the importance of principled design choices.  
\end{abstract}


% Main text - USE EMPTY LINES BETWEEN DIFFERENT \INPUT TO NOT BREAK FLATTENING
%\input{abstract}

\input{introduction}

\input{methods}

\input{results}

\input{conclusion}

\section*{Declaration of competing interest}
The authors declare no competing interests.

\section*{Data/code availability}
The code for this project is available under \url{https://github.com/jhuebotter/spiking_control}.

% Acknowledgements
\section*{Acknowledgments}
This work was partially supported by the Spikeference project, Human Brain Project SGA3 (Grant Agreement Nr. 945539).

% Bibliography
\bibliographystyle{apalike}  
\bibliography{references}  

% Appendix
\newpage
\appendix
\label{sec:appendix}
\setcounter{section}{0}

\include{appendix}

\end{document}

%% file: introduction.tex
\section{Introduction}
\label{sec:intro}

% Framing ANN-based control and rise of SNNs
Modern artificial intelligence has transformed robotic control into a learning problem.
Classical approaches typically rely on carefully specified dynamic models; yet real-world environments often introduce significant variability and uncertainty, limiting their generality.
As a result, many researchers now treat robot control as a learning task, frequently deploying artificial neural networks (ANNs) to learn control policies from data.
Spiking neural networks (SNNs) offer a compelling alternative to conventional ANNs, particularly in settings where temporal dynamics and energy efficiency are central, but they also introduce additional complexity in the learning process.

% SNNs work well in classification / temporal tasks, but not yet in control
Despite these challenges, spiking networks have already proven highly effective in sensory processing, classification, and unsupervised learning \citep{neftci2019surrogate, yin2021accurate, zenke2018superspike, rossbroich2022fluctuation, tavanaei2019deep}, where their sparse, event-driven nature and internal state dynamics provide concrete advantages.
These models, inspired by how biological neurons communicate, are increasingly adopted in abstract computational settings for their ability to process temporally structured input and operate with remarkable efficiency.
Yet most practical machine learning, especially in continuous control, remains dominated by deep learning architectures with continuous-valued activations.

% The existence proof from biology
From a neuroscience perspective, this is striking: biological brains achieve robust, adaptive, and low-latency motor behavior using spike-based communication. 
This offers a powerful existence proof that spike-based control can be both effective and scalable in embodied systems. 
However, translating this biological insight into practical machine learning systems for robotic control remains a significant challenge.

% The training challenge for SNNs in control
One central challenge limiting broader use of SNNs in control settings is the difficulty of training spiking networks end-to-end for continuous tasks.
Unlike standard ANNs and gated recurrent neural networks (RNNs), which benefit from smooth gradients and well-established optimization pipelines, SNNs exhibit discontinuities due to spiking non-linearities and require surrogate approximations to enable gradient-based learning.
This has traditionally pushed spiking control research toward local plasticity, online adaptation, or indirect training paradigms, rather than principled end-to-end optimization.
In the following, we review recent research on SNN control to identify the methodological gaps our approach aims to address.

\paragraph{Literature overview.}
% Surveys and context
Research on spiking control spans diverse methodologies, learning approaches, and tasks, with multiple surveys synthesizing progress to date.
While these studies have advanced understanding of adaptive neuron models, network architectures, bio-plausibility, local learning rules, and various hardware constraints, they often focus on aspects somewhat peripheral to scalable, general-purpose control.
Several overviews emphasize the growing relevance of robotics applications \citep{bing2018survey, lanillos2021active, dewolf2021spiking, oikonomou2025reinforcement, mompo2024brain}, while methodological surveys cover deep-learning-inspired SNN learning and training advances \citep{tavanaei2019deep, neftci2019surrogate, zenke2021remarkable, eshraghian2023training, oikonomou2025reinforcement, yamazaki2022spiking}.
Together, these establish SNN control as promising but technically challenging, especially in high-dimensional continuous domains.

% analytical controllers
One of the earliest approaches embeds analytical control equations directly into spiking networks.
\citet{slijkhuis2023closed} developed closed-form spike coding controllers, \citet{agliati2025spiking} introduced predictive spiking control for linear systems, and \citet{traub2021dynamic} extended this paradigm with recurrent spiking networks for dynamic action inference.
Such controllers offer elegance and interpretability, drawing inspiration from control theory, but typically rely on one-step least-squares optimization rather than learning.
As a result, they depend heavily on accurate system models and often lack robustness to unmodeled dynamics in real-world environments.

% Local learning rules
Moving beyond purely analytical formulations, a large body of work has investigated bio-inspired local plasticity.
\citet{fernandez2021biological} implemented STDP-like adaptation for robotic arm reaching, \citet{juarez2022r} applied R-STDP to arm control with changing friction, and \citet{chen2020bio} developed a bio-inspired controller for a 4-DoF arm.
Complementary to plasticity-driven controllers, compact bio-inspired neuromorphic SNN controllers have also been proposed to produce smooth reaching dynamics for multi-DoF arms \citep{polykretis2023bioinspired}.
Building on the Neural Engineering Framework (NEF), \citet{dewolf2016spiking}, \citet{iacob2020models}, and \citet{marrero2024novel} applied NEF-based models to arm control, linking adaptive computation, cognition, and robotics, with later Loihi-based extensions to 7-DoF arms \citep{dewolf2023neuromorphic}.
Beyond arm reaching, \citet{schmidgall2023synaptic} proposed synaptic motor adaptation for locomotion, while \citet{jiang2025fully} introduced a fully spiking locomotion controller.
These approaches make use of data-driven adaptation and offer some biological plausibility, but they remain limited to relatively low-complexity tasks and do not yet scale to high-DoF continuous control.

% Conversion + neuromorphic deployment
Another important direction focuses on ANN-to-SNN conversion/transfer.
Representative conversion pipelines include converting trained networks for robotic inverse kinematics and control \citep{volinski2022data}, converting RL policies for event-based obstacle avoidance \citep{salvatore2020neuro}, and converting discrete-action DQN policies to spiking Q-networks \citep{patel2019improved}.

A closely related direction is the deployment of spiking controllers on neuromorphic hardware.
As an example of directly trained spiking DRL agents targeting neuromorphic accelerators, \citet{zanatta2023directly} demonstrated spiking models optimized for neuromorphic accelerators, highlighting energy efficiency benefits.
\citet{zhao2020closed} implemented closed-loop control on the iCub robot using DYNAP-SE, \citet{paredes2024fully} realized fully neuromorphic drone flight, and \citet{dewolf2023neuromorphic} deployed Loihi-based control of a 7-DoF arm.
Deployment-focused studies show the feasibility of deploying spiking control policies on neuromorphic hardware, but they are typically constrained by hardware-specific neuron models and architectural limitations, often prioritizing efficiency over generality.

% Surrogate gradients
In contrast to hardware-driven or conversion-based methods, surrogate gradient techniques \citep{neftci2019surrogate, zenke2021remarkable} enable true end-to-end optimization in spiking domains by approximating spike derivatives, bringing deep-learning-style training into SNNs. 
Originally developed for classification, these methods have since been adapted to control, with stabilization strategies such as fluctuation-driven initialization further improving training dynamics \citep{rossbroich2022fluctuation}. 
Applied to reinforcement learning, \citet{tang2021deep} used population coding for MuJoCo tasks, \citet{chen2024noisy} proposed noisy spiking actors to enhance exploration, \citet{park2025designing} extended TD3 to 3D arm control, and \citet{oikonomou2023hybrid} applied deep RL to 6-DoF reaching. 
Beyond arms, \citet{kumar2025dsqn} developed a spiking DQN for mobile robot navigation, \citet{traub2021many} demonstrated recurrent SNNs scaling to many-joint arms, and \citet{zanatta2024exploring} explored PPO-based DRL integration in robotic tasks. 
Together, these works highlight scalability and flexibility, overcoming many limitations of earlier local or conversion approaches. 
However, most efforts focus on policy optimization alone, without predictive modeling components.

% Predictive models
A particularly promising next step is the integration of learned predictive models into the control loop, not only for state estimation, but as generative world models capable of producing synthetic experience for policy learning \citep{ha2018recurrent, hafner2025mastering, huebotter2022learning, stockl2024local}.
This paradigm, already transformative in the ANN-based reinforcement learning community, could help SNN controllers scale to high-dimensional, long-horizon tasks without prohibitively expensive data collection.
In the SNN domain, several first steps exist: \citet{taniguchi2023world} reviewed predictive coding for cognitive robotics, \citet{agliati2025spiking} provided a formal account of predictive spiking control, and \citet{zhu2024autonomous} applied SNNs to autonomous driving by combining perception and planning.
Beyond our control setting, recent neuromorphic-motivated work has shown that local prediction learning can support planning in high-dimensional domains \citep{stockl2024local}, that cognitive-map structure combined with neural sampling can enable goal-directed imagination and planning \citep{lin2025neural}, and that similar cognitive-map learners can be applied to mapless navigation under edge constraints \citep{polykretis2024mapless}.
Most closely aligned with model-based RL, \citet{capone2024towards} proposed dreaming in recurrent SNNs, though their approach remains limited to simpler settings.
Together, these works suggest that predictive modeling may help scale SNN controllers, but the joint, end-to-end optimization of predictive and control networks in fully spiking architectures for high-DoF continuous arm control remains underexplored—a gap directly addressed by the present work.

% Task overview
Grouping by task domains further illustrates this methodological progression.
For arm reaching, analytical formulations such as closed-form spike coding and predictive control have been explored \citep{slijkhuis2023closed, agliati2025spiking}, followed by local plasticity and NEF-inspired approaches \citep{chen2020bio, juarez2022r, fernandez2021biological, dewolf2016spiking, iacob2020models, marrero2024novel, volinski2022data}.
Related spiking joint-control motifs have also been explored for articulated multi-joint motion generation outside classical robotics, e.g., via evolutionarily selected spiking controllers for realistic 3D motion synthesis \citep{polykretis2023interactive}.
Neuromorphic deployments have extended these ideas to real-world settings, with Loihi-based 7-DoF arms \citep{dewolf2023neuromorphic} and iCub controllers implemented on DYNAP-SE \citep{zhao2020closed}.
More recently, surrogate-gradient-driven RL has scaled SNN controllers to high-DoF arms, including 3D reaching with TD3 \citep{park2025designing}, 6-DoF hybrid RL control \citep{oikonomou2023hybrid}, and recurrent architectures spanning many-joint systems \citep{traub2021many}.
Beyond arms, locomotion has been approached with adaptive rules \citep{schmidgall2023synaptic, jiang2025fully}, drones with neuromorphic controllers \citep{paredes2024fully, salvatore2020neuro}, mobile robot navigation \citep{polykretis2024mapless}, and autonomous driving with predictive frameworks \citep{zhu2024autonomous}.
Broader contributions include surveys synthesizing the field \citep{bing2018survey, lanillos2021active, oikonomou2025reinforcement}, methodological advances in training \citep{tavanaei2019deep, eshraghian2023training}, and stabilization techniques for scalable optimization \citep{rossbroich2022fluctuation}.
Finally, predictive-model-based approaches remain in their infancy, with only initial steps toward integrating generative world models into spiking control \citep{capone2024towards, taniguchi2023world}.

% Gap and transition to this work
Taken together, spiking control research now spans arms, locomotion, drones, and autonomous driving, utilizing local rules, analytical models, RL, conversion, and predictive approaches.
Yet no systematic evaluation exists of how deep learning training tools can be effectively transferred to fully spiking architectures for generalizable prediction and control, and only a handful of works demonstrate end-to-end training that combines predictive modeling with high-DoF continuous arm control—the specific gap addressed by the present work.

% This work—goals, architecture, novelty
In this work, we systematically investigate how techniques from the deep learning paradigm can be adapted to train SNNs for continuous control.
To this end, we introduce the predictive-control (Pred-Control) SNN model, a fully spiking, model-based control architecture composed of two trainable spiking networks (see \autoref{fig:schematic}):
a forward model that predicts future states given current sensory input and motor commands, and a policy network that infers actions based on current and target states.
This structure mirrors classical forward/inverse models from control theory while enabling end-to-end optimization using surrogate gradients.
Rather than relying on complex reinforcement learning pipelines or neuromorphic deployment constraints, we adopt a simpler yet sufficiently rich experimental setting: a goal-directed reaching task with a simulated robotic arm in both 2D and 3D.

Using this framework, we show that many of the techniques that enable effective ANN training can be transferred to spiking networks, provided care is taken to stabilize dynamics and ensure smooth optimization.
We demonstrate that fully spiking networks, trained end-to-end with surrogate gradients, can produce accurate, low-latency torque control for both planar and 6-DOF robotic manipulators.
Through extensive ablation studies, we identify which architectural and algorithmic essential components and optional features enable stable learning and high task performance, and which offer diminishing returns.
Our final models retain the adaptability and expressiveness of deep architectures while embracing the temporal and event-driven nature of spiking computation.
These results support the view that SNNs, when trained with principled methods from deep learning, can function as scalable, adaptable control systems.
Together, these findings offer a clear path for constructing scalable, adaptable, and biologically inspired control systems that bridge the gap between deep learning, SNNs, and low-level continuous control (motor-learning / robotics).

% Paper structure
The remainder of this paper is structured as follows.
\autoref{sec:methods} introduces the Pred-Control SNN architecture in detail, including %the employed LIF dynamics and 
the set of training strategies evaluated for spiking networks, as well as a description of the experimental setup.
\autoref{sec:results} presents the main results, with a focus on continuous control of a 3D robotic arm.
\autoref{sec:conclusion} then discusses the implications and limitations of our findings and outlines promising directions for future research.
An extensive \hyperref[sec:appendix]{Appendix} complements the main text by detailing additional experiments (e.g., parameter analysis, surrogate functions comparison, initialization, training details, regularizations, etc), which were critical in shaping our methodological insights but are too detailed for inclusion in the main narrative.

%% file: methods.tex
\section{Methods}
\label{sec:methods}

\begin{figure}[h!]
    \centering
    \includegraphics[width=0.9\textwidth]{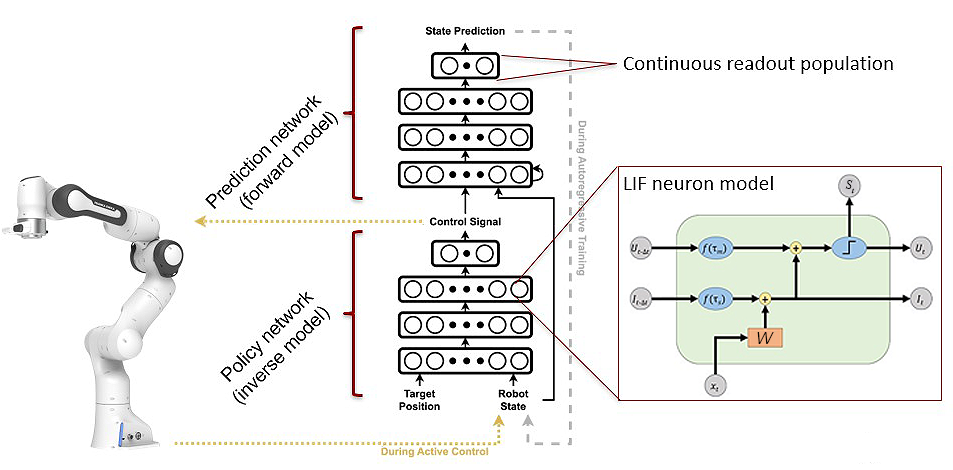}
    \caption{\textbf{Pred-Control SNN architecture.}
    The system consists of two spiking neural networks composed of LIF neurons with learnable parameters $\bm{\theta}$:
    a \emph{prediction network} $\bm{\upsilon}$ (forward model), which receives the current robot state $\bm{s}_t$ and control signal $\bm{u}_t$ to predict the state change $\Delta \hat{\bm{s}}_t$, and a \emph{policy network} $\bm{\pi}$ (inverse model), which takes in the current state $\bm{s}_t$ and target state $\bm{s}^*_t$ to compute a control output $\bm{u}_t$.
    During active control, only the policy network is used; during training, the prediction network is rolled out autoregressively to provide differentiable state estimates for optimizing the policy.
    Each network ends in a continuous readout layer decoding membrane voltages into output vectors.
    A schematic of the LIF neuron model is shown on the right.
    }
    \label{fig:schematic}
\end{figure}

\begin{table}[h!]
\centering
\caption{\textbf{Notation summary} for task and model variables used in the Pred-Control SNN.}
\label{tab:notation}
\footnotesize
\begin{tabularx}{\linewidth}{@{} l L @{}} \toprule
\textbf{Symbol} & \textbf{Description} \\
\midrule
$\bm{s}_t$               & State of the system at time $t$ (e.g., joint angles, end-effector pose). \\
$\bm{s}^*_t$             & Target state at time $t$ (e.g., goal position of end-effector). \\
$\hat{\bm{s}}_t$         & Predicted system state at time $t$. \\
$\Delta \hat{\bm{s}}_t$  & Predicted change in state at time $t$. \\
$\bm{u}_t$               & Control input applied to the system (e.g., torque, velocity). \\
$\bm{x}_t$               & Input vector to the network at time $t$: 
                           $\bm{x}_t = [\bm{s}_t,\bm{u}_t]$ for the prediction network, and 
                           $\bm{x}_t = [\bm{s}_t,\bm{s}^*_t]$ for the policy network. \\
$\bm{y}_t$               & Network output (predicted change or control signal). \\
$ I_{\text{inj},t}$      & Injected current into the first spiking layer at time $t$. \\
$e_{\hat{s}}$            & Prediction error. \\
$e_{\bm{\pi}}$           & Policy error over a trajectory. \\
$\bm{\theta}_{\upsilon}$ & Learnable parameters of the prediction network. \\
$\bm{\theta}_{\pi}$      & Learnable parameters of the policy network. \\
\bottomrule
\end{tabularx}
\end{table}

\subsection{Pred-Control SNN}
Our SNN architecture is depicted in \autoref{fig:schematic}. It draws inspiration from world models~\cite{ha2018recurrent} and predictive coding approaches to robot control~\cite{lanillos2021active,taniguchi2023world} and is composed of: $i)$ a \emph{prediction} network $\bm{\upsilon}$, which learns to predict the robot’s future state evolution as a function of its current state and actions; and $ii)$ a \emph{policy} network $\bm{\pi}$, which computes the best action (continuous control signals) to apply depending on the state of the system and a given end-effector target position.
Breaking up the network into two parts allows us to use the prediction model $\bm{\upsilon}$ to simulate state trajectories during learning of the policy model $\bm{\pi}$.
We can directly propagate the gradients of the distance between the predicted future positions and the target (policy loss) with respect to the policy network parameters through the prediction network.

All neurons inside the networks are modeled as leaky integrate-and-fire (LIF) units with optional adaptive thresholds (ALIF). 
These neuron models introduce several time constant parameters $\tau$, which require careful tuning, as they control the temporal dynamics of information integration and eligibility traces for surrogate-gradient-based backpropagation.
Details of the LIF and ALIF implementations, including time constants and reset mechanisms, are described in detail in \autoref{sec:lif} and  \autoref{sec:adaptive_lif}.
For each environment time step, the SNNs are running for a set of internal sub-steps (e.g., 7) to propagate input information through the network with spike signals for sufficient temporal resolution.
All networks and learning dynamics have been implemented using our Control Stork framework\footnote{\url{https://github.com/jhuebotter/control_stork}} (based on PyTorch) and the specific experiments for this project are available on GitHub\footnote{\url{https://github.com/jhuebotter/spiking_control}}.
All relevant hyperparameters for the final model are summarized in \autoref{sec:hyperparameters}.
A brief overview of the notation for the following sections is given in \autoref{tab:notation}.
%We develop two SNNs for our experiments: a \emph{prediction network} that forecasts future states of the robotic arm, and a \emph{policy network} that produces continuous control signals to track a target in either 2D or 3D.

\paragraph{Prediction Network}
The prediction network $\bm{\upsilon}$ comprises two spiking LIF populations and one non-spiking leaky integrator readout population.
The first spiking layer includes full recurrent connectivity.
Network size and architectural variants were explored empirically (see \autoref{sec:network_architecture}).
Its input consists of the current robot state $\bm{s}_t$ (joint angles, end-effector pose) and control action $\bm{u}_t$ (joint velocity or acceleration).
The network outputs a predicted state change $\Delta \hat{\bm{s}}_t$, which is added to the previous estimate $\hat{\bm{s}}_t$ to yield $\hat{\bm{s}}_{t+1} = \hat{\bm{s}}_t + \Delta \hat{\bm{s}}_t$.
The prediction error $e_{\hat{s}}$ at time $t$ is measured by
\begin{equation}
\label{eq:pred_error}
e_{\hat{s}}(t) = \frac{1}{K}\sum_{i=1}^{K}(s_i(t) - \hat{s}_i(t))^2,
\end{equation}
where $K$ denotes the dimensionality of the full state vector $\bm{s}_t$. During learning this error further is averaged over a trajectory of length $T$.
All learnable parameters are summarized as $\bm{\theta_\upsilon}$, yielding the compact notation $\bm{\upsilon}(\hat{\bm{s}}_{t+1}|\bm{s}_t,\bm{u}_t,\bm{\theta_\upsilon})$.
For implementation and training details, refer to \autoref{sec:prediction_training}.

\paragraph{Policy Network}
The policy network $\bm{\pi}$ shares the architecture of the prediction model, except that it omits recurrent connections (see \autoref{sec:network_architecture} for details).
Its inputs are the current state $\bm{s}_t$ and target position $\bm{s}^*_t$, and it outputs an action vector $\bm{u}_t$.
This vector can represent joint velocity, acceleration, or torque, depending on the task.
Since it is not immediately obvious whether $\bm{u}_t$ leads to the target, the output is evaluated by simulating the trajectory using the prediction model to obtain a new predicted position.
We define the policy loss as
\begin{equation}
\label{eq:pol_error}
e_{\bm{\pi}} = \frac{1}{KT} \sum_{t=1}^T\sum_{i=1}^{K^*}(s_i^*(t) - \hat{s}_i(t))^2,
\end{equation}
where $K^*$ refers to the subset of state dimensions describing the task-relevant target, specifically the Cartesian coordinates of the end-effector position.
All learnable parameters are denoted by $\bm{\theta_\pi}$, and we write the policy model as $\bm{\pi}(\bm{u}_t|\bm{s}_t,\bm{s}_t^*,\bm{\theta_\pi})$.
See \autoref{sec:policy_training} for details.

\paragraph{Encoding and Decoding}
All state and action variables are linearly rescaled to $[-1,1]$ and encoded via a learned linear projection.
Specifically, the scalar input vector $\bm{x}_t$ ($[\bm{s}_t, \bm{u}_t]$ for prediction network, $[\bm{s}_t, \bm{s}^*_t]$ for policy network) is linearly transformed by
\begin{equation}
I_{\text{inj},t} = W_{\text{in}}^\intercal\bm{x}_t + \bm{b}_{\text{in}},
\end{equation}
where $W_{\text{in}}$ and $\bm{b}_{\text{in}}$ are learned parameters. $I_{\text{inj},t}$ serves as the input current to the first spiking layer.
Injected currents are not constrained to be positive; instead, both the sign and magnitude of $I_{\text{inj},t}$ are determined by the learned input projection, allowing negative-valued inputs to be represented naturally through negative currents.
Alternative encoding schemes were explored but did not yield consistent improvements and are therefore omitted from the final model.

To obtain the continuous-valued output vector $\bm{y}_t$ (actions or predictions), we use the membrane potentials $U_t$ of a non-spiking layer.
These first and final weight matrices are therefore scaled differently compared to other weight parameters in the network to generate stable internal dynamics as well as output on the correct expected magnitude.
The output scaling also differs for prediction and control models, as the predicted changes in state $\Delta \hat{\bm{s}}(t)$ are typically much smaller than the control signals $\bm{u}(t)$.
In the policy network, a $\tanh$ activation function ensures that control outputs remain within $[-1, 1]$.
To mitigate saturation-induced gradient collapse, we apply an auxiliary $\ell_2$ penalty to the policy’s pre-$\tanh$ activations when they exceed the range $[-3,3]$.

\subsection{Leaky integrate-and-fire neuron model}
\label{sec:lif}

Leaky integrate-and-fire models form a broad class of neuron models that vary in complexity, parameterization, and numerical methods \citep{gerstner2014neuronal}. 
Choosing the LIF neuron model for robotics and machine learning tasks is based on a suitable trade-off between computational efficiency and biological plausibility because its simple, interpretable dynamics enable fast simulation and stable gradient-based learning, while still capturing key temporal integration properties necessary for control and sequence-based tasks. 
Variants of the LIF model differ in whether they include explicit synaptic currents, refractory periods, or adaptive thresholds; here, we adopt a current-based leaky integrate-and-fire (LIF) neuron with explicit synaptic current dynamics with subtractive reset and no refractory period, as this provides both computational efficiency and sufficiently rich temporal traces to support surrogate gradient learning.
While a single-variable LIF model may suffice for this task, we adopt a current-based LIF with synaptic dynamics to remain compatible with the fluctuation-driven initialization of \citet{rossbroich2022fluctuation}, avoiding re-derivation for a different neuron model.

The temporal dynamics of our non-spiking LIF model are described by two differential equations for the membrane potential $U(t)$ and synaptic current $I(t)$:

\begin{align}\label{eq:mem}
    \tau_{\text{mem}}\,\dv{U(t)}{t} &= -\bigl(U(t) - U_{\text{rest}}\bigr) + R\,I(t)\\
\label{eq:syn}
    \tau_{\text{syn}}\,\dv{I(t)}{t} &= -I(t) + W\sum_{f}\delta\bigl(t - t_f\bigr) + I_{\text{inj}}(t)
\end{align}
where $R$ is the membrane resistance, $U_{\text{rest}}$ is the resting potential, and $t_f$ are presynaptic spike times which contribute jumps of magnitude $W$.
The optional term $I_{inj}(t)$ can add a bias or continuous input, while $\tau_{\text{mem}}$ and $\tau_{\text{syn}}$ set the timescales of membrane and synaptic decay.

The LIF neurons compute firing events $S(t)$ using a Heaviside step function $\Theta(U)$, defined by:
\begin{equation}\label{eq:fire}
\Theta(U) = \begin{cases}
    1 & \text{ if }U(t) \ge \vartheta,\\
    0 & \text{otherwise}.
\end{cases}
\end{equation}
When $U(t)$ exceeds the threshold $\vartheta$, the neuron spikes and its membrane potential is reset. 

All experiments in this work are performed in discrete time using Euler updates. 
Let $\Delta t$ be the simulation step, and define:
\begin{equation}\label{eq:beta_defs}
\beta_{\text{mem}} = \exp\!\left(-\tfrac{\Delta t}{\tau_{\text{mem}}}\right),
\quad
\beta_{\text{syn}} = \exp\!\left(-\tfrac{\Delta t}{\tau_{\text{syn}}}\right).
\end{equation}

We then express the Euler updates for $I(t)$ and the pre-reset membrane potential $\tilde{U}_t$ as:
\begin{equation}\label{eq:lif_i}
I_t = \beta_{\text{syn}}\,I_{t-1} + W\,x_t + I_{inj,t},
\end{equation}
\begin{equation}\label{eq:lif_u}
\tilde{U}_t = \beta_{\text{mem}}\,U_{t-1} + \bigl(1-\beta_{\text{mem}}\bigr)\,I_t.
\end{equation}
Here, $\tilde{U}_t$ is a temporary variable used to determine spiking before reset and is not an independent state variable.

After computing $\tilde{U}_t$, we check for a spike using $S_t = \Theta\bigl(\tilde{U}_t\bigr)$. 
If the neuron spikes ($S_t=1$), the membrane potential is updated using a subtractive reset:
\begin{equation}\label{eq:subreset_new}
U_t = \tilde{U}_t - \vartheta\,S_t.
\end{equation}

We set $\vartheta=1$ and $U_{\text{rest}}=0$ for simplicity. 
Note that if nonzero resting potentials were desired, the reset formulation would have to be adapted accordingly.

\begin{figure}[h!]
    \centering
    \includegraphics[width=0.38 \linewidth]{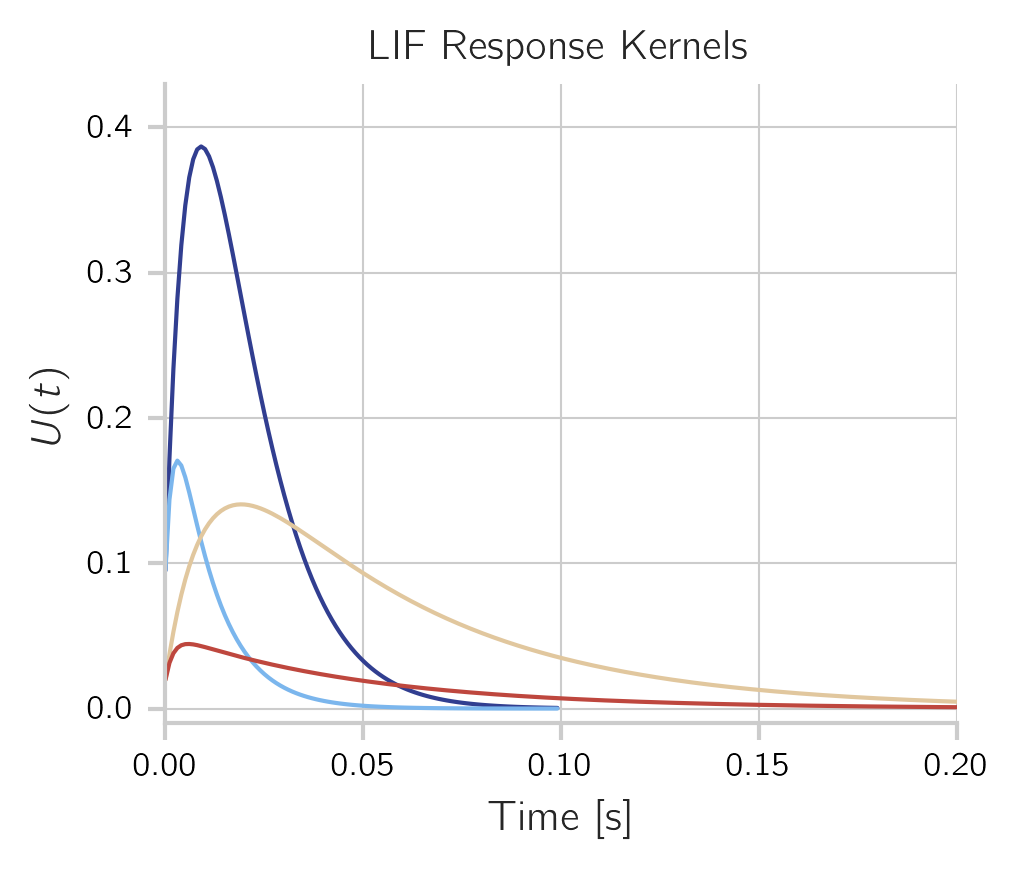}
    \includegraphics[width=0.6\linewidth]{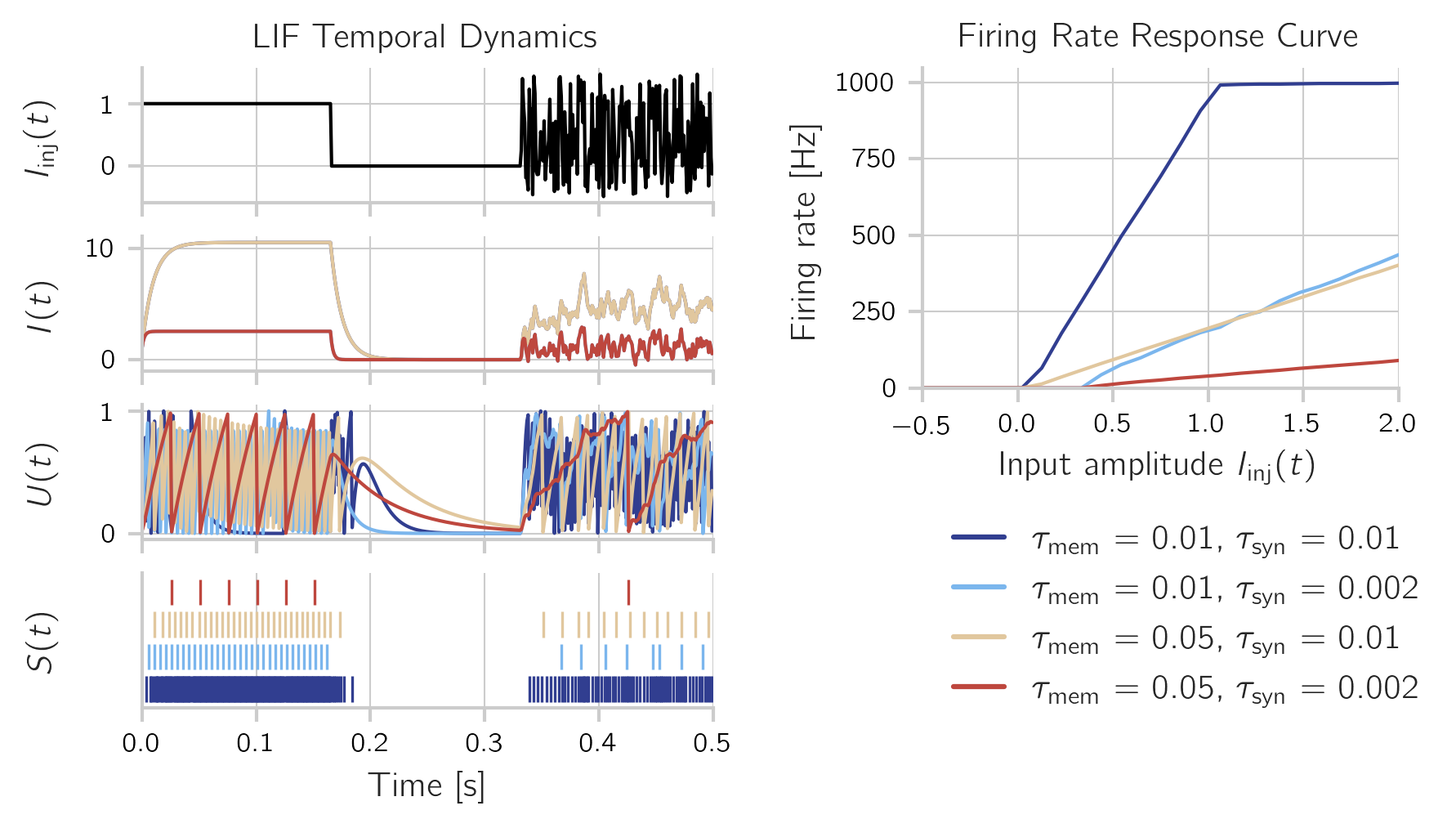}
    \caption{\textbf{Dynamics of the LIF neuron model and the influence of temporal parameters.}
    \textbf{Left:} The membrane voltage response $U(t)$ to a single input spike varies in amplitude and duration depending on the membrane and synaptic time constants $\tau_{\text{mem}}$ and $\tau_{\text{syn}}$. 
    \textbf{Center:} Full LIF temporal dynamics under three regimes of injected current $I_{\text{inj}}(t)$: constant input, silence, and high-frequency noise.
    Traces show filtered current $I(t)$, membrane potential $U(t)$, and spike activity $S(t)$.
    \textbf{Right:} Firing rate response curves over constant current injection amplitudes.
    The orange and light blue traces, corresponding to larger and smaller time constants respectively, show overall similar response curves but with distinct differences.
    While the light blue (fast) neuron requires higher input to spike, it responds more rapidly once active; the slower orange neuron integrates more gradually but yields longer activity traces.
    This tradeoff between response speed, firing magnitude, and trace duration critically affects not only neuronal responsiveness but also the effective temporal horizon over which surrogate gradients can propagate.
    Balancing these dynamics is a central challenge in the design and training of spiking neural networks.
    }
    \label{fig:lif_kernels}
\end{figure}

The complete LIF model dynamics are summarized in \autoref{alg:euler} and visualized in \autoref{fig:lif_kernels} for different settings of time constants $\tau_{\text{mem}}$ and $\tau_{\text{syn}}$.

\begin{algorithm}[h!]
\caption{\textbf{Discrete-time update procedure for leaky integrate-and-fire neurons.}
Each population updates its synaptic current $I(t)$ and membrane voltage $U(t)$ according to exponential decay dynamics, receiving weighted spike inputs $x(t)$ and external current $I_{\text{inj}}(t)$.
Spikes $S(t)$ are generated when the voltage exceeds the threshold $\vartheta$, after which a subtractive reset is applied.
All updates are differentiable via surrogate gradients during training.
}
\label{alg:euler}
\begin{algorithmic}
\For{all populations $p$ in network}
    \State $U_0 \gets U_{\text{rest}},\; I_0 \gets 0,\; S_0 \gets 0$
    \Comment{Initialize population states}
\EndFor
\For{$t \in [1 \dots T]$}
    \For{all populations $p$ in network}
        \State $x_t \gets [S_{t-1,i}\;\text{for all populations } i \text{ projecting to } p]$
        \Comment{Gather spike outputs from the previous step}
        \State $I_t \gets \beta_{\text{syn}}\,I_{t-1} + W\,x_t + I_{inj,t}$
        \Comment{Synaptic current update (Eq.\,\ref{eq:lif_i})}
        \State $\tilde{U}_t \gets \beta_{\text{mem}}\,U_{t-1} + \bigl(1-\beta_{\text{mem}}\bigr)\,I_t$
        \Comment{Membrane integration (Eq.\,\ref{eq:lif_u})}
        \State $S_t \gets \Theta\bigl(\tilde{U}_t\bigr)$
        \Comment{Check if neuron fires (Eq.\,\ref{eq:fire})}
        \State $U_t \gets \tilde{U}_t - \vartheta\,S_t$
        \Comment{Subtractive reset (Eq.\,\ref{eq:subreset_new})}
    \EndFor
\EndFor
\end{algorithmic}
\end{algorithm}

Having declared the LIF dynamics for our model, there are several other choices which can be made for model training, some of which are mandatory for surrogate gradient learning, while others remain optional.

\subsection{Mandatory Training Considerations}
\label{sec:necessary_steps}

\paragraph{Surrogate Gradients}
Non-differentiable spike functions are approximated with smooth surrogate gradient methods during the backward pass.
We evaluated several commonly used surrogate functions (sigmoid, Gaussian Spike~\cite{yin2021accurate}, and Super Spike~\cite{zenke2018superspike}) and investigated the impact of steepness $\beta$ of the function on gradient magnitude and learning stability (see \autoref{sec:surrogate_gradients}).

\paragraph{Parameter Initialization}
SNNs (similar to classic recurrent neural networks) are sensitive to weight initialization, as only certain parameter regimes yield stable behavior and effective learning.
We employ a fluctuation-driven initialization \citep{rossbroich2022fluctuation}, which incorporates LIF time constants $\tau_{\text{mem}}$ and $\tau_{\text{syn}}$ into the scaling of synaptic weights, improving upon common ANN-based initialization methods.
The original scheme was developed for SNNs in (temporal) classification tasks, which allows for a priori assessment of input firing rates to identify an appropriate weight scale factor $\nu$.
In our case, we estimate $\nu$ empirically as the training data is not known at time of network initialization and the network input is continuous instead of precomputed spike patterns.
Our networks require careful tuning of parameters to achieve stable membrane potentials and gradients throughout the networks (see \autoref{sec:empirical_init}).
%We fix $\sigma_U=1$ and $\mu_U=0$ throughout.

\paragraph{Training Loops}
The prediction and policy networks are trained iteratively using batches from a replay buffer.
We adopt unroll-based training (truncated backpropagation through time) with $w = 10$ environment warmup steps and teacher forcing for the transition model.
The policy model is updated via autoregressive predictions through the frozen transition model
Unless stated otherwise, networks are trained using mini-batches of 256 samples for 25 batches per iteration.
See \autoref{sec:training_loop}, \ref{sec:prediction_training}, and \ref{sec:policy_training}.

\paragraph{Parameter Optimization}
The gradient magnitude in our networks is influenced by several factors. 
The LIF time constants $\tau_{\text{mem}}$ and $\tau_{\text{syn}}$ (as well as $\tau_{\text{ada}}$ for ALIF neurons) control the length of eligibility traces any given input spike event leaves on the membrane voltage of downstream neurons. 
Further, the number of simulation steps $r$, the number of SNN sub-steps per simulation step, and the total number of spike events $S$ collectively affect the risk of exploding or vanishing gradients in partially recurrent networks.
Finally, the choice of surrogate gradient function and the steepness parameter $\beta$ scale the gradients directly. 
Jointly searching the parameter space quickly becomes infeasible due to the large number of degrees of freedom in this optimization problem. 
We found that using the adam optimizer~\cite{kingma2014adam}, which can dynamically adapt learning rates per parameter, proved effective in stabilizing training.
As the gradient magnitudes for prediction and policy networks vary, we evaluated a number of different learning rates for each model (see \autoref{sec:learning_rate}).
The final learning rates used are $\alpha_{\tau} = 0.01$ for time constants and $\alpha = 0.001$ for all other parameters across both networks without decay.

\subsection{Optional Training Components}
\label{sec:optional_methods}

% [Optional components overview]
Several optional architectural and training components were evaluated to improve stability, temporal credit assignment, and parameter efficiency in spiking predictive control.
Learnable membrane, synaptic, and adaptation time constants allow neurons to tune their intrinsic temporal dynamics and substantially improve convergence and final performance (\autoref{sec:learnable_timescales}).
Adaptive LIF (ALIF) neurons introduce activity-dependent thresholds that stabilize firing rates, re-engage silent neurons, and improve long-horizon credit assignment (\autoref{sec:adaptive_lif}).
Latent-space compression constrains the dimensionality of inter-layer representations, enabling wider spiking layers at fixed parameter budgets and yielding favorable performance–efficiency trade-offs (\autoref{sec:reducing_parameters}).
Additional components, including scheduled learning rates (\autoref{sec:learning_rate}), weight decay (\autoref{sec:weight_reg}), activity regularization (\autoref{sec:activity_reg}), action penalties (\autoref{sec:action_reg_loss}), and exploratory action noise (\autoref{sec:action_reg_noise}), were explored but did not consistently improve final performance beyond the components above and are therefore omitted from the final model.

\begin{comment}
\paragraph{Scheduled Learning Rates}
Instead of using a constant learning rate, we explored an exponential decay schedule (see \autoref{sec:learning_rate}).

\paragraph{Learnable Time Constants}
LIF time constants $\tau_{\text{mem}}$ and $\tau_{\text{syn}}$ (as well as $\tau_{\text{ada}}$ for ALIF neurons) can be learned for each neuron individually.
This allows the network to tune its temporal processing dynamics.
For a detailed overview, see \autoref{sec:learnable_timescales}.

\paragraph{Regularization}
In SNNs regularization may be critical to reduce the bursting and dying networks effect, and also to improve communication efficiency \cite{hubotter2021training}. 
We evaluated multiple regularization techniques to stabilize training and improve generalization.
These include $L_2$ weight decay (\autoref{sec:weight_reg}), network activity constraints (\autoref{sec:activity_reg} and \ref{sec:adaptive_lif}), as well as action penalties (\autoref{sec:action_reg_loss}) and random action space exploration (\autoref{sec:action_reg_noise}).

\paragraph{Reducing Network Parameters}
Full rank connectivity between layers can quickly lead to a large number of parameters to optimize. 
We investigated a method to restrict the latent dimension of the encoded information, which can reduce the number of parameters needed in the networks by several orders of magnitude (see \autoref{sec:reducing_parameters}).
%This over-parametrization is often considered as a key factor for the effective function approximation of neural networks in arbitrary task domains. 
\end{comment}

\subsection{Experimental setup}
We evaluated Pred-Control SNNs on two simulated continuous control environments, a 2D planar arm and a 6-DOF robotic arm. Both tasks are goal-reaching problems trained end-to-end using identical data collection, unrolling, and optimization pipelines. Performance is assessed using cumulative distance to target, success rate, time on target, and prediction mean squared error (Prediction MSE).
Prediction MSE is computed from autoregressive rollouts of the learned forward model and reflects long-horizon prediction accuracy rather than one-step prediction loss (see \autoref{tab:performance_metrics}).

\paragraph{2D Planar Arm}
The 2D task simulates a two-joint planar arm with fixed segment lengths in PyGame.
The state includes joint angles and end-effector position, while actions specify joint accelerations.
Observations are partially observable (no velocity information in state) and include mild noise, enabling fast experimentation and hyperparameter screening. 
%The arm segment lengths are given as 0.5 and 0.4 arbitrary units (au) and a reaching attempt qualifies as successful when the end effector arrives within 0.05 au of the target.
%This task enables fast experimentation and analysis due to its short rollout time and low computational requirements.
The unrestricted robot joint angles are given as both $\sin$ and $\cos$ of their angle, as well as the $x$ and $y$ coordinates of the end effector and the target position.
Both the state and action representations are scaled such that they lie within $[-1, 1]$.

\paragraph{3D Robot Arm (6-DOF)}
The 3D task uses a simulated Franka Emika Panda robot with six actuated degrees of freedom and torque control, implemented in NVIDIA Isaac Sim / Isaac Lab \cite{mittal2023orbit}.
The goal is to reach target positions in Cartesian space within a fixed distance threshold.
All 3D experiments reported in this work are conducted with gravity enabled.
The state representation includes joint rotations (scaled to $[-1,1]$), end-effector position, and target position.
The threshold for a successful reaching attempt is defined as 0.123\,m.
The 3D task introduces more complex dynamics in the forward model with higher degrees of freedom and joint rotation limits.

\paragraph{Simulation Setup and Evaluation Protocol}
We fix the simulation time step at $\Delta t = 0.02$\,s (50\,Hz) for both the 2D and 3D tasks.
2D and 3D episodes last 200 environment steps while data is collected in parallel across 64 simulated environments.
At the start of each episode, a random robot configuration is sampled to define the target end-effector position, and a new random configuration is selected for the initial state.
The initial velocities are always set to zero.

For evaluation, the 2D task uses 8 hand-crafted initial and target configurations shared across runs.
In the 3D setting, we sample 64 fixed random initial and target states per run.
Although these vary with the seed and prevent exact comparisons between runs, they offer internal consistency.
Unless stated otherwise, x-axes in learning curves refer to training iterations rather than environment or network time steps.

\paragraph{Hyperparameter search}
To identify a robust training configuration, we first applied the necessary methods from \autoref{sec:necessary_steps} to the 2D reaching task, to identify viable time constants ($\tau_{\text{mem}}$, $\tau_{\text{syn}}$, and $\tau_{\text{ada}}$ for ALIF neurons), weight scalings (input, output, recurrent proportion $\rho$, and expected pre-synaptic spike rate $\nu$), learning rates ($\alpha_{\bm{\upsilon}}$ and $\alpha_{\bm{\pi}}$), surrogate gradient function (type and scale $\beta$), network architecture, batch size ($n$), memory capacity ($M$), and unroll steps ($T_{\bm{\upsilon}}$ and $T_{\bm{\pi}}$), as well as the optional methods outlined in \autoref{sec:optional_methods}.
Hyperparameters optimized for the 2D task did not transfer reliably to the 3D setting, requiring task-specific retuning of temporal parameters and weight scalings.
The search results are shown in the \hyperref[sec:appendix]{Appendix} and all final hyperparameters used in the reported experiments are summarized in \autoref{sec:hyperparameters}.

\begin{table}
    \centering
    \caption{\textbf{Overview of task performance metrics (top) and network spiking statistics (bottom).} Arrows indicate whether higher ($\uparrow$) or lower ($\downarrow$) values are preferred.}
    \label{tab:performance_metrics}
    \footnotesize
    \begin{tabularx}{\linewidth}{@{} p{34mm} L c @{}}       \toprule
        \textbf{Metric} & \textbf{Description} & \textbf{Preferred} \\
        \hline
        Mean Cumulative Distance & Average Euclidean distance to the target over the episode. & $\downarrow$ \\
        Success Rate & Fraction of episodes in which the end-effector reaches within a fixed distance threshold of the target. & $\uparrow$ \\
        Time on Target & Number of time steps during which the end-effector remains within the target zone. & $\uparrow$ \\
        Prediction MSE & Mean squared error of the prediction model during autoregressive rollout. & $\downarrow$ \\
        \midrule
        Mean Active Neurons & Number of neurons that spike at least once per episode, averaged across all layers. & $\uparrow$ \\
        Mean Spike Activity & Average number of spike events per episode, normalized from 0 (no spiking) to 1 (constant spiking). No preferred direction. & --- \\
        \bottomrule
    \end{tabularx}
\end{table}

%% file: results.tex
\section{Results}
\label{sec:results}

We systematically evaluated the training and control performance of the Pred-Control SNN on a 6-DOF reaching task using a torque-controlled 3D robotic arm. 
This setup served to assess whether key components from deep learning pipelines, such as trainable time constants, adaptive thresholds, and latent-space compression, transfer effectively to fully spiking architectures.
Our findings are based on extensive ablation experiments and provide several insights into SNN training dynamics and model performance.

We compare four model classes in the main results:
(i) a \emph{full Pred-Control SNN} (large variant, 2048 neurons per layer, $\approx$0.9M parameters) combining learnable time constants, ALIF neurons, and latent-space compression;
(ii) a \emph{basic SNN} that disables these components (512 neurons per layer, $\approx$0.8M parameters);
and non-spiking recurrent baselines with identical training pipelines but standard floating-point RNN modules, evaluated in both a \emph{small} (S) (512 neurons per layer, $\approx$0.8M parameters) and \emph{large} (L) (2048 neurons per layer, $\approx$12.7M parameters) configuration.
All models are trained under identical data collection, unrolling, and optimization protocols. A more complete overview on baseline model performance is given in \autoref{sec:rnn_baseline}.

\begin{figure}[h!]
    \centering
    \includegraphics[width=0.85\linewidth]{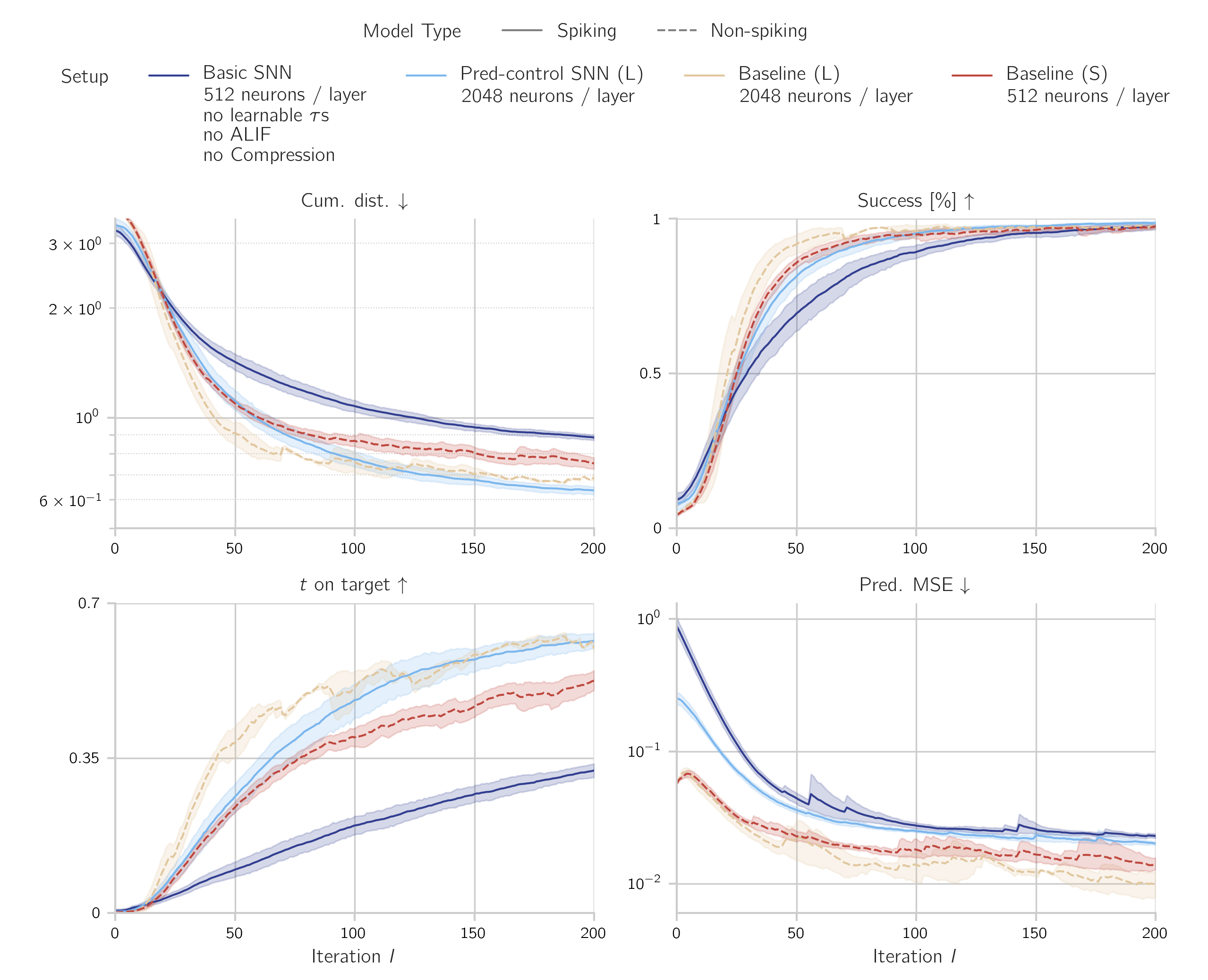}
    \caption{
    \textbf{Main results on the 3D reaching task.}
    Learning curves comparing a non-spiking recurrent baseline (small and large variants), a basic spiking controller, and the full Pred-Control SNN (large).
    Top row: task-level performance metrics (cumulative distance to target, success rate, time on target, and Prediction MSE).
    The Pred-Control SNN achieves task performance on par with the large non-spiking baseline despite using more than an order of magnitude fewer parameters, while clearly outperforming the basic SNN and the small non-spiking baseline.
    The non-spiking baselines converge slightly faster and achieve lower long-horizon prediction error, but this advantage does not translate into superior control performance.
    The Pred-Control SNN learns more gradually but exhibits smoother and more stable optimization dynamics.
    Curves show mean $\pm$ s.e.m. over 10 random seeds.
    }
    \label{fig:6dof-comparison}
\end{figure}

% Task-level performance and parameter efficiency
As shown in \autoref{fig:6dof-comparison}, the full Pred-Control SNN reaches comparable cumulative distance, success rate, and time-on-target performance to the large non-spiking baseline, while using more than an order of magnitude fewer parameters.
Both models substantially outperform the basic SNN, indicating that architectural extensions beyond a minimal spiking controller are required to solve the 3D task reliably.
The small non-spiking baseline achieves reasonable control performance but remains consistently below both large models, highlighting the importance of model capacity for this task.

% Prediction accuracy versus control quality
The non-spiking baselines achieve lower Prediction MSE during long-horizon autoregressive rollouts, reflecting more accurate forward predictions.
However, this advantage does not translate into improved control performance.
Despite higher Prediction MSE, the full Pred-Control SNN matches or exceeds the task-level metrics of the baselines.
This suggests that effective control does not require a perfectly accurate predictive model, but rather a model that is sufficiently expressive to capture task-relevant dynamics and provide stable gradients for policy optimization.

% Learnable time constants
We find that adding learnable time constants consistently improves performance, especially when paired with appropriately scaled learning rates (see \autoref{sec:learnable_timescales} and \ref{sec:snn_ablation}).
This adaptation enables the model to fine-tune its intrinsic temporal dynamics in response to task demands and improves both convergence and final performance, partly mitigating the sensitivity to initialization.
Notably, we found that assigning higher learning rates to time constants than to weights and biases improves convergence speed and stability without destabilizing training.

% Role of ALIF neurons
Replacing the standard LIF neurons with adaptive LIF (ALIF) units further enhances performance (see \autoref{sec:adaptive_lif} and \ref{sec:snn_ablation}). 
Our ALIF neurons incorporate two key mechanisms: I) a linearly decaying threshold to force spiking behavior and II) an adaptive component that increases with recent activity and gradually relaxes back to baseline.
This mechanism provides dual benefits: it enables previously inactive neurons to re-engage in encoding and learning, and it prevents persistent overactivation of units during long episodes. 
Together, these dynamics facilitate better credit assignment over time and help stabilize network activity during training and inference.

% Compression and spiking representation
We also observe clear benefits from introducing latent-space compression between the spiking layers (see \autoref{sec:reducing_parameters} and \ref{sec:snn_ablation}). 
This method reduces the dimensionality of inter-layer representations, enabling substantially wider spiking layers at a fixed parameter budget.
In our best-performing compressed architecture, the full Pred-Control SNN with 2048 spiking neurons per layer and a 64-dimensional latent bottleneck clearly exceeds the performance of uncompressed models with similar parameter counts.
The same model shows comparable performance to the large baseline, despite having \emph{14$\times$} fewer parameters.
While removing compression yields marginal absolute gains in some metrics, this comes at a dramatic increase in model size, indicating that compression provides an overall favorable performance–efficiency trade-off.
We hypothesize that the higher neuron count improves precision in encoding and decoding continuous-valued signals, while the compression enforces a compact, task-relevant representation that aids generalization.
A detailed quantitative comparison is provided in \autoref{sec:snn_ablation}.

% Summary of ablation insights
Taken together, our ablations show that each component, learnable time constants, adaptive thresholds, and latent compression, offers a clear performance benefit. 
Their contributions are complementary: time constant learning mitigates sensitivity to initialization, ALIF neurons enhance credit assignment, and compression improves spiking representation efficiency under parameter constraints.
These findings support our design choice to combine all three in the final architecture.

\begin{figure}[h!]
    \centering
    \includegraphics[width=0.72\linewidth]{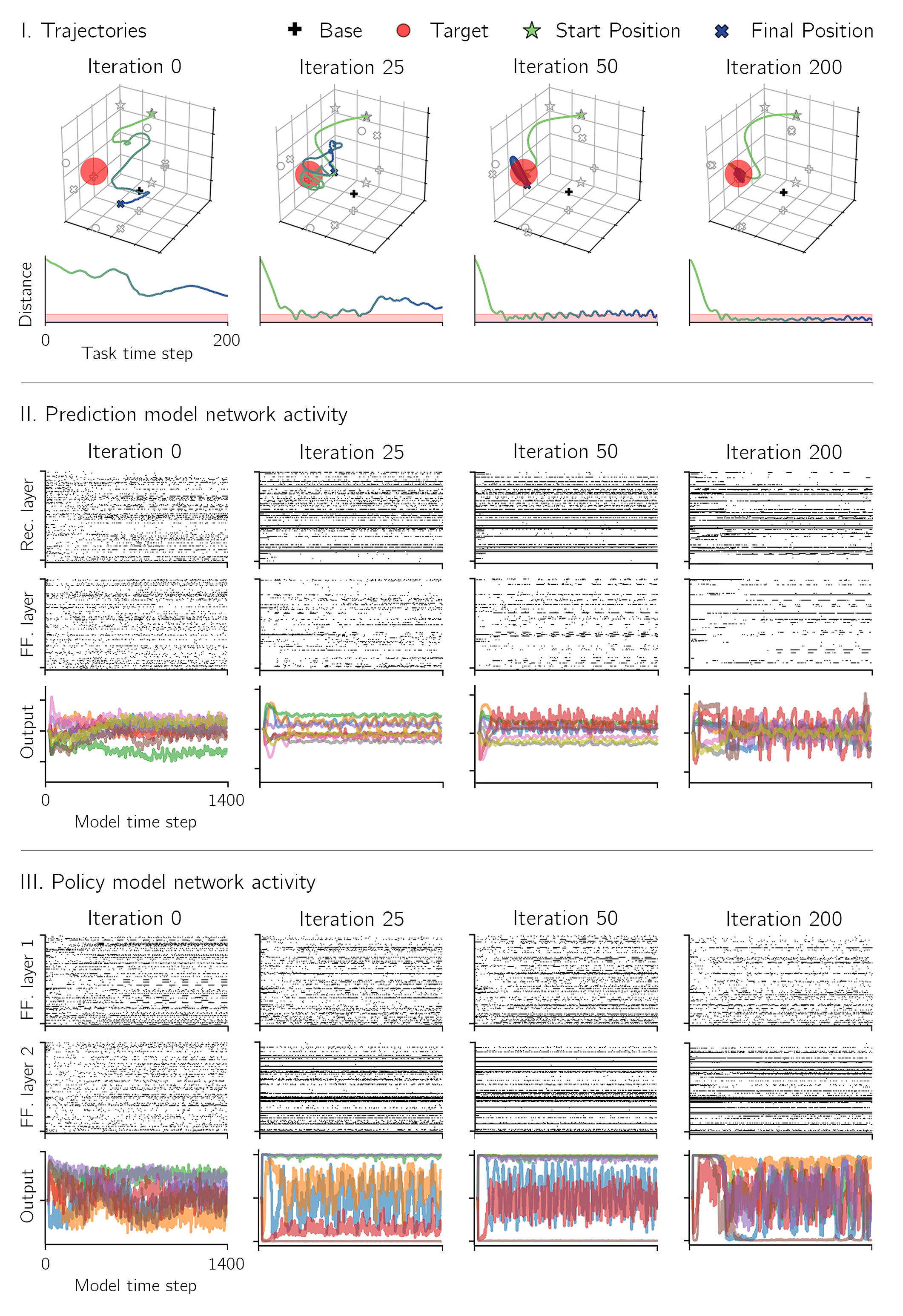}
    \caption{
    \textbf{Pred-Control SNN behavior during task execution.}
    Shown are 3D trajectories and Euclidean error traces (I) over the course of training, along with spiking activity (II) and voltage traces (III) from the prediction and policy networks. 
    Top-row plots are shown over 200 environment time steps per episode.
    The middle and bottom rows are plotted over internal spiking model time, where each environment step corresponds to 7 spiking substeps, resulting in 1400 model time steps per episode.
    This distinction reflects the multi-timescale structure of the controller and is consistent across all experiments.
    The model exhibits progressively smoother control and more consistent activity as training proceeds.
    Spiking activity in the policy network becomes denser and more regular, while the prediction network shows and opposing trend.
    Mild oscillatory behavior near the target is visible in later training stages.
    }
    \label{fig:6dof-example}
\end{figure}

% Learning dynamics and network behavior
\autoref{fig:6dof-example} illustrates representative trajectories and internal network dynamics during task execution. 
As training progresses, end-effector trajectories become smoother and more consistent, and distance-to-target errors decrease steadily.
Simultaneously, the membrane voltages and spike trains of both the prediction and policy networks stabilize, indicating that the network develops a robust internal representation of both control dynamics and task structure.
Notably, as training stabilizes, spiking activity in the policy network becomes denser and more regular than in the prediction network, consistent with the policy’s role in producing continuous control signals rather than sparse state-change representations.
We interpret this as a sign that the model transitions from exploration to a settled control regime.
At later training stages, small oscillations around the target can be observed in the distance traces, despite high time-on-target performance.
We further evaluate robustness to within-episode target shifts and observe rapid re-acquisition of perturbed targets without online learning (see \autoref{sec:robustness}).

% Closing sentences
These experiments confirm that fully spiking networks, when equipped with adaptive temporal dynamics and efficient representations, can solve continuous, high-dimensional motor tasks with high accuracy and stability comparable to non-spiking recurrent baselines. 
Importantly, this is achieved with substantially fewer parameters and without requiring highly accurate long-horizon predictions, highlighting a key distinction between prediction fidelity and control effectiveness in model-based spiking systems.
Altogether, our findings demonstrate that SNNs can match the functional demands of complex control settings, opening the door to scalable, adaptable, and energy-efficient spiking systems for robotics and beyond.

\FloatBarrier

%% file: conclusion.tex
\section{Discussion and Conclusion}
\label{sec:conclusion}

% Framing & conceptual summary
Spiking neural networks offer a compelling model of temporal computation for embodied systems, drawing inspiration from biological principles while promising advantages in energy efficiency and dynamic signal processing.  
Previous work has demonstrated that SNNs can be applied to continuous control, often via equation-based controllers \citep{slijkhuis2023closed, agliati2025spiking} or local adaptation frameworks such as the Neural Engineering Framework \citep{dewolf2016spiking, iacob2020models, marrero2024novel}.  
Yet these approaches typically remain limited to low-dimensional tasks or constrained by analytical assumptions.  
Our work demonstrates that, when equipped with modern training tools adapted from deep learning, spiking architectures like the Pred-Control SNN can scale to high-dimensional robotic control.  
To our knowledge, this is the first demonstration of a fully spiking, end-to-end trainable predictive-control architecture operating in high-DoF continuous robotic arm tasks, helping to close a long-standing gap between biologically inspired modeling and practical machine learning.  
This positions our study beyond earlier demonstrations of spiking controllers, showing that deep-learning–style methods enable SNNs to function as competitive continuous controllers rather than as proof-of-concept demonstrations.

% Key findings in narrative form
By training our Pred-Control SNN using surrogate gradients and evaluating a suite of architectural extensions, we systematically identified several components that significantly affect performance and learning dynamics.  
Adaptive thresholds improved neuron participation and supported sparse activity patterns when well-tuned, consistent with prior findings on adaptive spiking neurons in classification settings \citep{yin2021accurate}.  
Decaying thresholds enabled silent neurons to reactivate over time, serving as an effective alternative to static lower-bound activity constraints.  
Spike regularization reduced bursting but required careful calibration to avoid performance loss.
Learning time constants, especially on a per-neuron basis in log space, offered flexibility to compensate for poor initialization, echoing stabilization strategies such as fluctuation-driven initialization \citep{rossbroich2022fluctuation}, provided that their learning rates were sufficiently high.  
Additionally, loss shaping and latent space compression translated well from ANN training, aiding credit assignment and signal precision within the constraints of sparse spiking dynamics.
Viewed together, learnable time constants, adaptive thresholds, and latent-space compression address complementary aspects of temporal credit assignment in spiking predictive control.
However, their combined effect enable stable, fast, and competitive spiking control.

% Prediction vs control
A central insight emerging from our results is that prediction accuracy does not directly translate to control quality.
While the non-spiking baselines achieve substantially lower long-horizon Prediction MSE, this advantage does not yield superior task-level performance.
Instead, the Pred-Control SNN achieves comparable or better control despite less accurate autoregressive predictions.
This suggests that effective control requires a predictive model that is sufficiently expressive to capture task-relevant dynamics and provide stable gradients for policy optimization, rather than one that minimizes prediction error in isolation.
From a model-based control perspective, this aligns with the notion that approximate forward models can be entirely sufficient for planning and control when errors remain structured and bounded.

% Broader implications and framing of contributions
Together, our results confirm that many principles of deep learning, including dynamic adaptation, sparsity control, and structured dimensionality reduction, are not only compatible with spiking architectures but may be essential for unleashing their potential in continuous tasks.  
They offer a reproducible blueprint for making SNNs robust, trainable, and effective in real-valued motor control.  
This moves beyond earlier demonstrations in arm reaching with STDP or NEF rules \citep{fernandez2021biological, juarez2022r, dewolf2016spiking}, neuromorphic hardware pilots \citep{zhao2020closed, dewolf2023neuromorphic, paredes2024fully}, or policy-focused surrogate gradient RL \citep{tang2021deep, park2025designing, oikonomou2023hybrid, zanatta2024exploring}.  
In contrast, our predictive-control architecture explicitly integrates forward modeling with control, positioning this work as a step toward scalable spiking systems that unify deep learning training with predictive computation.  
For the neural computation audience, this suggests that SNNs can move from classification benchmarks toward the kind of adaptive, embodied computation that biological systems perform.

% Limitations and potential remedies
Our study has limitations worth noting.  
First, the reliance on backpropagation-through-time introduces significant computational overhead in both memory and runtime, an issue exacerbated in SNNs due to their internal sub-timesteps and fine-grained temporal resolution \citep{neftci2019surrogate, zenke2021remarkable}.  
This imposes a bottleneck for scaling to longer task horizons, real-time execution, or complex multi-agent scenarios, and represents a central obstacle to broader applicability.  
Second, while our models generalize across varying initial conditions, they remain sensitive to hyperparameters such as regularization strength and adaptation rates.  
From a robotics or ML perspective this is a limitation, but from a neural computation perspective it may be better viewed as a result: biological systems also rely on finely tuned dynamics to maintain robustness.  
Future work should explore additional homeostatic mechanisms inspired by biological regulation, akin to threshold adaptation or learnable time constants.  
Likewise, incorporating structured connectivity patterns such as partially inhibitory lateral connections or sparsity-promoting weight matrices may enable decorrelation across neurons while preserving the benefits of distributed population codes.  
Finally, most existing SNN control work, including our own, currently uses rate coding for motor output and state representation, leaving the potential of alternative spike coding schemes for control largely untapped \citep{slijkhuis2023closed}.

% Oscillations
Additionally, we observe mild oscillatory behavior near the target in some trajectories at later training stages for the Pred-Control SNN as well as the non-spiking baseline models.
These oscillations occur despite high success rates and time-on-target performance and likely reflect a combination of local attractor dynamics exhibited by the policy network, limited prediction horizon, and the absence of strong action-smoothing penalties.
Notably, the controller is not explicitly incentivized to come to rest at the target and the $\ell_2$ loss is weak close to the target position.
We explored alternative loss formulations, including mixed $\ell_1/\ell_2$ distance penalties designed to increase attraction toward the target center, but these did not reliably eliminate oscillations (data not shown).
Similarly, action regularization reduced overall movement speed and time-on-target but did not fully suppress oscillatory behavior unless set to levels that prevented motion altogether (see \autoref{sec:action_reg_loss}), suggesting a trade-off between control smoothness and learning efficiency.
This suggests that oscillations arise not from instability in the spiking dynamics, but from an underconstrained stopping criterion in continuous control via supervised learning, a challenge shared with non-spiking controllers.
Future work could explore structured action regularizers, multi-rate control schemes, or predictive uncertainty estimates to mitigate such effects without compromising performance.
An alternative approach to reaching tasks in the domain of reinforcement learning is known by applying more complex reward shaping compared to a simple distance based measure, where extra rewards might be granted close to the target, to give a stronger incentive (steeper loss landscape / stronger attractor) towards the target position.

% Energy efficiency and sparsity
Although spiking neural networks are often motivated by energy efficiency, our work does not explicitly optimize for sparsity or minimal spike counts.
Under a rate-coding assumption, reducing spike activity necessarily trades representational precision for efficiency, which may or may not be desirable depending on task demands.
For continuous motor control, population size and spike density jointly determine how finely continuous variables can be represented.
In principle, equivalent control signals could be produced by larger populations with proportionally lower firing rates, or by discretized action spaces with fewer output levels.
However, in the absence of explicit hardware constraints, sparsity becomes an ambiguous objective.
We therefore adopt a hardware-ignorant perspective: rather than enforcing biological or neuromorphic limits, we focus on identifying architectural principles that enable robust learning and control.
The reported spike activity metrics are thus interpreted as descriptive signals of network dynamics rather than direct proxies for energy consumption.
Explicit energy optimization and deployment on neuromorphic hardware remain important directions for future work.

% Outlook and research trajectory
Looking ahead, we aim to extend this architecture to reinforcement learning tasks, which introduce challenges in exploration, long-term credit assignment, and reward sparsity.  
Our Pred-Control SNN is well-suited for this transition due to its modularity and internal state dynamics.  
However, the reliance on BPTT underscores the importance of exploring alternative learning methods.  
Three complementary directions appear especially promising.  
First, online training approaches such as e-prop \citep{bellec2020solution} and forward propagation through time \citep{yin2023accurate} provide more memory-efficient schemes for recurrent spiking networks.  
Second, noise-driven credit assignment methods such as node and weight perturbation \citep{zuege2023weight} may offer biologically inspired, event-driven alternatives that scale more gracefully than full backpropagation.  
Third, benchmarking these methods on high-dimensional continuous control remains a crucial test.  
MuJoCo-style continuous control environments, already used in recent SNN-based reinforcement learning studies \citep{tang2021deep, zanatta2024exploring}, provide a natural next step for assessing whether these algorithms can scale to complex, high-DoF robotic tasks.  
In parallel, we also intend to explore the generative power of model-based reinforcement learning algorithms adapted to the spiking domain, where predictive components may further reduce reliance on expensive gradient propagation.  
Such work will be crucial for assessing whether spiking systems can rise to the challenges of real-world robotics, offering not only compact and energy-efficient control but also biologically inspired online adaptability.

% Wrap-up and broader vision
In conclusion, this study shows that SNNs, when trained with principled, deep learning–inspired methods, can scale to high-dimensional continuous control without requiring ANN pretraining, conversion, or hardware-specific constraints.  
By demonstrating stable, effective motor control in end-to-end trained spiking systems, we move closer to a new generation of adaptive, low-power control systems that draw on the best of both neuroscience-inspired computation and machine learning.  
Although demonstrated here in simulation, the architectural principles are hardware-agnostic and could guide deployment on emerging neuromorphic platforms, providing a pathway toward real-world, low-power robotics.  
This convergence between machine learning and neuroscience-inspired models points toward the next generation of intelligent, adaptive machines.

%% file: appendix.tex
\section{Learning Rate \texorpdfstring{$\alpha$}{alpha}}
\label{sec:learning_rate}

As with most neural network training regimes, the learning rate is one of the most critical hyperparameters.
Its appropriate setting depends on several factors, including the choice of optimizer, the magnitude of the gradients, and the curvature of the loss landscape.
A common strategy is to begin with a relatively high learning rate to enable rapid initial learning and then reduce it gradually to promote convergence and stability.

In this experiment, we investigate suitable learning rates for two separately trained networks: the policy network $\bm{\pi}$ and the prediction (dynamics) model $\bm{\upsilon}$.
Their respective learning rates are denoted as $\alpha_{\bm{\pi}}$ and $\alpha_{\bm{\upsilon}}$.
We treat both as independent hyperparameters and sample their values logarithmically between $10^{-5}$ and $10^{-1}$.
Training is performed using the Adam optimizer~\cite{kingma2014adam} for both networks, due to its adaptive update rules and strong empirical performance across a wide range of tasks.

\begin{figure}[h!]
    \centering
    \includegraphics[width=0.8\linewidth]{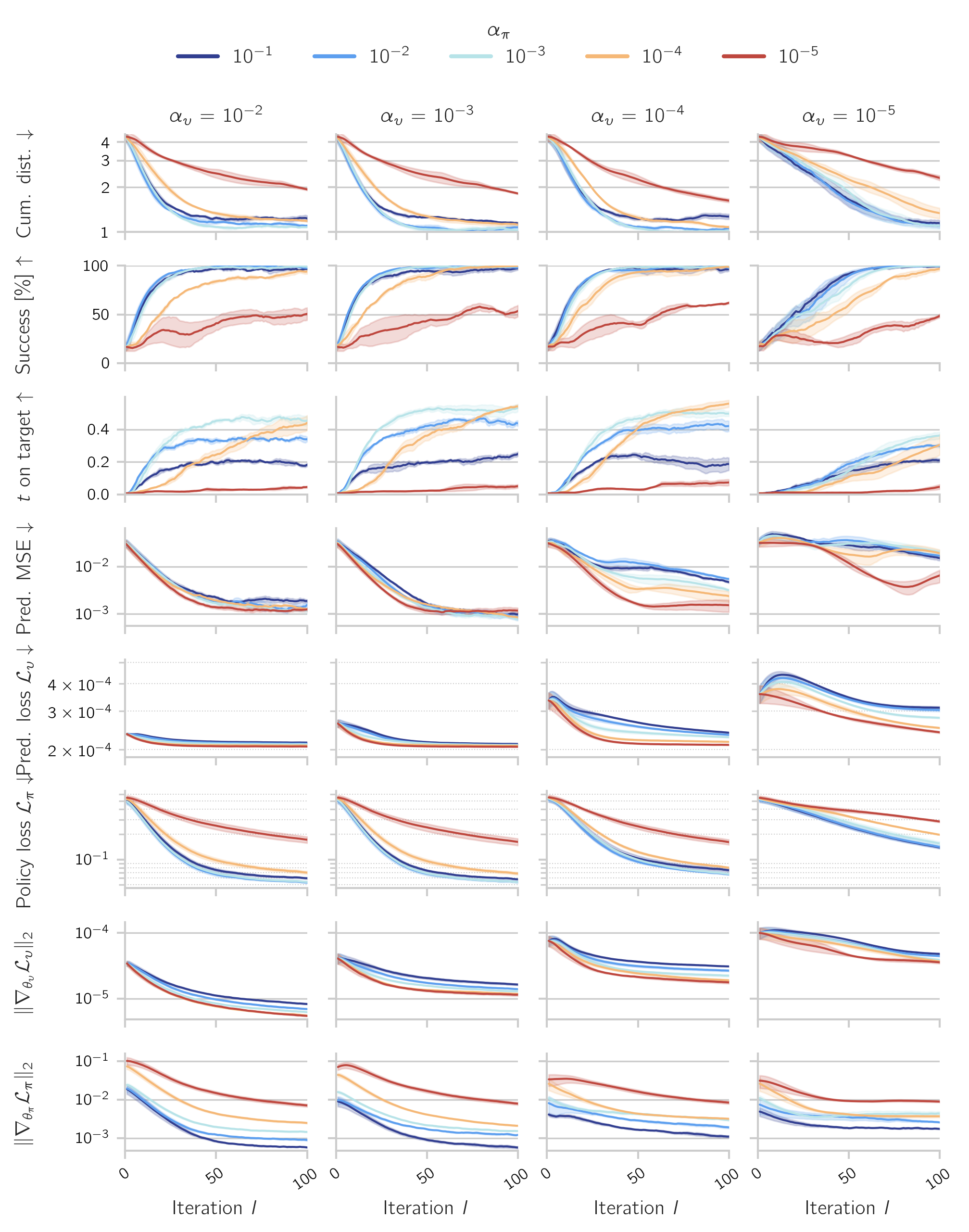}
    \caption{
    \textbf{Effect of learning rates $\alpha_{\bm{\pi}}$ and $\alpha_{\bm{\upsilon}}$ on training in the 2D control task.}
    We vary $\alpha_{\bm{\upsilon}}$ across columns and $\alpha_{\bm{\pi}}$ across line colors.
    Performance is most easily distinguished by the time spent on target (row 3), which clearly peaks when both networks use $\alpha = 10^{-3}$.
    Learning stability is preserved across all settings, with no signs of vanishing or exploding gradients, yet performance still varies markedly with learning rate choice.
    The policy is sensitive to prediction quality, but the inverse does not hold: policy failure does not impair model learning.
    Overall, the best choice of parameter based on the performance metrics was found at $\alpha_{\bm{\pi}} = \alpha_{\bm{\upsilon}} = 10^{-3}$.
    }
    \label{fig:2Dlr_results}
\end{figure}

While most configurations reduce the cumulative distance and increase the success rate to some extent, the clearest separation between well- and poorly-performing models is observed in the time-on-target metric (see \autoref{fig:2Dlr_results}).
This measure integrates both accuracy and stability over time and proves more sensitive than raw loss values, which can decrease even when control performance remains poor.

Notably, the best performance is achieved when both networks use a learning rate of $\alpha_{\bm{\pi}} = \alpha_{\bm{\upsilon}} = 10^{-3}$, which was also selected for all subsequent experiments.
Across all configurations, the recorded gradient norms remain within a stable range and do not exhibit signs of vanishing or exploding.
However, their absolute magnitude shows little correlation with task performance: both poorly and well-performing models can exhibit similarly sized gradients.
This suggests that gradient norm alone is not a reliable indicator of effective learning progress or eventual task success.
Instead, successful training depends more critically on the interaction between the learning rates and the underlying optimization landscape.
It is plausible that the Adam optimizer mitigates the effects of absolute gradient scale differences, allowing training to remain numerically stable even when learning progress diverges.

The results further highlight an asymmetry in the dependency between the two networks: if the prediction model learns poorly, policy optimization fails due to inaccurate gradients.
However, the prediction model can converge even when the policy performs suboptimally, as it is trained independently on state transitions.

Taken together, these findings highlight the importance of co-tuning learning rates and reinforce the central role of accurate dynamics in enabling effective policy optimization.

\paragraph{Scheduled Learning Rates.}

To modulate the learning rate during training, we apply an exponentially decaying schedule defined by a decay factor $\gamma \in \{1.0, 0.99, 0.97, 0.9\}$, where $\gamma = 1.0$ corresponds to a constant learning rate.
Decay is applied once per epoch according to $\alpha_t = \max(\gamma^t \alpha_0, \alpha_{\text{min}})$, with $\alpha_{\text{min}} = 0.0001$ ensuring that learning does not halt entirely in later training stages.

\begin{figure}[h!]
    \centering
    \includegraphics[width=0.6\linewidth]{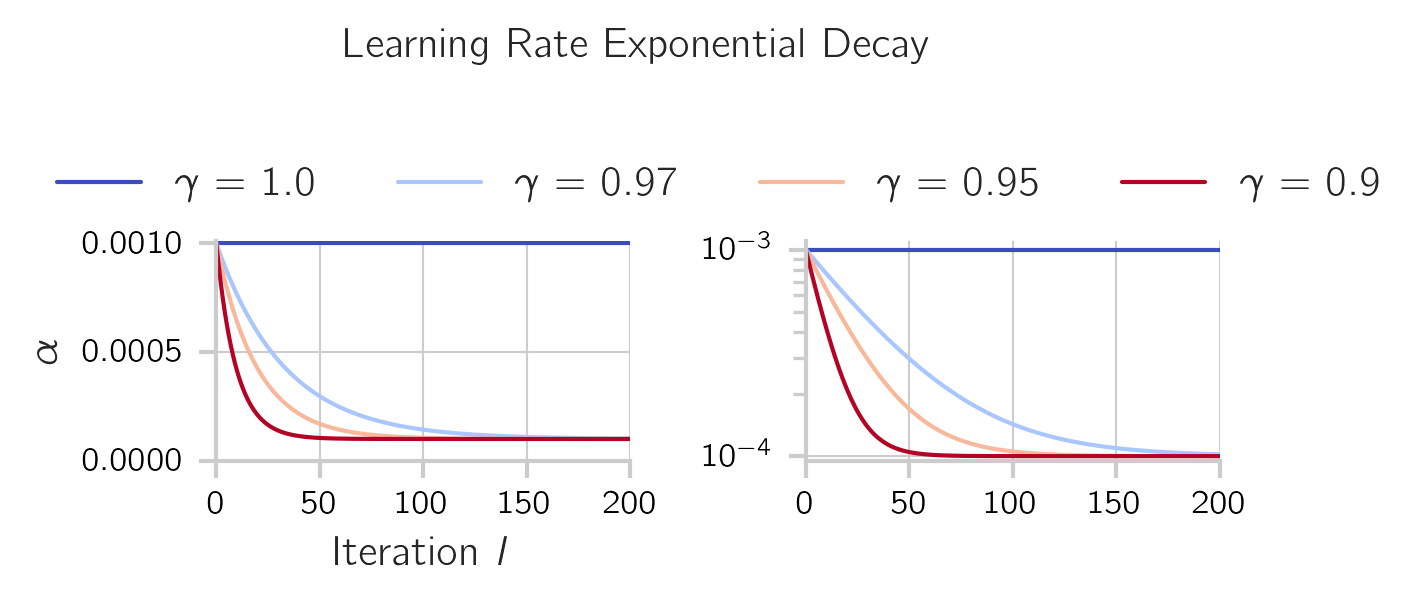}
    \includegraphics[width=0.9\linewidth]{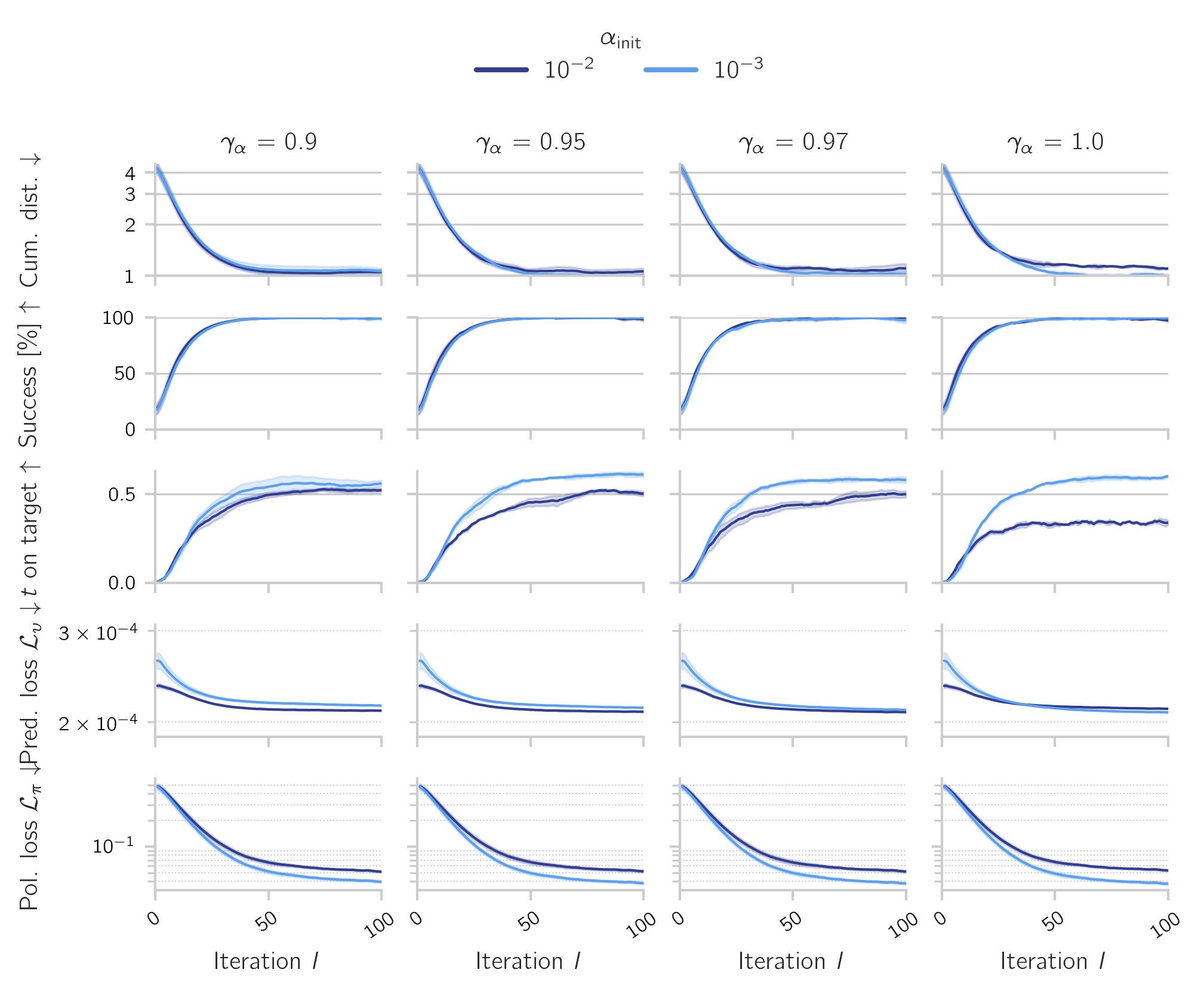}
    \caption{
    \textbf{Exponential learning rate decay.}
    \textbf{Top:} Exponential decay schedules for different values of $\gamma$, starting from $\alpha_0 = 0.001$.
    All schedules are bounded below by $\alpha_{\text{min}} = 0.0001$ to prevent the learning rate from vanishing entirely.
    \textbf{Bottom:} Control performance for different decay factors $\gamma$ and two initial learning rates $\alpha_{\text{init}} \in \{10^{-3}, 10^{-2}\}$, applied symmetrically to both networks.
    While exponential decay mitigates instability at higher initial learning rates, performance on cumulative distance and success rate remains similar across all settings.
    The time-on-target metric reveals a consistent advantage for the constant $10^{-3}$ schedule.
    }
    \label{fig:lr_decay}
\end{figure}

We evaluate the effect of exponential decay schedules on training stability and performance for both the policy and prediction networks.
Each decay factor is tested with initial learning rates of $\alpha_0 = 10^{-3}$, the best-performing value identified previously—and a larger alternative $\alpha_0 = 10^{-2}$.

As shown in \autoref{fig:lr_decay}, decay schedules can compensate for overly aggressive initial rates to some extent, but do not yield improvements over the constant baseline.
This is particularly clear in the time-on-target metric, which remains highest and most stable for the non-decayed $10^{-3}$ setting.
More aggressive decay (e.g., $\gamma = 0.9$) reduces the effective learning rate too quickly and leads to premature plateauing, while milder decay schedules offer no tangible advantage.

Overall, exponential decay provides a degree of robustness in unstable configurations but does not improve performance when a well-tuned constant learning rate is already available.
It is plausible that the Adam optimizer already compensates for learning rate scale variations through its adaptive update mechanism, reducing the potential benefits of external scheduling.
For simplicity and consistency, we therefore use a constant learning rate of $\alpha = 10^{-3}$ for both networks in all subsequent experiments.

\FloatBarrier
\clearpage

\section{Surrogate Gradient Function \texorpdfstring{$f'(U)$}{f'(U)}}
\label{sec:surrogate_gradients}

\begin{figure}[h!]
    \centering
    \includegraphics[width=1.0\linewidth]{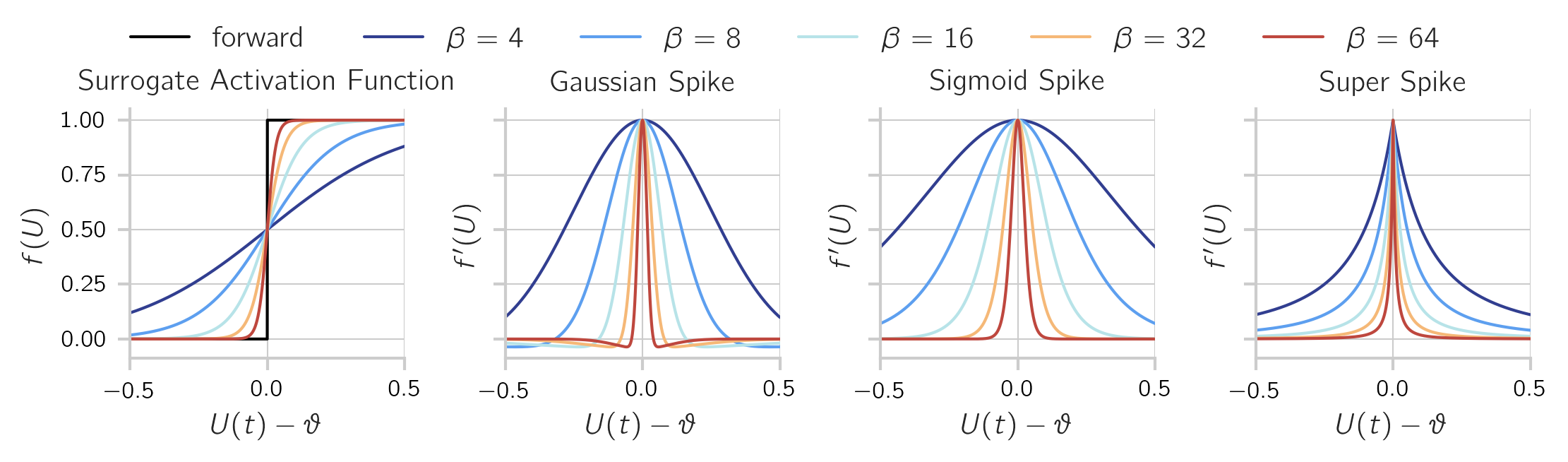}
    \includegraphics[width=0.9\linewidth]{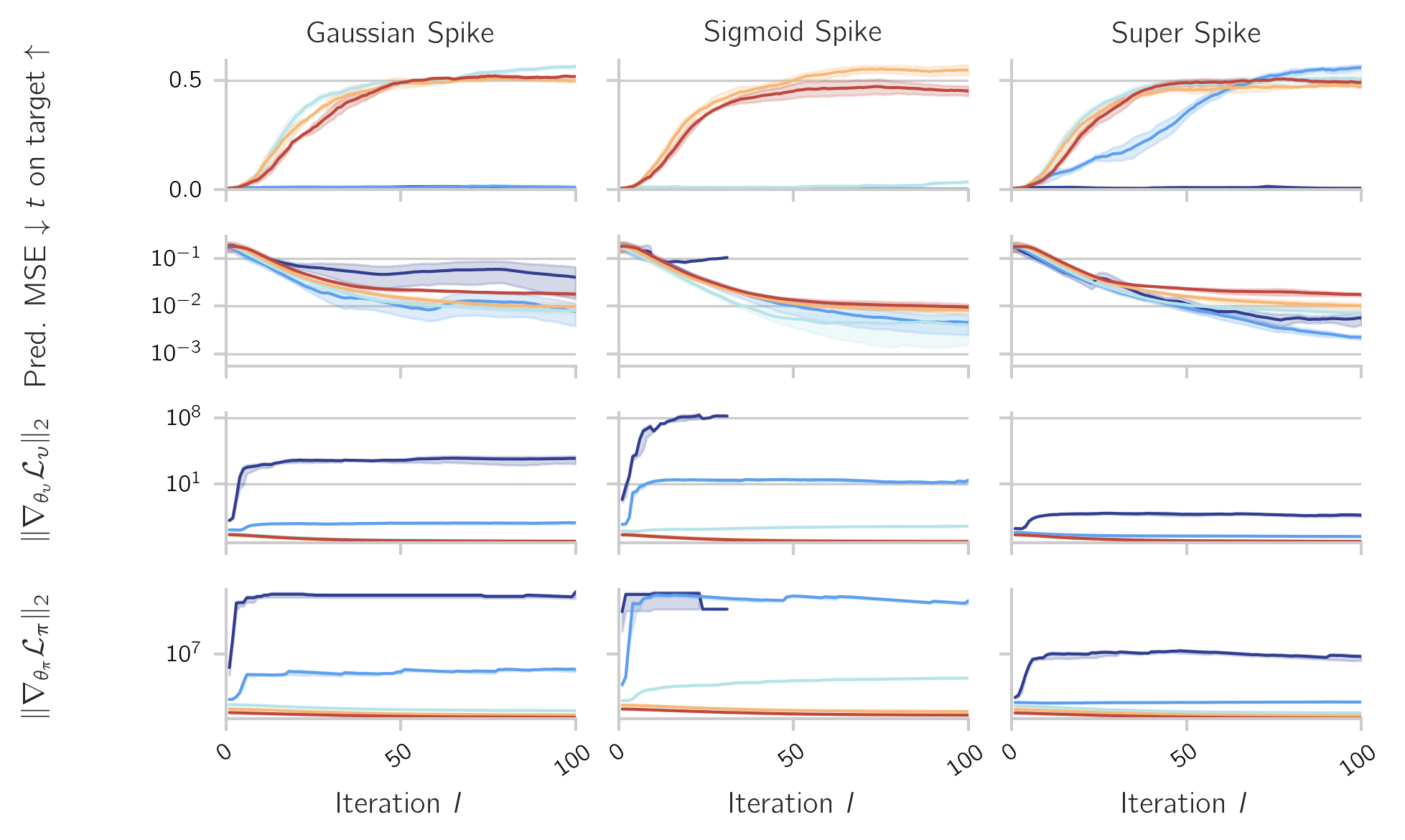}
    \caption{
    \textbf{Surrogate gradient functions and training outcomes.}
    \textbf{Top:} Gradient profiles $f'(U)$ of the Sigmoid, SuperSpike, and Gaussian surrogates across different steepness values $\beta$.
    Higher $\beta$ localizes the gradient more sharply around the spike threshold.
    \textbf{Bottom:} Training dynamics using each surrogate with varying $\beta$ in the 2D control task.
    All three surrogates can support effective learning if $\beta$ is appropriately tuned.
    Low $\beta$ values lead to unstable gradient magnitudes and training failure, while moderate values yield comparable time-on-target and prediction performance.
    We proceed with the Gaussian Spike surrogate at $\beta = 16$ for the remainder of this study.
    }
    \label{fig:surrogate_grad}
\end{figure}

Training spiking neural networks (SNNs) with backpropagation relies on surrogate gradients $f'(U)$ to approximate the non-differentiable spike function during the backward pass~\citep{neftci2019surrogate, zenke2021remarkable} (see \autoref{fig:surrogate_grad}, top).
Several surrogate functions have been proposed, each differing in shape and defined by key parameters: the steepness $\beta$, which controls the sharpness of the response around threshold, and the scaling factor $\gamma$, which sets the overall gradient amplitude.

Surrogate gradients do not operate in isolation.
Their effectiveness depends on interactions with broader dynamical parameters that shape gradient flow, including the membrane and synaptic time constants ($\tau_{\text{mem}}$, $\tau_{\text{syn}}$), spike activity levels, and initialization parameters such as the baseline firing rate $\nu$.
Sparser activity or rapidly decaying synapses reduce the temporal window over which gradients can propagate.

A full grid search over all interactions is computationally infeasible.
Instead, we evaluate the practical effect of different surrogate functions and steepness values while keeping other parameters fixed at reasonable working defaults.

\vspace{0.5em}
In this experiment, we compare three commonly used surrogate gradient functions:

\begin{itemize}
    \item \textbf{Sigmoid surrogate}, based on the derivative of the logistic function
    \item \textbf{SuperSpike}~\citep{zenke2018superspike}, a heavy-tailed approximation with smoother decay
    \item \textbf{GaussianSpike}~\citep{yin2021accurate}, a mixture-based method with localized support
\end{itemize}

Each function defines a differentiable approximation to the spike function's gradient during training.
The Sigmoid Spike surrogate is defined as:
\begin{equation}
\label{eq:sigmoidspike}
\frac{\partial L}{\partial x} = \frac{\gamma}{n_{\text{sig}}} \left( \sigma(\beta x) \left[1 - \sigma(\beta x)\right] \right) \frac{\partial L}{\partial y},
\end{equation}
where $\sigma$ is the logistic sigmoid and $n_{\text{sig}} = 0.25$ normalizes the peak gradient to 1 when $\gamma = 1$.

The Super Spike approximation is:
\begin{equation}
\label{eq:superspike}
\frac{\partial L}{\partial x} = \gamma \left( \frac{1}{(\beta |x| + 1)^2} \right) \frac{\partial L}{\partial y}.
\end{equation}

The Gaussian Spike surrogate is defined by a weighted mixture of Gaussians:
\begin{equation}
\label{eq:gaussian_function}
G(x; \mu, \sigma) = \frac{\exp\left(-\frac{(x - \mu)^2}{2 \sigma^2}\right)}{\sqrt{2\pi} \sigma},
\end{equation}
\begin{equation}
\label{eq:gaussianspike}
\frac{\partial L}{\partial x} = \frac{\gamma}{n_{\text{gaus}}} \left[ G\left(x; 0, \tfrac{1}{\beta}\right)(1 + h) - G\left(x; \tfrac{1}{\beta}, s \tfrac{1}{\beta}\right)h - G\left(x; -\tfrac{1}{\beta}, s \tfrac{1}{\beta}\right)h \right] \frac{\partial L}{\partial y},
\end{equation}
with constants $h=0.15$, $s=6$, and normalization factor $n_{\text{gaus}}$ ensuring unit peak when $\gamma = 1$.

\autoref{fig:surrogate_grad} (top) visualizes these surrogate profiles for varying steepness $\beta$. 
Higher $\beta$ values result in sharper gradients concentrated near threshold.
We fix $\gamma = 1.0$ throughout and vary $\beta$ to assess each function’s impact on training performance.

The bottom panel of \autoref{fig:surrogate_grad} shows the results of these experiments in the 2D control task.
For each surrogate function, we observe that a moderate steepness $\beta$ yields the fastest and most stable learning.
While overly low $\beta$ values result in shallow gradients and slow convergence, excessively large values can trigger gradient explosion and numerical instability—especially in the Sigmoid and Gaussian surrogates.
SuperSpike appears slightly more tolerant to variation in $\beta$, but all three surrogates support learning when well-tuned.
At their respective optimal $\beta$ values, final task performance is comparable across surrogate types.

Since no surrogate clearly outperforms the others, we continue with the Gaussian Spike surrogate at $\beta = 16$ for all subsequent experiments, chosen for its reliable and comparably fast convergence in this setting.

\FloatBarrier
\clearpage

\section{Empirical Initialization}
\label{sec:empirical_init}

Spiking neural networks are highly sensitive to weight initialization, as only specific parameter regimes yield sufficient activity and stable learning.
We adopt a fluctuation-driven initialization scheme~\citep{rossbroich2022fluctuation}, which incorporates the neuron time constants $\tau_{\text{mem}}$ and $\tau_{\text{syn}}$ into the weight scaling rules.
This method aims to place the network in a dynamic regime where neurons spike regularly but not excessively, facilitating gradient-based learning.
Unlike standard methods such as Kaiming or Xavier initialization—which do not account for temporal filtering—this scheme is better suited to SNNs with explicit dynamics.

The original formulation requires an estimate of $\nu$, the expected presynaptic spike rate per neuron per second.
While this can be computed analytically for static datasets, our closed-loop control setting generates inputs online.
We therefore treat $\nu$ as a tunable hyperparameter and determine suitable values empirically via grid search.

We assess initialization quality using two criteria: (i) the fraction of neurons that spike at least once per episode (per layer), and (ii) the average spike rate per neuron.
We do not enforce a specific target rate but seek configurations that preserve signal propagation without saturating or silencing activity.

Although our networks are relatively shallow (3 layers), they are unrolled over many time steps during task execution, producing significant temporal depth.
This contrasts with feedforward SNNs used in classification tasks~\citep{rossbroich2022fluctuation}, where stability was more closely linked to performance.
Here, we observe that well-performing configurations are not always those with the lowest loss or the highest activity levels, indicating that different metrics may emphasize different aspects of task behavior.

We fix the target membrane statistics to $\mu_U = 0$ and $\sigma_U = 1$ for all initializations.
The full derivation of how $\nu$, $\tau_{\text{mem}}$, and $\tau_{\text{syn}}$ influence the initialization scale is provided in~\citep{rossbroich2022fluctuation}.

\autoref{fig:lif_kernels} shows how $\tau_{\text{mem}}$ and $\tau_{\text{syn}}$ shape the temporal dynamics and firing behavior of individual neurons.
We use this as a reference when selecting the parameter range explored in the grid search.

\begin{figure}[h!]
    \centering
    \includegraphics[width=0.8\linewidth]{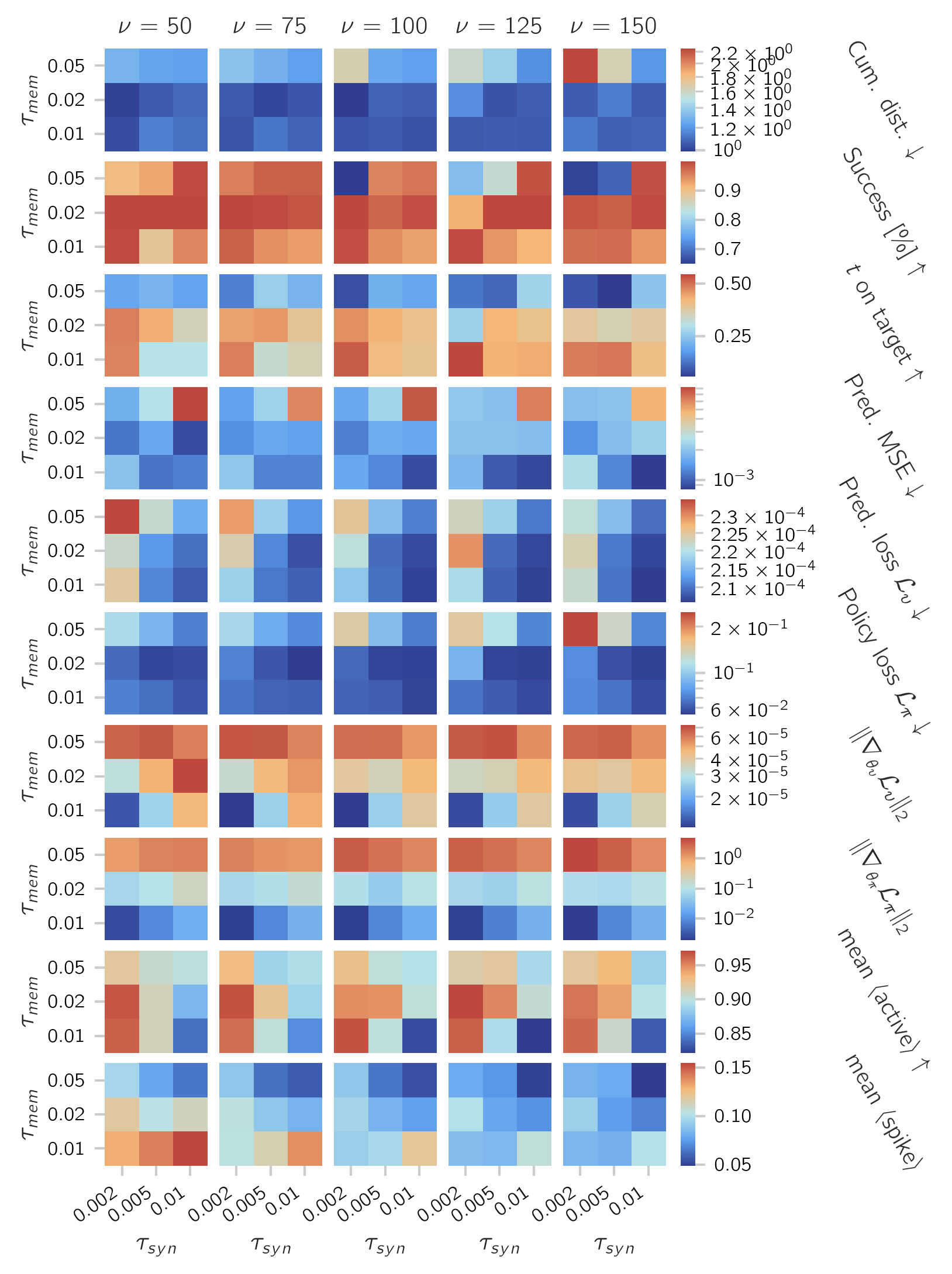}
    \caption{
    \textbf{Influence of initialization parameters on SNN performance, gradients, and spiking activity.}
    Each column corresponds to an empirically selected input rate $\nu$, while each heatmap varies $\tau_{\text{mem}}$ and $\tau_{\text{syn}}$.
    We report task performance (top), prediction quality (middle), and gradient/spike metrics (bottom).
    Each cell shows the average over 3 seeds at the iteration of minimum cumulative distance.
    The influence of $\tau_{\text{mem}}$ and $\tau_{\text{syn}}$ is generally stronger than that of $\nu$ across most metrics.
    Time-on-target peaks for $\tau_{\text{mem}} = 0.01$, $\tau_{\text{syn}} = 0.002$, and $\nu = 125$, which we use for all further experiments.
    Surprisingly, some configurations with low loss do not yield the best behavioral outcomes, indicating a partial decoupling between error reduction and effective control.
    }
    \label{fig:empirical_init_results}
\end{figure}

\autoref{fig:empirical_init_results} summarizes the effect of $\nu$, $\tau_{\text{mem}}$, and $\tau_{\text{syn}}$ on control performance, spiking statistics, and gradient norms.
Each result is averaged over 3 random seeds, and reported at the training iteration where the cumulative distance is minimized.
While cumulative distance is used as the optimization objective, we find that time on target provides a more discriminative view of model quality across hyperparameter settings.

The influence of $\tau_{\text{mem}}$ and $\tau_{\text{syn}}$ is notably more pronounced than that of $\nu$ on most task-related metrics.
By contrast, $\nu$ mainly controls the number of active neurons and spike density, confirming its role in setting the network's baseline excitability.

Interestingly, the configurations with the lowest policy and prediction losses are not those with the highest time-on-target.
This mismatch highlights that minimizing error signals does not always correspond to robust goal-directed behavior—reaching the target briefly and maintaining presence there are qualitatively different challenges.
Moreover, some parameter settings that would be considered unstable by classical criteria (e.g., low spike coverage or imbalanced layer activity) still yield strong task performance.
This stands in partial contrast to prior work on SNN classification~\citep{rossbroich2022fluctuation}, where stable activity and balanced firing were stronger predictors of success.
Our findings suggest that the link between stability and performance is more nuanced in closed-loop, temporally extended control tasks.

We select $\tau_{\text{mem}} = 0.01$, $\tau_{\text{syn}} = 0.002$, and $\nu = 125$ for all subsequent experiments, based on their strong performance on the time-on-target metric.
This setting also yields among the lowest gradient norms for both networks, suggesting that efficient learning can occur with well-conditioned, moderate gradients.
While other settings might perform well with different learning rates, a joint search over learning rates and initialization parameters is computationally infeasible.
While our approach explores a fixed grid of values for $\nu$, $\tau_{\text{mem}}$, and $\tau_{\text{syn}}$, it does not include a mechanism to adapt $\nu$ during training or to automatically tune it to satisfy predefined stability criteria.
Incorporating an outer loop that adjusts $\nu$ to achieve a target spiking profile at initialization time—or during training—could be a promising avenue for future work, particularly in deeper or more recurrent architectures where initialization fragility is more pronounced.

\FloatBarrier
\clearpage

\section{Network Architecture}
\label{sec:network_architecture}

In this section, we investigate the influence of architectural choices on learning performance by systematically varying the structure of both the policy and prediction networks. We focus on two main factors: (i) the number of hidden layers, and (ii) whether the first layer of each network is implemented as a recurrent spiking layer. This design choice determines the temporal context each model can access and, by extension, the type of information it can extract from its input. 
Further, we investigate the shape of the network (neurons per layer and number of layers) to see which setup is sufficient to solve the control task.

\subsection{Recurrent Connections}
\label{sec:recurrent_connections}

We begin our investigation into architectural structure by examining whether recurrent connectivity in the first spiking layer improves learning for the prediction model $\bm{\upsilon}$ and/or the policy model $\bm{\pi}$.

In principle, recurrence is expected to benefit both models. 
The input to $\bm{\upsilon}$ includes the robot's current joint position, but omits velocity information, which must be internally inferred over time. 
Adding a recurrent layer provides the models with increased temporal memory, allowing it to better accumulate information across timesteps and approximate hidden state variables.
Without recurrence, both SNNs would need to rely solely on the implicit memory of LIF neurons, which may be insufficient in this partially observable setting.

\begin{figure}[h!]
    \centering
    \includegraphics[width=\linewidth]{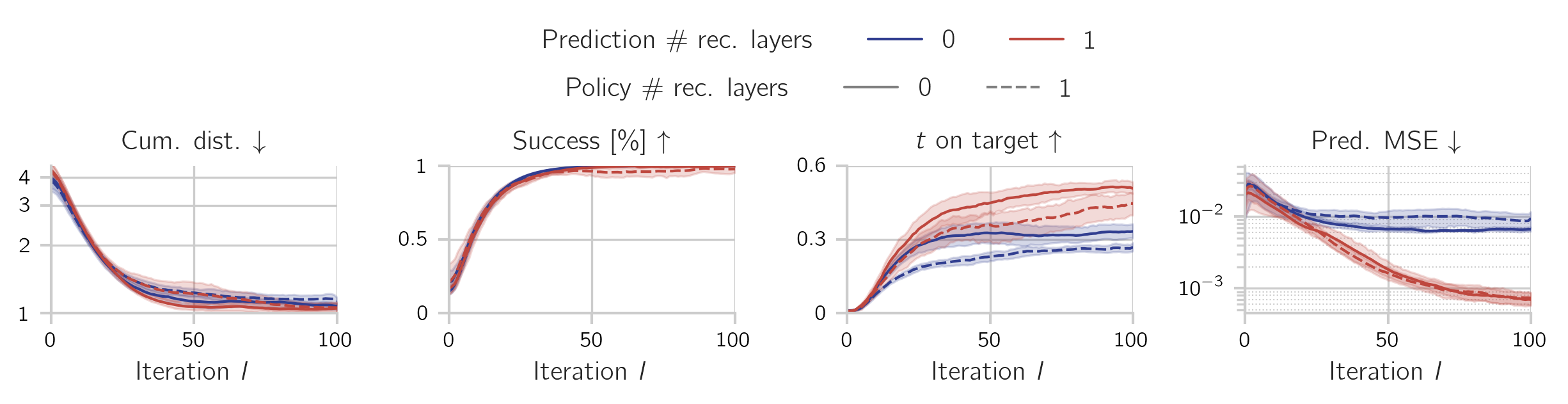}
    \includegraphics[width=0.9\linewidth]{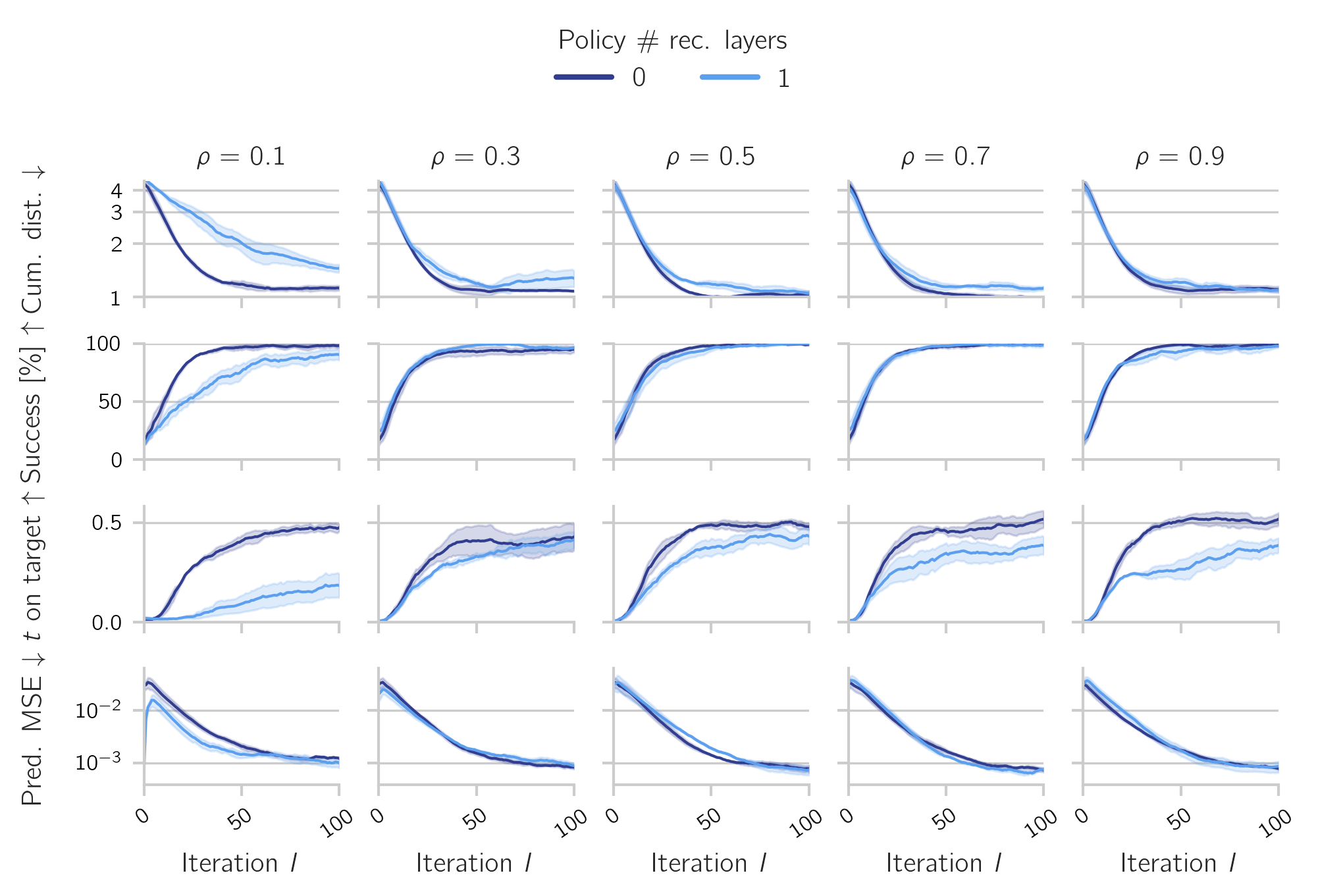}
    \caption{
    \textbf{Influence of recurrence and input/recurrent weight ratio $\rho$ on network performance.}
    \textbf{Top:} Task performance metrics show that no recurrence is necessary to learn the control task.
    Recurrence in the prediction network $\bm{\upsilon}$ clearly improves time on target, while recurrence in the policy $\bm{\pi}$ offers no benefit and often impairs performance.
    \textbf{Bottom:} Detailed sweep over values of $\rho$ from 0.1 to 0.9.
    Excessively strong recurrence (low $\rho$) destabilizes training in the policy network, while moderate recurrence is safe but unnecessary.
    A weakly recurrent prediction model with $\rho = 0.9$ performs well and is used in all following experiments.
    }
    \label{fig:recurrent_architecture_results}
\end{figure}

We first perform a high-level comparison across all tested conditions with and without recurrence in either network (see \autoref{fig:recurrent_architecture_results}).
The results confirm that the models can learn the reaching task without any recurrent connections. 
However, adding recurrence to the prediction model consistently improves both prediction accuracy and task performance.
Surprisingly, recurrence in the policy model either yields no improvement or slightly impairs learning.
The reason for this difference is not obvious.

To explore how strongly the recurrent layer should rely on internal versus external input, we introduce a tunable parameter $\rho \in [0,1]$ that controls the initialization ratio between external inputs and recurrent feedback:
\[
\rho = \frac{\text{external input weight}}{\text{external input weight} + \text{recurrent weight}}.
\]
A value of $\rho = 0.9$ implies a weakly recurrent layer (mostly driven by external input), while $\rho = 0.1$ yields a strongly recurrent layer (dominated by recurrent feedback).
We sweep $\rho$ from 0.1 to 0.9 in steps of 0.2, using the fluctuation-driven initialization strategy of~\citet{rossbroich2022fluctuation}.

\autoref{fig:recurrent_architecture_results} (bottom) shows the results of this sweep, with and without recurrence in the policy model.
We observe that recurrence in the prediction model yields clear benefits across all $\rho$ values.
By contrast, recurrence in the policy model degrades performance for all tested values of $\rho$, likely due to the introduction of excessive memory and less interpretable credit pathways.

Another key finding is that prediction and control performance vary only mildly across the tested range.
For robustness and simplicity, we therefore continue with $\rho = 0.9$ in all subsequent experiments involving recurrent layers.

Together, these findings support the architectural choice of using a recurrent first layer in the prediction model and a purely feedforward structure in the policy model.
This hybrid design achieves the best trade-off between memory capacity and training stability in our control setting.

\subsection{Network Shape}
\label{sec:network_shape}

Spiking neurons such as LIF units emit discrete spikes, producing a binary signal at each timestep.
As a result, individual neurons cannot easily represent continuous quantities.
Instead, effective encoding typically relies on population coding, where multiple neurons collectively represent scalar values through coordinated activity patterns.
This motivates the use of larger hidden layers in SNNs compared to non-spiking neural networks.

In addition to size, network depth is another key architectural factor.
Multiple hidden layers enable richer nonlinear transformations and may be necessary to map raw inputs to suitable control or prediction outputs.
However, increasing depth also introduces optimization challenges and computational cost.

To identify a suitable network configuration, we evaluate task performance across a range of architectures where both the prediction network $\bm{\upsilon}$ and the policy network $\bm{\pi}$ share the same structure.
We sweep the number of spiking layers (1, 2, or 3) and the number of neurons per layer (32, 64, 128, 256, 512, or 1024).

\begin{figure}[h!]
    \centering
    \includegraphics[width=0.8\linewidth]{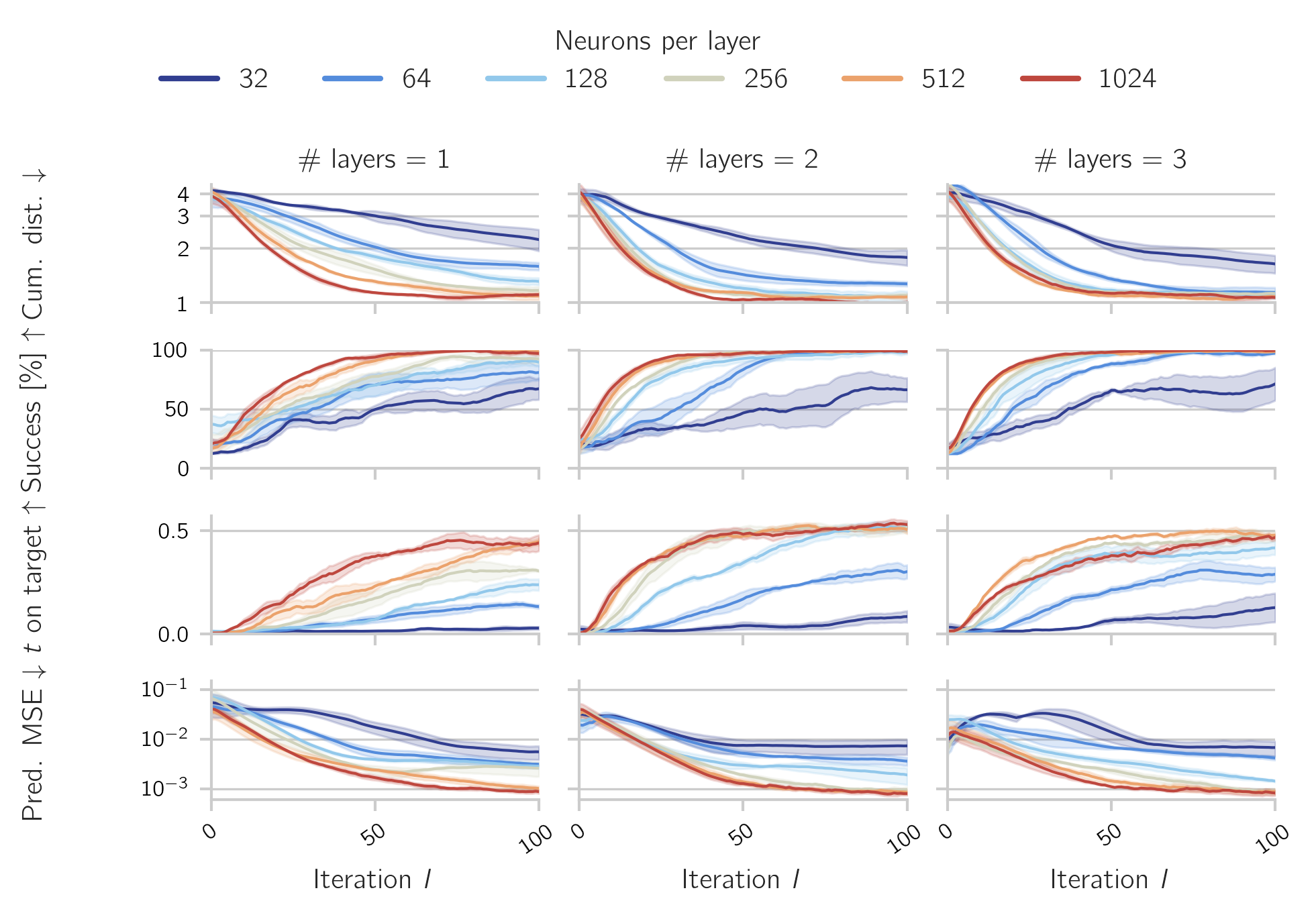}
    \caption{
    \textbf{Effect of network size on task performance.}
    Each column varies the number of spiking layers (1–3), while each line color indicates the number of neurons per layer.
    Metrics include cumulative distance, success rate, time on target, and prediction MSE.
    Performance generally improves with increasing network size, but saturates around 512 units.
    A two-layer architecture provides a good trade-off between expressivity and stability, and is used in all subsequent experiments.
    }
    \label{fig:shape_results}
\end{figure}

\autoref{fig:shape_results} summarizes the results across four key metrics on the 2D control task.
We observe that increasing the number of neurons generally improves task performance, especially in shallow networks.
Performance gains diminish beyond 512 neurons per layer, and models with 1024 units exhibit some instability across seeds.

Network depth has a subtler effect: moving from one to two layers provides a noticeable benefit, but adding a third layer does not yield consistent further improvement.
In some cases, deeper networks even show slower convergence or reduced robustness.

Based on these findings, we adopt a shared structure of two spiking layers with 512 neurons each for both networks in all remaining experiments.
This configuration offers strong performance with reasonable model size and training stability.

\FloatBarrier
\clearpage

\section{Training Procedure}
\label{sec:training_loop}

We adopt an iterative offline training approach to jointly optimize the prediction network $\bm{\upsilon}$ and the policy network $\bm{\pi}$.
At each iteration, a new batch of experience is collected by rolling out the current policy $\bm{\pi}$ in the environment.
These episodes are stored in a replay buffer of capacity $M$, which acts as a sampling pool for learning.
Training then proceeds in two stages: first, the prediction model is updated for $n_{\bm{\upsilon}}$ mini-batches; then, the policy model is trained for $n_{\bm{\pi}}$ mini-batches.
This process is repeated over $I = 100$ training iterations and all reported results are averaged over 3 random seeds.
The full procedure is outlined in \autoref{alg:training_loop}.

\begin{algorithm}
\caption{\textbf{Offline training of prediction and policy networks.}}
\label{alg:training_loop}
\begin{algorithmic}[1]
\State \textit{Inputs:} $\bm{\upsilon}(\hat{\bm{s}}_{t+1}|\bm{s}_t,\bm{u}_t,\bm{\theta_{\upsilon}})$, $\bm{\pi}(\bm{u}_t|\bm{s}_t,\bm{s}_t^{*},\bm{\theta_{\pi}})$, environment $(\bm{s}_{t+1}, \bm{s}^{*}_{t+1}|\bm{u}_t)$, memory capacity $M$
\State Initialize parameters $\bm{\theta_{\upsilon}}$, $\bm{\theta_{\pi}}$ and replay buffer of size $M$
\For{$I$ training iterations}
    \State Collect $E$ new episodes with $\bm{\pi}$, append to replay buffer
    \State Train $\bm{\upsilon}$ for $n_{\bm{\upsilon}}$ mini-batches \Comment{See \autoref{alg:prediction_learning}}
    \State Train $\bm{\pi}$ for $n_{\bm{\pi}}$ mini-batches \Comment{See \autoref{alg:policy_learning}}
    \State Update training schedule (e.g., decay $\alpha$, $p_{\text{tf}}$, $\sigma_u$)
\EndFor
\end{algorithmic}
\end{algorithm}

We vary two key parameters that govern the training dynamics: the memory buffer size $M$ and the number of gradient updates per iteration for each model.
Each iteration collects $E = 64$ new episodes, and both networks are trained using mini-batches of size 256.
The memory buffer is tested in three regimes: minimal memory ($M = E$), intermediate memory ($M = 20E$), and full memory ($M = IE$), where $I = 100$ is the total number of training iterations.
In parallel, we vary the number of mini-batches per iteration $n_{\bm{\upsilon}} = n_{\bm{\pi}} \in \{5, 15, 25, 35\}$.

\begin{figure}[h!]
    \centering
    \includegraphics[width=0.9\linewidth]{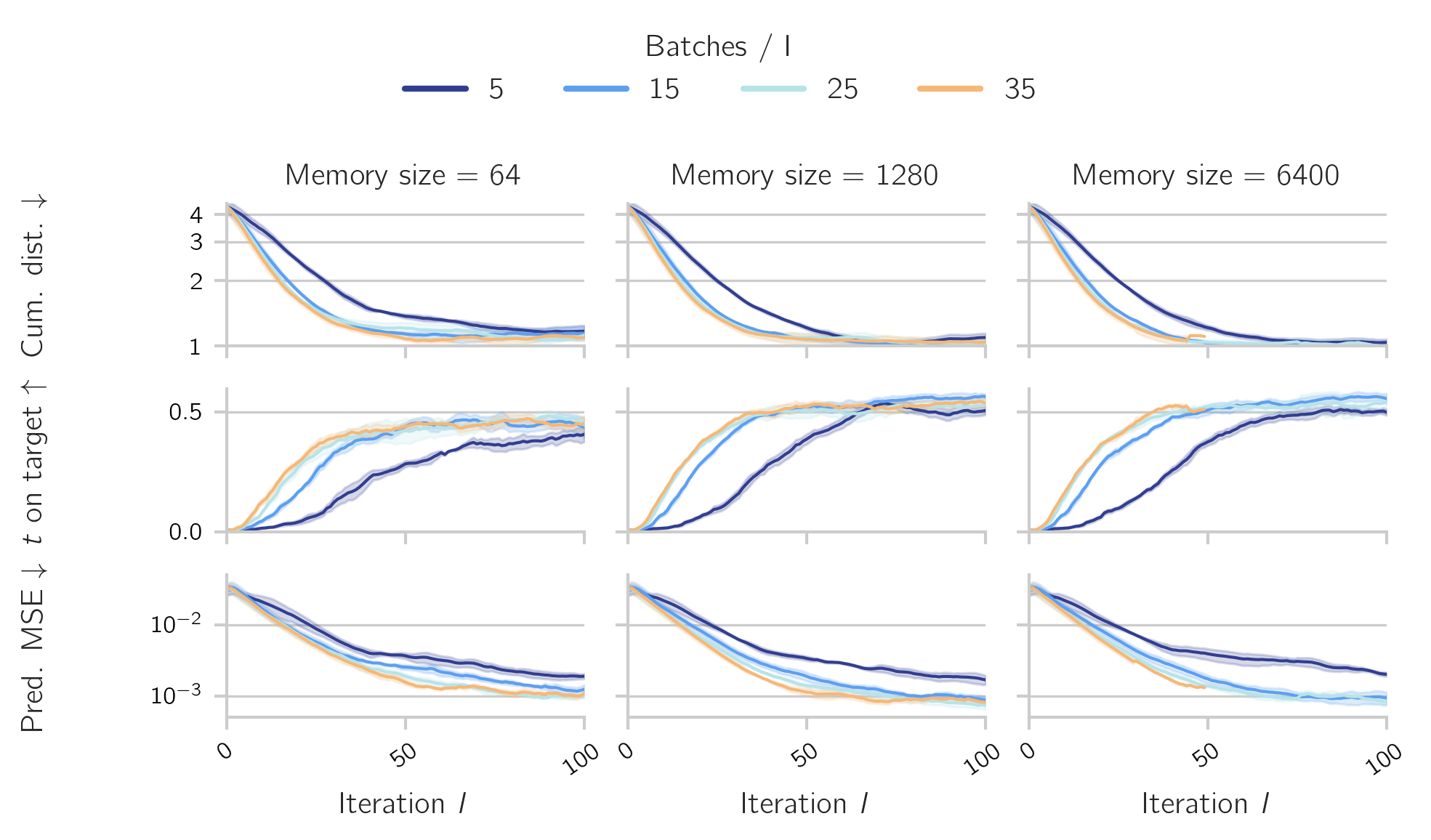}
    \caption{
    \textbf{Effect of replay buffer size and training frequency on model performance.}
    Each column corresponds to a different memory size $M$, while lines indicate the number of mini-batches per iteration.
    Metrics include cumulative distance, time-on-target, and prediction MSE.
    Moderate memory sizes (e.g., $M = 1280$) yield stable performance.
    Excessive update frequency or limited memory can lead to slower convergence or unstable behavior.
    }
    \label{fig:memory_results}
\end{figure}

\autoref{fig:memory_results} shows the resulting performance across buffer sizes and training frequencies.
We observe that larger memory buffers lead to better generalization and more stable convergence, particularly at higher training frequencies.
However, the gains diminish between $M = 1280$ and $M = 6400$, indicating that even limited memory can suffice for this relatively simple 2D control task.
Conversely, using only the most recent batch ($M = 64$) causes slower learning, particularly in the early stages, and can result in suboptimal final performance.

We also find that increasing the number of training batches per iteration improves performance up to around 25 batches, beyond which returns diminish.
This suggests that moderate reuse of experience is beneficial, but excessive replay may reduce sample diversity and hurt adaptation.

Based on these findings, all subsequent experiments use the full memory buffer ($M = IE$) and 25 mini-batches per iteration.
While memory capacity was not a bottleneck in this study, larger and more complex reinforcement learning tasks may require more careful buffer management.

Finally, we note that our training setup does not explore the regime of truly online learning, where no buffer is used and only one episode is collected and trained on per iteration.
Such a regime—where each training batch is sampled from a single episode without storage—would approximate a fully online learning setup and may require significantly different strategies for stability and plasticity.
We leave such scenarios to future work focused on continual and online adaptation in spiking control networks.

\FloatBarrier
\clearpage

\section{Prediction Network \texorpdfstring{$\bm{\upsilon}$}{upsilon} Training}
\label{sec:prediction_training}

Training the prediction network $\bm{\upsilon}$ requires special consideration due to its recurrent structure and its role in generating multi-step predictions of future robot states. At each timestep, the model receives the current robot state $\bm{s}_t$ and action $\bm{u}_t$ and predicts a state increment $\Delta \hat{\bm{s}}_t$, which is added to the previous estimate $\hat{\bm{s}}_t$ to produce $\hat{\bm{s}}_{t+1}$. The training objective is to minimize the discrepancy between these predicted and true future states over short unrolled sequences.

To enable stable gradient propagation, we adopt a two-phase rollout strategy comprising a warmup and an unroll period. During warmup, the network is driven by ground-truth observations $\bm{s}_t$ for $T_{\text{warm}}$ steps without gradient updates, allowing the internal state to converge to a stable regime. This is followed by an unroll phase of $T_{\text{unroll}}$ steps, during which predictions are made autoregressively and losses are accumulated.

To mitigate drift and improve convergence, we use teacher forcing during unrolling: at each timestep, the model either receives the true state $\bm{s}_t$ or its own previous prediction $\hat{\bm{s}}_t$ with a fixed probability $p_{\text{tf}}$. In this experiment, we test both extremes—fully enabled ($p_{\text{tf}} = 1.0$) and fully disabled ($p_{\text{tf}} = 0.0$)—as well as varying unroll lengths $T_{\text{unroll}} \in \{10, 20, 40, 80\}$. While it is possible to gradually decay $p_{\text{tf}}$ over time in a curriculum-style fashion, we found that this did not lead to further improvements for the present task.

As $\bm{\upsilon}$ predicts changes in state rather than absolute values, the target outputs are typically small. To maintain healthy spiking activity and avoid overly large predictions early in training, we initialize the weights of the readout layer with a learnable scaling factor set to 0.005. Without this scaling, gradients become unstable and the policy receives misleading updates.

The prediction error is computed using the mean squared error between predicted and ground-truth states (\autoref{eq:pred_error}), averaged across all unroll steps. Additional regularizers penalize extreme levels of spiking activity (see \autoref{sec:activity_reg}) and large weights (see \autoref{sec:weight_reg}), leading to a total loss:
\begin{equation}
\label{eq:loss_pred}
L_{\bm{\upsilon}} = e_{\hat{s}} + \lambda_{\text{low}} L_{\text{low}}(X) + \lambda_{\text{up}} L_{\text{up}}(X) + \lambda_{\text{L2}} L_{\text{L2}}.
\end{equation}

The training algorithm is summarized in \autoref{alg:prediction_learning}. The network consists of one recurrent spiking layer followed by spiking and non-spiking readout layers.

\begin{algorithm}
\caption{\textbf{Updating prediction network parameters} with warmup and teacher forcing for a single epoch.}
\label{alg:prediction_learning}
\begin{algorithmic}[1]
\State \textit{Inputs:} $\bm{\upsilon}(\hat{\bm{s}}_{t+1}|\bm{s}_t,\bm{u}_t,\bm{\theta_\upsilon})$, memory buffer, $\mathcal{L}_{\bm{\upsilon}}$, $\alpha_{\bm{\upsilon}}$, warmup steps $T_{\text{warm}}$, unroll steps $T_{\text{unroll}}$, teacher forcing prob. $p_{\text{tf}}$
\For{$n_{\bm{\upsilon}}$ mini-batches}
    \State Sample $N_{\bm{\upsilon}}$ episodes    
    \State $L \gets 0$
    \For{each episode} \Comment{Computed in parallel}
        \State Select subsequence of length $T_{\text{warm}} + T_{\text{unroll}}$
        \State Reset hidden state $h_0$
        \For{$t = 1$ to $T_{\text{warm}}$}
            \State $\hat{\bm{s}}_{t+1} \gets \bm{\upsilon}(\bm{s}_t, \bm{u}_t, \bm{\theta_\upsilon}, h_t)$
            \Comment{Warmup phase, no loss}
        \EndFor
        \For{$t = T_{\text{warm}}+1$ to $T_{\text{warm}} + T_{\text{unroll}}$}
            \State Sample $b \sim \text{Uniform}(0, 1)$
            \If{$b < p_{\text{tf}}$}
                \State $\hat{\bm{s}}_{t+1} \gets \bm{\upsilon}(\bm{s}_t, \bm{u}_t, \bm{\theta_\upsilon}, h_t)$ \Comment{Teacher forcing}
            \Else
                \State $\hat{\bm{s}}_{t+1} \gets \bm{\upsilon}(\hat{\bm{s}}_t, \bm{u}_t, \bm{\theta_\upsilon}, h_t)$ \Comment{Autoregressive input}
            \EndIf
            \State $L \gets L + \mathcal{L}_{\bm{\upsilon}}(\hat{\bm{s}}_{t+1}, \bm{s}_{t+1})$
        \EndFor
    \EndFor
    \State $\bm{\theta_\upsilon} \gets \bm{\theta_\upsilon} + \alpha_{\bm{\upsilon}} \nabla_{\theta_{\bm{\upsilon}}} \left( \frac{L}{N_{\bm{\upsilon}} T_{\text{unroll}}} \right)$ \Comment{Use optimizer for parameter update}
\EndFor
\State \Return $\bm{\theta_\upsilon}$
\end{algorithmic}
\end{algorithm}

\begin{figure}[h!]
    \centering
    \includegraphics[width=0.95\linewidth]{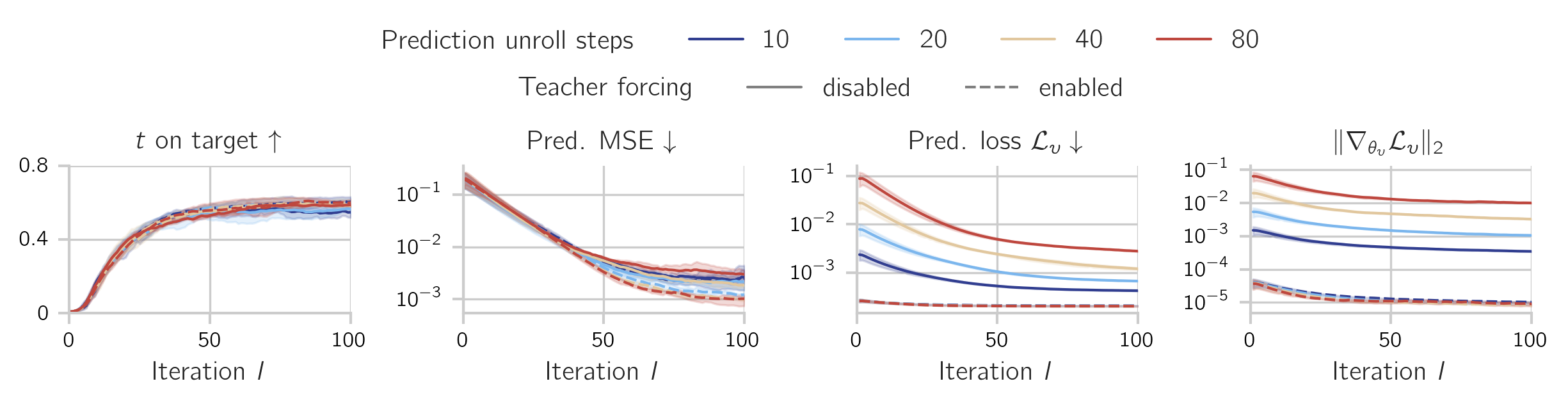}
    \caption{
    \textbf{Effect of unroll length and teacher forcing on training dynamics of the prediction model.}
    Solid lines indicate full autoregression ($p_{\text{tf}} = 0.0$), dashed lines indicate teacher forcing ($p_{\text{tf}} = 1.0$).
    Longer unrolls lead to increased loss and gradient magnitudes due to error accumulation, but task performance (time on target) remains similar across all configurations.
    Teacher forcing stabilizes gradient flow and speeds up convergence, especially at longer horizons.
    }
    \label{fig:unroll_tf_results}
\end{figure}

\autoref{fig:unroll_tf_results} shows how unroll length and teacher forcing influence model behavior. Longer unrolls without teacher forcing lead to larger prediction losses and gradient magnitudes due to error accumulation during backpropagation. 
However, these differences have only marginal effects on final prediction accuracy and virtually no impact on downstream task performance.

These findings suggest that precise long-horizon prediction is not strictly necessary for learning successful control, and that the policy network can compensate for small inaccuracies in the model's dynamics. 
Teacher forcing plays a stabilizing role, reducing gradient scale and improving convergence stability. 
Importantly, increasing the unroll horizon also leads to substantially higher computational and memory costs per batch.
Balancing predictive depth, computational efficiency, and learning stability, we select an unroll length of $T_{\text{unroll}} = 10$ and teacher forcing enabled ($p_{\text{tf}} = 1.0$) for all subsequent experiments.

\FloatBarrier
\clearpage

\section{Policy Network \texorpdfstring{$\bm{\pi}$}{pi} Training}
\label{sec:policy_training}

The policy network $\bm{\pi}$ generates control signals that guide the robot arm toward a desired end-effector position. 
Its architecture mirrors the prediction model, consisting of two spiking populations followed by a leaky integrator layer, but it does not include recurrent connections (see \autoref{sec:recurrent_connections}). 
The network receives the current state $\bm{s}_t$ and the target state $\bm{s}_t^*$ as input and outputs a continuous action $\bm{u}_t$.

Since $\bm{u}_t$ alone does not directly convey how effective the action is, we use the differentiable prediction model $\bm{\upsilon}$ to simulate the trajectory resulting from the policy's action sequence. 
By comparing predicted future states $\hat{\bm{s}}_{t+1}$ to the desired trajectory $\bm{s}^*_{t+1}$, we compute a policy loss that provides direct supervision.
This setup enables end-to-end learning through backpropagation, avoiding the complexities of reinforcement learning while maintaining signal fidelity for credit assignment.

As in the prediction model, we use a warmup and unroll procedure: during warmup, both models are driven by ground-truth states to stabilize internal dynamics. 
During the unroll, the policy outputs actions and receives feedback through the prediction model.

\paragraph{Bounded Action Output.}
To ensure valid control commands, we apply a $\tanh$ activation to the output layer, constraining actions to $[-1, 1]$. 
A learnable linear scaling precedes the $\tanh$ to ensure the pre-activations fall within the stable range of the nonlinearity. 
To further stabilize training, we add a soft regularization term:
\begin{equation}
    L_{\text{tanh}} = \sum_i \left( \max\left(0, |U_i| - 3 \right) \right)^2,
\end{equation}
where $U_i$ are the pre-activation values.

The full policy loss becomes:
\begin{equation}
\label{eq:loss_pol}
L_{\bm{\pi}} = e_{\bm{\pi}} + \lambda_{\text{tanh}} L_{\text{tanh}} + \lambda_{\text{low}} L_{\text{low}}(X) + \lambda_{\text{up}} L_{\text{up}}(X) + \lambda_{\text{L2}} L_{\text{L2}},
\end{equation}
with regularization terms defined in \autoref{sec:activity_reg} and \autoref{sec:weight_reg}.

\autoref{alg:policy_learning} outlines the full training loop.

\begin{algorithm}
\caption{\textbf{Updating policy network parameters} with warmup and unrolling for a single epoch.}
\label{alg:policy_learning}
\begin{algorithmic}[1]
\State \textit{Inputs:} $\bm{\pi}(\bm{u}_t|\bm{s}_t, \bm{s}^*_t, \bm{\theta_\pi})$, $\bm{\upsilon}(\hat{\bm{s}}_{t+1}|\bm{s}_t, \bm{u}_t, \bm{\theta_\upsilon})$, memory buffer, $\mathcal{L}_{\bm{\pi}}$, $\alpha_{\bm{\pi}}$, warmup steps $T_{\text{warm}}$, unroll steps $T_{\text{unroll}}$
\For{$n_{\bm{\pi}}$ mini-batches}
    \State Sample $N_{\bm{\pi}}$ episodes
    \State $L \gets 0$
    \For{each episode} \Comment{Computed in parallel}
        \State Select subsequence of length $T_{\text{warm}} + T_{\text{unroll}}$
        \State Reset hidden states of $\bm{\pi}$ and $\bm{\upsilon}$
        \For{$t = 1$ to $T_{\text{warm}}$}
            \State $\hat{\bm{s}}_{t+1} \gets \bm{\upsilon}(\bm{s}_t, \bm{u}_t, \bm{\theta_\upsilon})$ 
            \Comment{Warmup phase, no loss}
        \EndFor
        \For{$t = T_{\text{warm}}+1$ to $T_{\text{warm}} + T_{\text{unroll}}$}
            \State $\hat{\bm{u}}_t \gets \bm{\pi}(\hat{\bm{s}}_t, \bm{s}^*_t, \bm{\theta_\pi})$
            \State $\hat{\bm{s}}_{t+1} \gets \bm{\upsilon}(\hat{\bm{s}}_t, \hat{\bm{u}}_t, \bm{\theta_\upsilon})$
            \State $L \gets L + \mathcal{L}_{\bm{\pi}}(\hat{\bm{s}}_{t+1}, \bm{s}^*_{t+1})$
        \EndFor
    \EndFor
    \State $\bm{\theta_\pi} \gets \bm{\theta_\pi} + \alpha_{\bm{\pi}} \nabla_{\theta_{\bm{\pi}}} \left( \frac{L}{N_{\bm{\pi}} T_{\text{unroll}}} \right)$
    \Comment{Use optimizer for parameter update}
\EndFor
\State \Return $\bm{\theta_\pi}$
\end{algorithmic}
\end{algorithm}

We now investigate how the length of the unroll window $T_{\text{unroll}}$ affects policy training. 
We test values of $T_{\text{unroll}} = \{5, 10, 20, 40, 80\}$ using three random seeds each.

\begin{figure}[h!]
    \centering
    \includegraphics[width=0.95\linewidth]{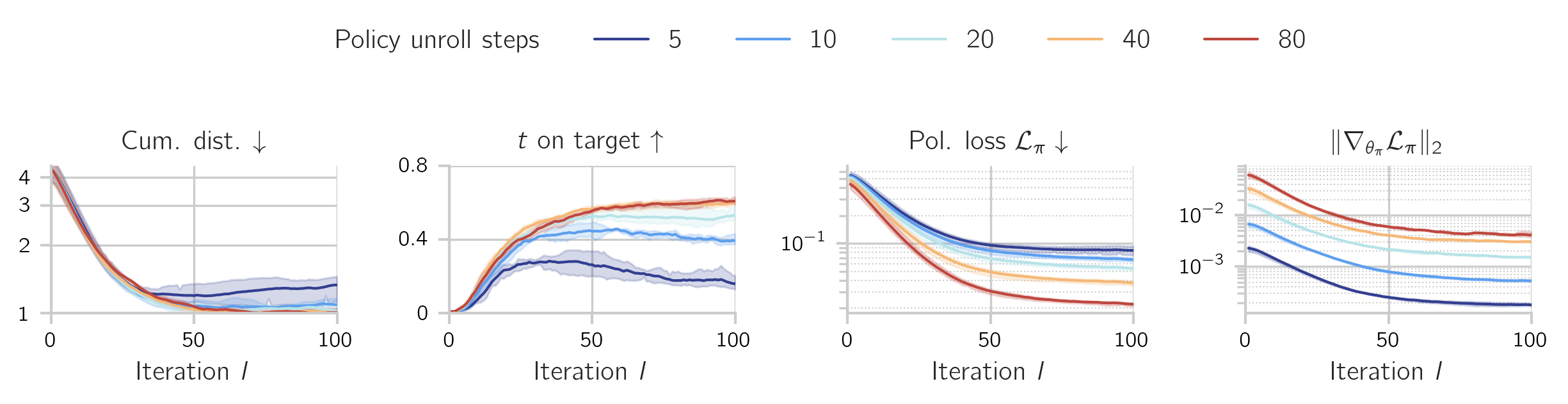}
    \caption{
    \textbf{Effect of policy unroll length on training dynamics and task performance.}
    Longer unrolls improve time on target and lower policy loss by providing richer gradient signals.
    However, gradient magnitudes increase with unroll length, indicating a growing risk of instability.
    Gains saturate at 40 steps, which is used in all subsequent experiments.
    }
    \label{fig:unroll_policy_results}
\end{figure}

\autoref{fig:unroll_policy_results} reveals clear trends. Task performance—measured as time spent on target—improves sharply with longer unroll windows and saturates around 40 steps. 
While even short unrolls reduce cumulative distance, they fail to produce stable control behaviors. 
Meanwhile, policy loss consistently decreases with longer horizons, reflecting greater opportunity for gradient-based corrections. 
However, this comes at the cost of growing gradient magnitudes, which suggests an increased risk of exploding gradients at long horizons. 

Interestingly, this pattern differs from the prediction model, where longer unrolls increased loss due to compounding error. 
Here, the policy benefits from longer feedback horizons, provided the prediction model remains stable.

For all subsequent experiments, we adopt $T_{\text{unroll}} = 40$ for policy training, which balances performance gains and computational efficiency.

\FloatBarrier
\clearpage

\section{Learnable Time Constants \texorpdfstring{$\tau$}{tau}}
\label{sec:learnable_timescales}

In all preceding sections, the neuronal time constants $\tau_{\text{mem}}$ and $\tau_{\text{syn}}$ were treated as fixed hyperparameters. 
However, in biological neurons and many real-world control settings, the optimal temporal integration windows may vary across tasks or layers. 
Allowing these parameters to be learned end-to-end gives the network flexibility to adapt its internal dynamics to the temporal structure of the task, potentially improving sample efficiency and robustness.

A key technical consideration is ensuring that learned time constants remain positive. 
To this end, we parameterize each time constant as $\tau = \exp(\tau')$, where $\tau'$ is an unconstrained real-valued parameter. 
This exponential mapping has several benefits: it guarantees positivity, enables multiplicative rather than additive updates (which are scale-invariant), and results in smooth gradients with respect to both small and large values of $\tau$.
These properties promote stable and interpretable learning behavior, and avoid pathological dynamics such as negative or vanishing time constants.

While one could alternatively parameterize and learn the decay factor $\beta = \exp(-\Delta t / \tau)$, we found learning in $\log \tau$ space to be more stable and easier to interpret. 
Preliminary experiments (not shown) also confirmed this choice resulted in more consistent convergence.

This learning mechanism can be applied per layer or per neuron. 
In the \textbf{per-neuron} configuration, each unit receives its own learned time constant, allowing fine-grained temporal specialization. 
In contrast, the \textbf{layer-wise} variant uses a single shared $\tau$ per layer, which reduces parameter count and computational cost.
In our experiments, we found that per-neuron learning yielded slightly better results without significant overhead, so this configuration is used throughout.

Here we evaluate whether learning time constants improves task performance, and whether it can compensate for suboptimal initialization. 
We focus on learning $\tau_{\text{mem}}$, while keeping $\tau_{\text{syn}}$ fixed at $2\,\text{ms}$. 
Other time constants (such as $\tau_{\text{ada}}$ in \autoref{sec:adaptive_lif}) can be learned using the same mechanism.

We consider two initializations of $\tau_{\text{mem}}$, $10\,\text{ms}$ (optimal, based on prior results) and $20\,\text{ms}$ (suboptimal), and three learning rates for $\tau$:
$\alpha_{\tau} \in \{0.0, 0.001, 0.01\}$, where $\alpha_{\tau}=0$ disables learning, and $\alpha_{\tau}=0.01$ corresponds to “fast learning”, i.e., ten times the learning rate used for other model parameters.

\begin{figure}[h!]
    \centering
    \includegraphics[width=0.9\linewidth]{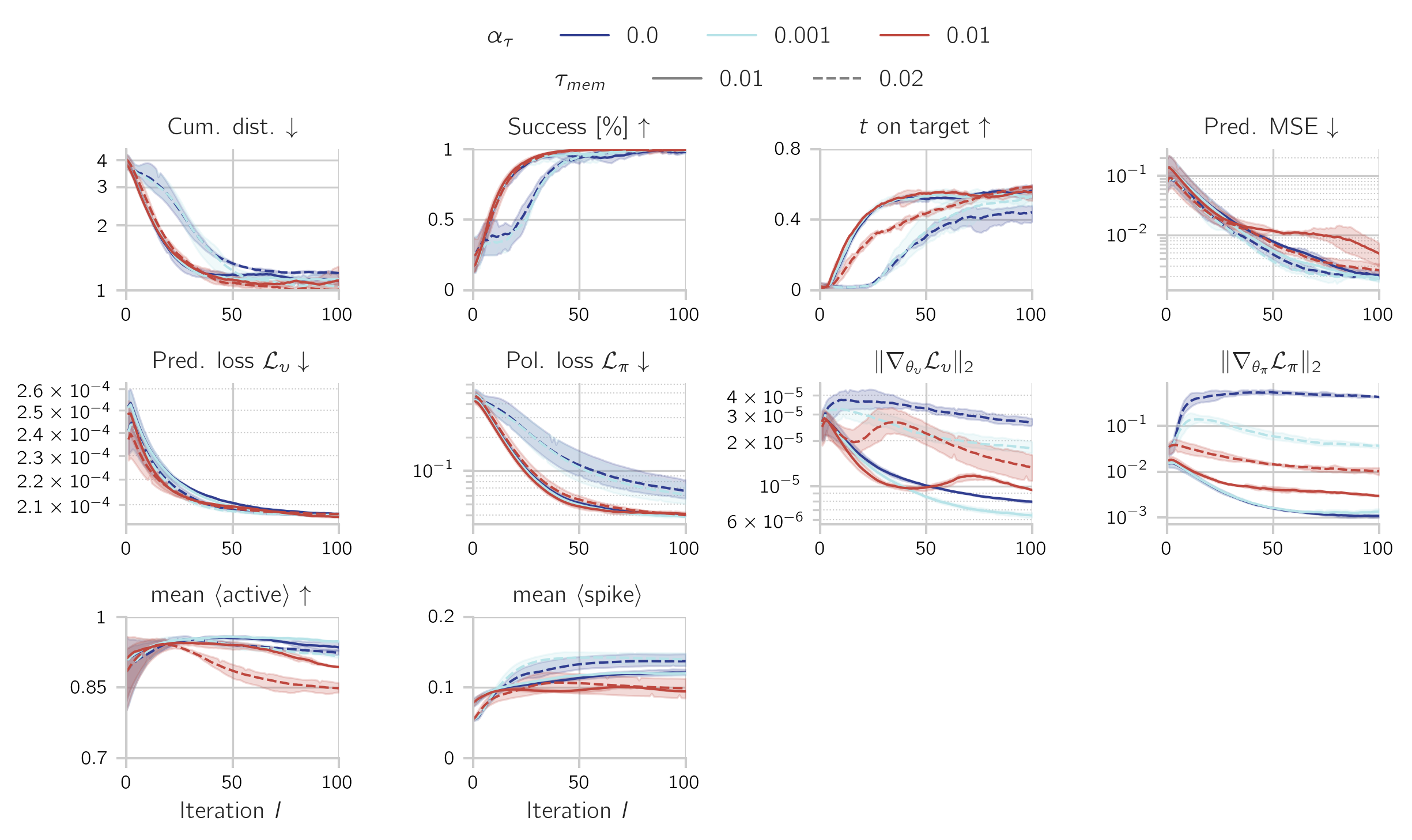}
    \includegraphics[width=0.7\linewidth]{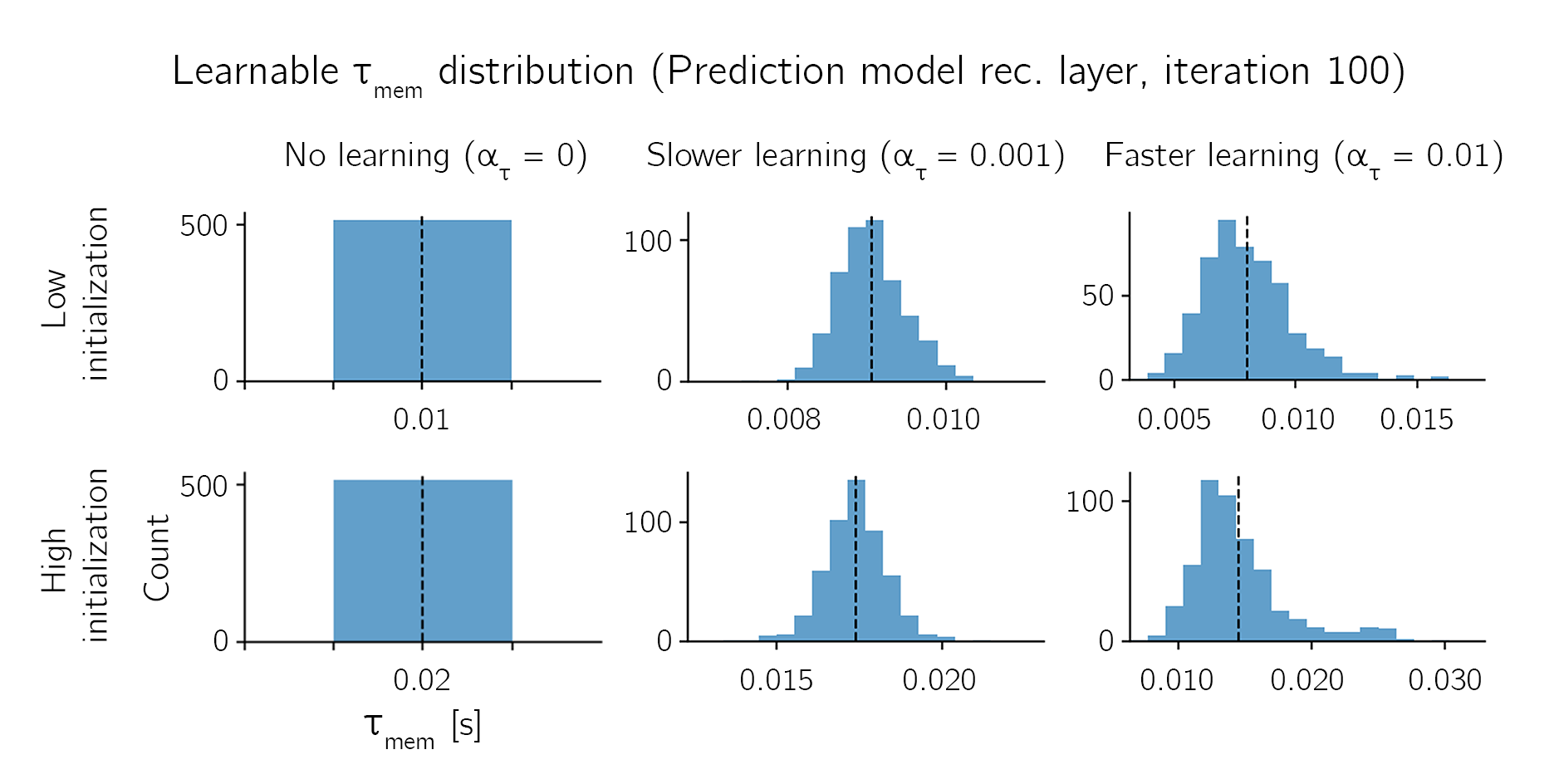}
    \caption{
    \textbf{Effect of learnable time constants on learning and neuron dynamics.}  
    \textbf{Top:} Learning curves and activity metrics for $\tau_{\text{mem}}$ initialization at $10\,\text{ms}$ (dashed) or $20\,\text{ms}$ (solid), with varying learning rates $\alpha_{\tau}$.  
    Fast learning ($\alpha_{\tau} = 0.01$) recovers performance even from poor initialization.  
    \textbf{Bottom:} Histogram of learned $\tau_{\text{mem}}$ values after training.  
    Learning shifts the distribution toward smaller, task-adapted values regardless of initialization.
    }
    \label{fig:learnable_tau_results}
\end{figure}

\autoref{fig:learnable_tau_results} (top) shows that with a good initialization ($10\,\text{ms}$), learning $\tau_{\text{mem}}$ makes little difference: both static and adaptive models perform well. 
However, when initialized at $20\,\text{ms}$, models without $\tau$ learning suffer in performance.
While slow learning ($\alpha_{\tau} = 0.001$) offers only mild improvements, fast learning ($\alpha_{\tau} = 0.01$) fully recovers performance, matching the results of the optimal initialization. 
This highlights the benefit of allowing models to escape suboptimal initial values by tuning their intrinsic timescales.

To understand how $\tau_{\text{mem}}$ evolves during training, \autoref{fig:learnable_tau_results} (bottom) shows the distribution of final time constants in the prediction model's recurrent layer.
When learning is enabled, the values shift consistently toward shorter integration windows, suggesting that the task favors fast neuronal responses.
The learned distributions remain unimodal but show slight leftward skew, with final means clearly below their respective initializations.

Based on these findings, we use per-neuron learnable $\tau_{\text{mem}}$ and $\tau_{\text{syn}}$ with fast learning rate $\alpha_{\tau} = 0.01$ in all subsequent experiments.

\FloatBarrier
\clearpage

\section{Adaptive Leaky Integrate-and-Fire Neuron Model}
\label{sec:adaptive_lif}

Spiking neurons often exhibit adaptive behavior, where their firing threshold increases temporarily after a spike. 
This mechanism reduces immediate re-firing and promotes sparse activity, improving network robustness and training dynamics. 
In this section, we study an adaptive variant of the LIF model (ALIF) that incorporates both fast and slow modulation of the spiking threshold to stabilize activity and promote efficient learning. 
We implement a discrete-time variant of ALIF where the baseline threshold provides slow homeostatic recovery, while the adaptive component provides spike-triggered suppression. 
Together, these mechanisms stabilize network activity and extend the effective eligibility traces available for gradient learning.

Specifically, the instantaneous firing threshold $\vartheta_t$ consists of two components:
\begin{itemize}
    \item a baseline component $\vartheta_{b,t}$ that gradually decays back toward a nominal value $\vartheta_0$ in the absence of spikes, and
    \item an adaptive component $\vartheta_{a,t}$ that increases with recent spiking activity and decays over time.
\end{itemize}

This formulation represents one of several possible ALIF variants. 
Other models introduce additional mechanisms such as explicit refractory periods, additive threshold increments instead of multiplicative scaling, or combined current- and threshold-based adaptation. 
We adopt this particular formulation because it balances simplicity with effectiveness for gradient-based training, and extends the eligibility traces of spiking neurons in a way that benefits backpropagation-through-time.
The total threshold is computed as:
\begin{equation}
\label{eq:adathresh_main}
\vartheta_t = \vartheta_{b,t} + \xi_{\vartheta}\,\vartheta_{a,t},
\end{equation}
where $\xi_{\vartheta}$ controls the strength of spike-triggered adaptation.

Spiking dynamics follow the standard discrete-time LIF updates (cf. Eqs.~\ref{eq:lif_i}--\ref{eq:lif_u}), but spike occurrence and membrane reset are now governed by the adaptive threshold:
\begin{equation}
\label{eq:alif_reset}
U_t = \tilde U_t - \vartheta_t S_t,
\end{equation}
where $\tilde U_t$ is the pre-reset membrane potential and $S_t$ the spike output.

The baseline and adaptation traces evolve according to:
\begin{align}
\label{eq:theta_b_update}
\vartheta_{b,t} &= \vartheta_{b,t-1} - \Delta \vartheta + \bigl[\vartheta_0 - \vartheta_{b,t-1}\bigr]\,S_t, \\
\label{eq:theta_a_update}
\vartheta_{a,t} &= \beta_{\text{ada}}\,\vartheta_{a,t-1} + S_t,
\end{align}
with $\beta_{\text{ada}} = \exp(-\Delta t / \tau_{\text{ada}})$. 
The scalar $\Delta \vartheta$ defines the slope of the baseline decay and allows silent neurons to slowly become excitable again over time. 
Our ALIF procedure is shown in detail in \autoref{alg:alif}.

\begin{algorithm}
\caption{Discrete-time update procedure for ALIF neurons.
Each population updates its synaptic current $I_t$, membrane potential $U_t$, and dynamic thresholds $\vartheta_{b,t}, \vartheta_{a,t}$ according to exponential decay dynamics.
Spikes $S_t$ are generated when the voltage exceeds the adaptive threshold $\vartheta_t$, after which a subtractive reset is applied.
All updates are differentiable via surrogate gradients during training.}
\label{alg:alif}
\begin{algorithmic}
\For{all populations $p$ in network}
    \State $U_0 \gets U_{\text{rest}},\; I_0 \gets 0,\; \vartheta_{b,0} \gets \vartheta_0,\; \vartheta_{a,0} \gets 0,\; S_0 \gets 0$
    \Comment{Initialize population states}
\EndFor

\For{$t \in [1 \dots T]$}
    \For{all populations $p$ in network}
        \State $x_t \gets [\,S_{t-1,i}\,\text{for all populations }i \text{ projecting to }p]$
        \Comment{Collect spike outputs from the previous time step}
        
        \State $I_t \gets \beta_{\text{syn}} I_{t-1} + W x_t + I_{\text{inj},t}$
        \Comment{Update synaptic current (Eq.\,\ref{eq:lif_i})}
        
        \State $\tilde U_t \gets \beta_{\text{mem}} U_{t-1} + (1-\beta_{\text{mem}})\,I_t$
        \Comment{Compute membrane potential (Eq.\,\ref{eq:lif_u})}
                
        \State $S_t \gets \Theta(\tilde U_t - \vartheta_t)$
        \Comment{Check if neuron fires}
        
        \State $U_t \gets \tilde U_t - \vartheta_t S_t$
        \Comment{Subtractive reset (Eq.\,\ref{eq:alif_reset})}

        \State $\vartheta_{b,t} \gets \vartheta_{b,t-1} - \Delta \vartheta + \bigl[\vartheta_0 - \vartheta_{b,t-1}\bigr]\,S_t$
        \Comment{Baseline threshold update (Eq.\,\ref{eq:theta_b_update})}
        
        \State $\vartheta_{a,t} \gets \beta_{\text{ada}} \vartheta_{a,t-1} + S_t$
        \Comment{Spike-triggered adaptation (Eq.\,\ref{eq:theta_a_update})}
        
        \State $\vartheta_t \gets \vartheta_{b,t} + \xi_{\vartheta}\,\vartheta_{a,t}$
        \Comment{Updated threshold (Eq.\,\ref{eq:adathresh_main})}
    \EndFor
\EndFor
\end{algorithmic}
\end{algorithm}

In the following subsections, we examine how each mechanism influences the behavior of single ALIF neurons and their firing responses to tonic and noisy inputs, as well as their impact on learning the 2D control task. 
To allow both $\vartheta_a$ and $\vartheta_b$ to contribute meaningfully, in the control experiments we use an unroll window $T_{\text{unroll}} = 40$ environment steps (= 280 SNN steps) for both the prediction and the policy model.

\subsection{Effect of Threshold Decay}

We first isolate the effect of the baseline decay rate $\Delta \vartheta$, which causes the baseline threshold $\vartheta_{b,t}$ to decrease over time in the absence of spikes.  
This enables otherwise inactive neurons to gradually recover excitability, acting as a simple homeostatic mechanism.  
To evaluate this effect in isolation, we fix $\xi_{\vartheta} = 0$ (disabling spike-triggered adaptation) and vary $\Delta \vartheta$ across trials.

\begin{figure}[h!]
    \centering
    \includegraphics[width=0.6\linewidth]{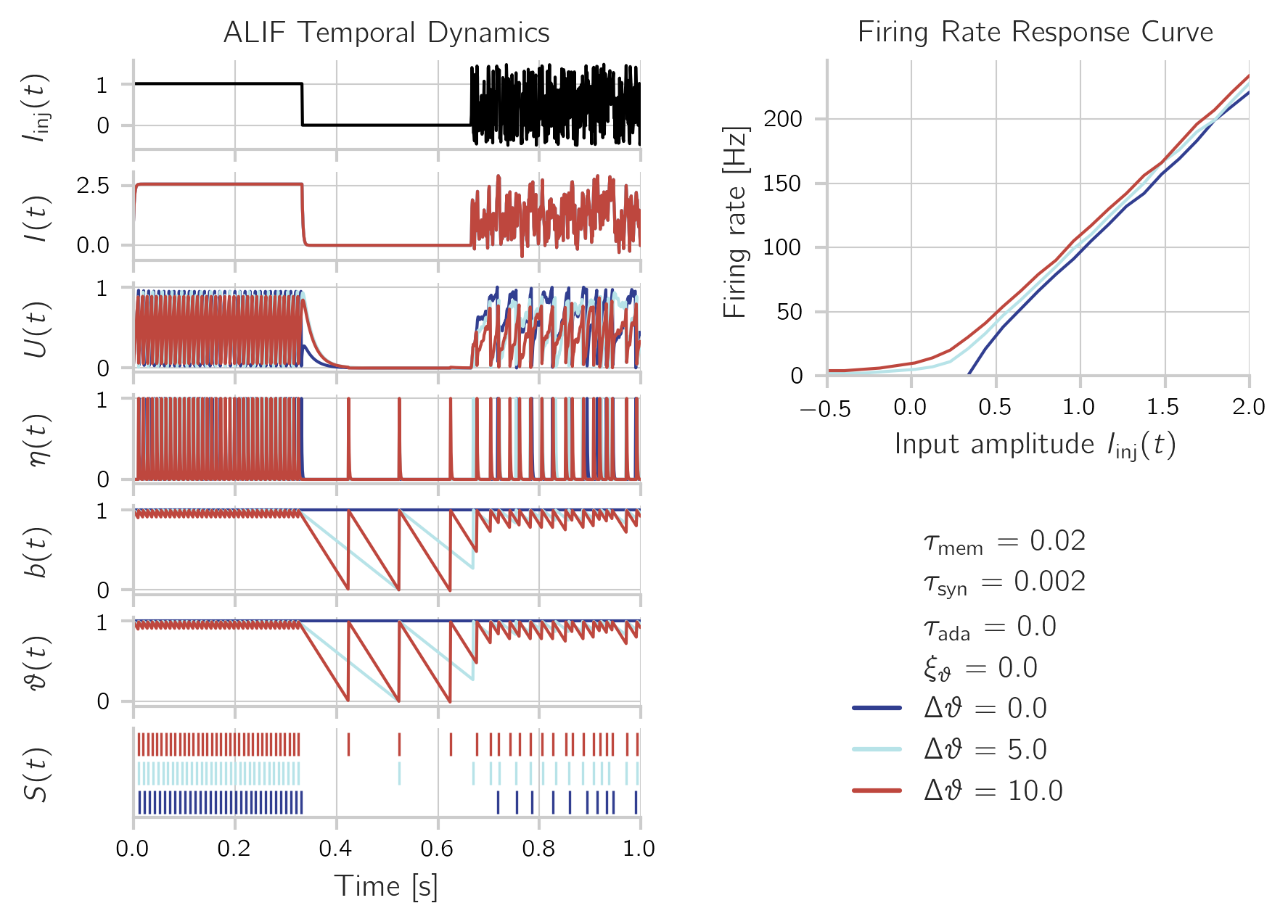}
    \caption{
    \textbf{Effect of baseline decay rate $\Delta \vartheta$ on single-neuron firing dynamics.}  
    Larger values of $\Delta \vartheta$ shorten the latency before silent neurons resume spiking, enabling recovery even under low or inhibitory input.
    }
    \label{fig:alif_decay_only}
\end{figure}

\autoref{fig:alif_decay_only} illustrates how neurons with higher decay rates recover from suppressed states and resume firing more quickly, even under constant or weakly negative current input.  
This leads to spontaneous reactivation and helps maintain population-level activity without relying on external noise or manual regularization.

\begin{figure}[h!]
    \centering
    \includegraphics[width=0.9\linewidth]{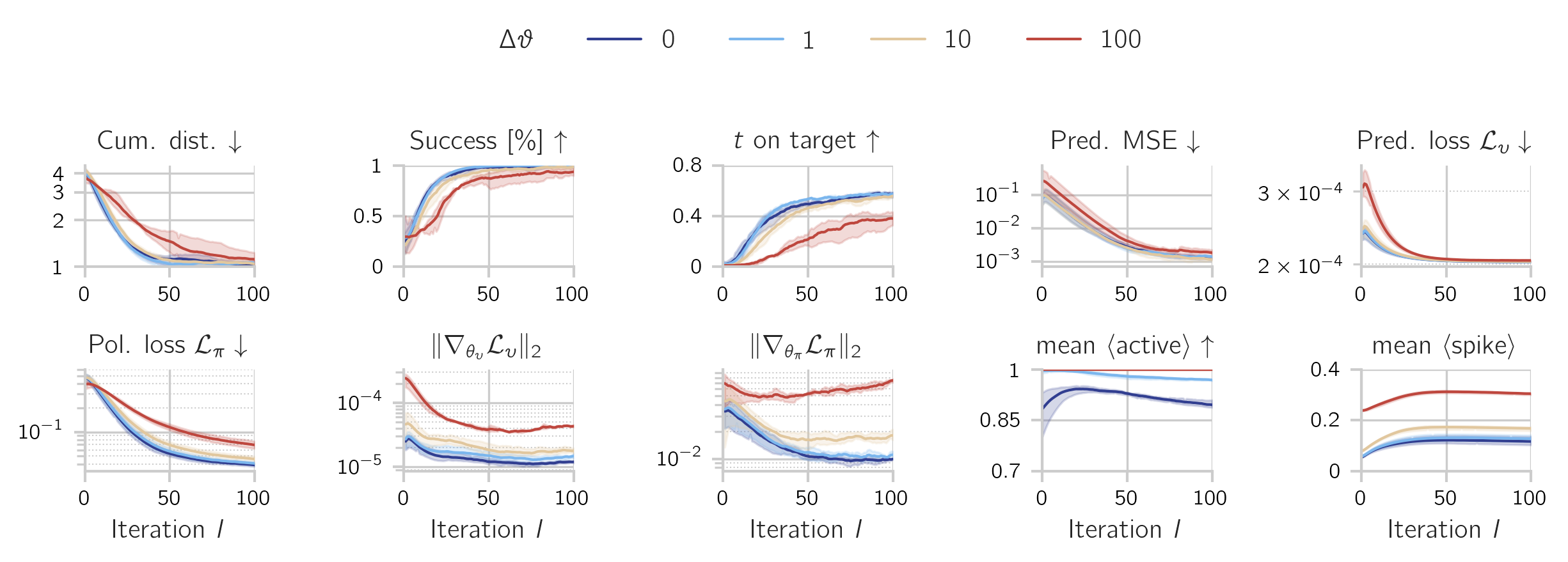}
    \caption{
    \textbf{Effect of baseline decay rate $\Delta \vartheta$ on control performance.}  
    Moderate decay values improve learning stability and neuron utilization without harming performance, while excessive decay (e.g., $\Delta \vartheta = 100$) causes overactivation, degraded accuracy, and slower convergence.
    }
    \label{fig:alif_decay_results}
\end{figure}

We next assess the effect of baseline decay on learning performance in the 2D control task.  
\autoref{fig:alif_decay_results} shows that intermediate values of $\Delta \vartheta$ improve policy learning by keeping more neurons active throughout training.  
Across runs, we find that increasing $\Delta \vartheta$ consistently raises the mean number of active and spiking neurons (bottom row), thereby reducing neuron silence.  
This also leads to larger gradient magnitudes in both the prediction and policy models, especially at $\Delta \vartheta = 100$, where the spiking activity becomes dense and overactive.  
While such aggressive decay eliminates silent units entirely, it impairs control behavior and slows convergence, likely due to the disruptive effect of excessive noise during training.  
By contrast, moderate decay rates (e.g., $\Delta \vartheta = 10$) yield stable and fast learning, suggesting a beneficial trade-off between activation and precision.  
The setting $\Delta \vartheta = 1$ shows only minimal differences compared to no decay ($\Delta \vartheta = 0$), and is therefore omitted from subsequent comparisons.  
We continue with $\Delta \vartheta = 0$ and $10$ as representative cases in the next sections.

While threshold decay successfully increases the number of active neurons, this does not necessarily imply that all units contribute meaningfully to information encoding or task performance.  
A deeper analysis of neuron selectivity and functional contribution would be needed to assess the representational benefits of this mechanism, which is beyond the scope of the present work.  
Such analyses could be particularly valuable when combined with pruning strategies that remove persistently uninformative units, offering a promising direction for future research.

\FloatBarrier

\subsection{Effect of Threshold Adaptation}

We now study the effect of the adaptive threshold trace $\vartheta_{a,t}$ in the absence of baseline decay.  
To isolate this mechanism, we fix $\Delta \vartheta = 0$ and explore different combinations of adaptation time constant $\tau_{\text{ada}}$ and scaling factor $\xi_{\vartheta}$.  
These parameters control how strongly recent spike history suppresses neuronal excitability. Note that based on the findings in \autoref{sec:learnable_timescales}, all time constants, including $\tau_{\text{ada}}$, are learned parameters. 

\begin{figure}[h!]
    \centering
    \includegraphics[width=0.6\linewidth]{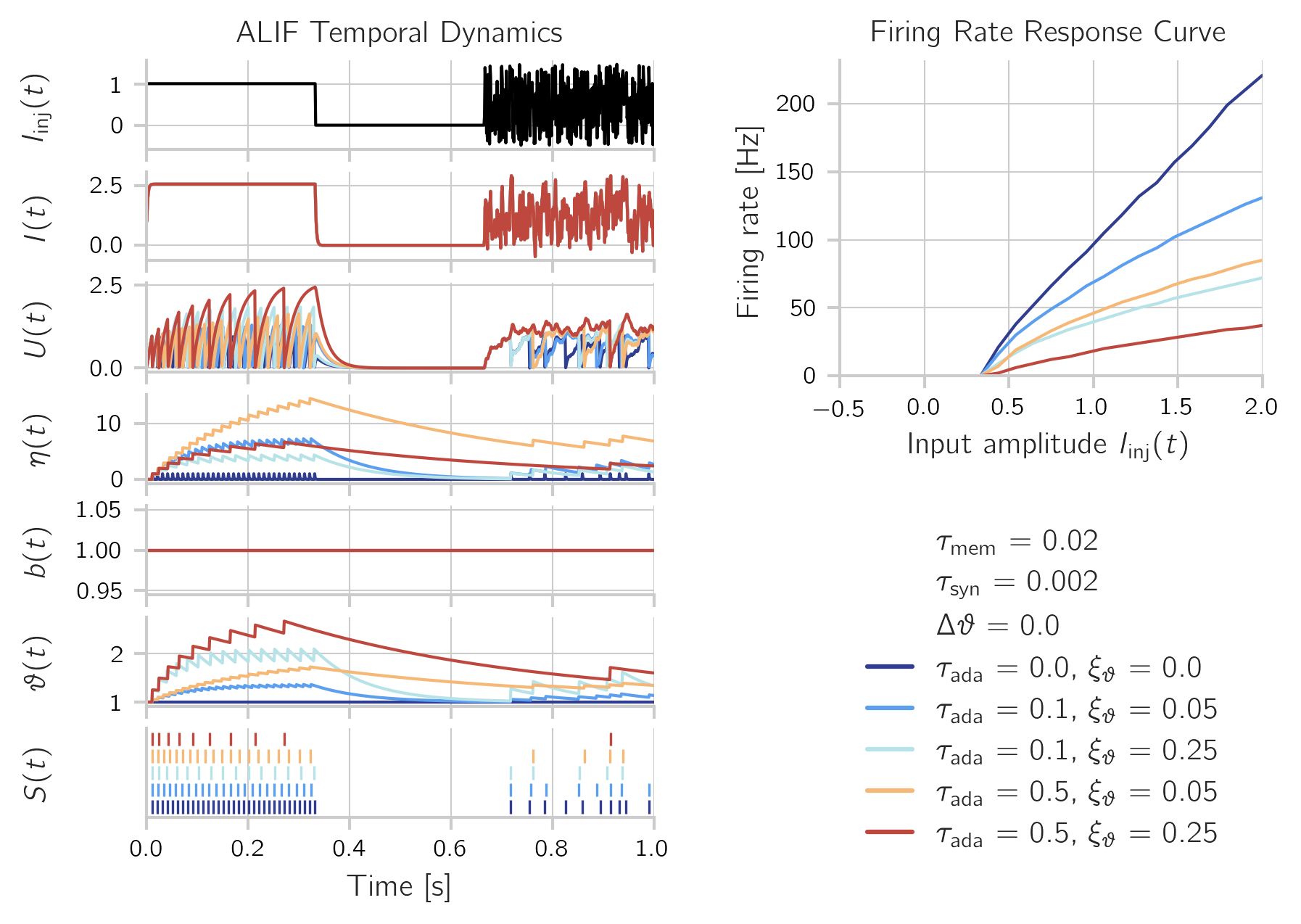}
    \caption{
    \textbf{Effect of spike-triggered threshold adaptation on single-neuron firing.}  
    Left: membrane and threshold traces under step + noise input.  
    Right: steady-state firing rates across input amplitudes.  
    Stronger adaptation (higher $\xi_{\vartheta}$ or longer $\tau_{\text{ada}}$) leads to more pronounced firing suppression.  
    While different settings may result in similar steady-state output (e.g., light blue vs.\ orange lines), their temporal responses differ significantly.
    }
    \label{fig:alif_adaptation_only}
\end{figure}

\autoref{fig:alif_adaptation_only} shows that spike-triggered adaptation effectively reduces firing rates under sustained stimulation.  
The suppression persists beyond the period of active input, illustrating a form of activity-dependent memory.  
Interestingly, different parameter combinations can produce similar steady-state firing rates but yield different temporal responses.  
For example, $\tau_{\text{ada}} = 0.1$, $\xi_{\vartheta} = 0.25$ (light blue) and $\tau_{\text{ada}} = 0.5$, $\xi_{\vartheta} = 0.05$ (orange) both suppress spiking to comparable levels, but with distinct adaptation dynamics that will influence gradient flow during training.

\begin{figure}[h!]
    \centering
    \includegraphics[width=0.9\linewidth]{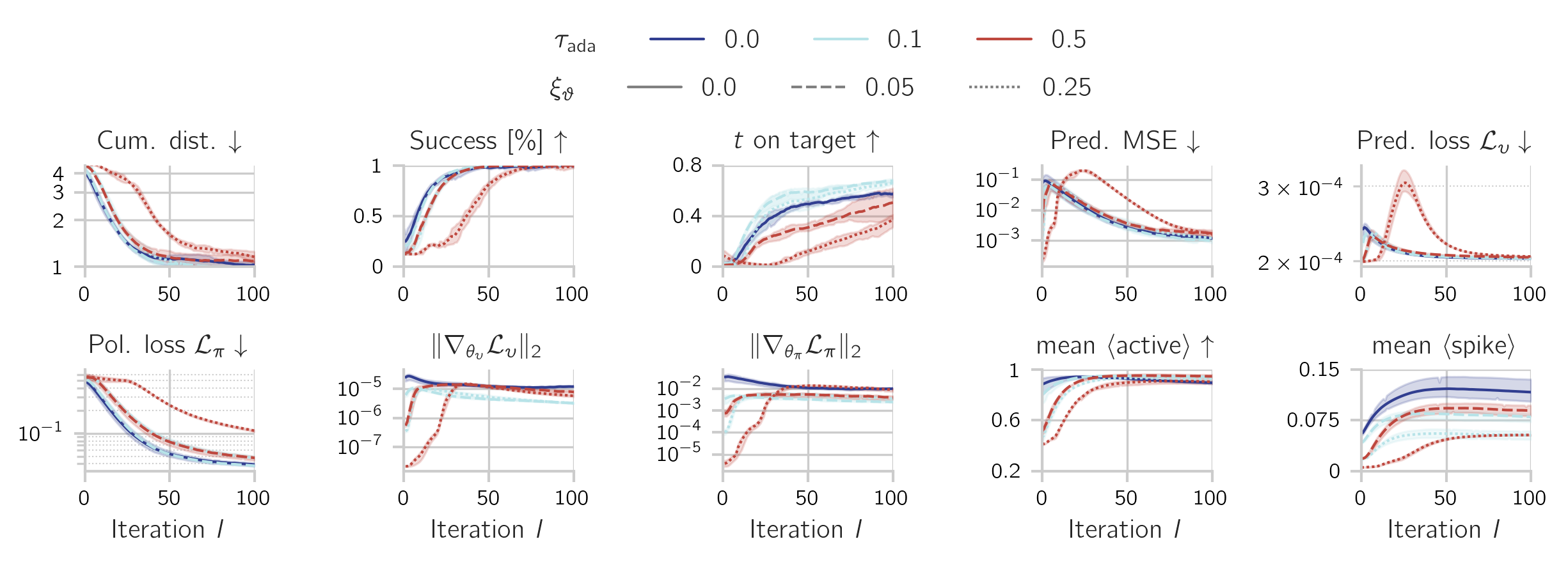}
    \caption{
    \textbf{Effect of threshold adaptation on task performance and activity.}  
    Moderate adaptation improves neuron utilization and accelerates convergence.  
    However, overly strong suppression (e.g., $\tau_{\text{ada}} = 0.5$, $\xi_{\vartheta} = 0.25$) can delay learning or reduce precision, despite achieving high final success rates.
    }
    \label{fig:alif_adaptation_results}
\end{figure}

When applied in the 2D control task, threshold adaptation leads to a reduction in overall spike counts, as expected (see \autoref{fig:alif_adaptation_results}).  
Although some neurons are initially suppressed, population-wide activity tends to recover during training.  
Configurations with $\tau_{\text{ada}} = 0.1$ lead to faster convergence and improved task performance relative to the baseline.  
Interestingly, the aforementioned parameter settings that induce similar firing rates result in markedly different learning trajectories.  
This highlights the sensitivity of adaptation-based dynamics to temporal structure and underscores the importance of careful tuning.
Examining the distribution of the $\tau_{\text{ada}}$ parameter at the end of learning shows a clear preference toward smaller values (data not shown).
Going forward, we adopt $\tau_{\text{ada}} = 0.1$ in all experiments using threshold adaptation.

\FloatBarrier

\subsection{Combined Behavior}

Finally, we evaluate the full ALIF model by combining spike-triggered adaptation with baseline decay.  
This formulation enables the suppression of overly active neurons while also reactivating silent ones, achieving a dynamic balance between stability and responsiveness.  
We fix the baseline decay rate to $\Delta \vartheta = 10$ and vary the adaptation parameters $\tau_{\text{ada}}$ and $\xi_{\vartheta}$ as in the previous section.

\begin{figure}[h!]
    \centering
    \includegraphics[width=0.6\linewidth]{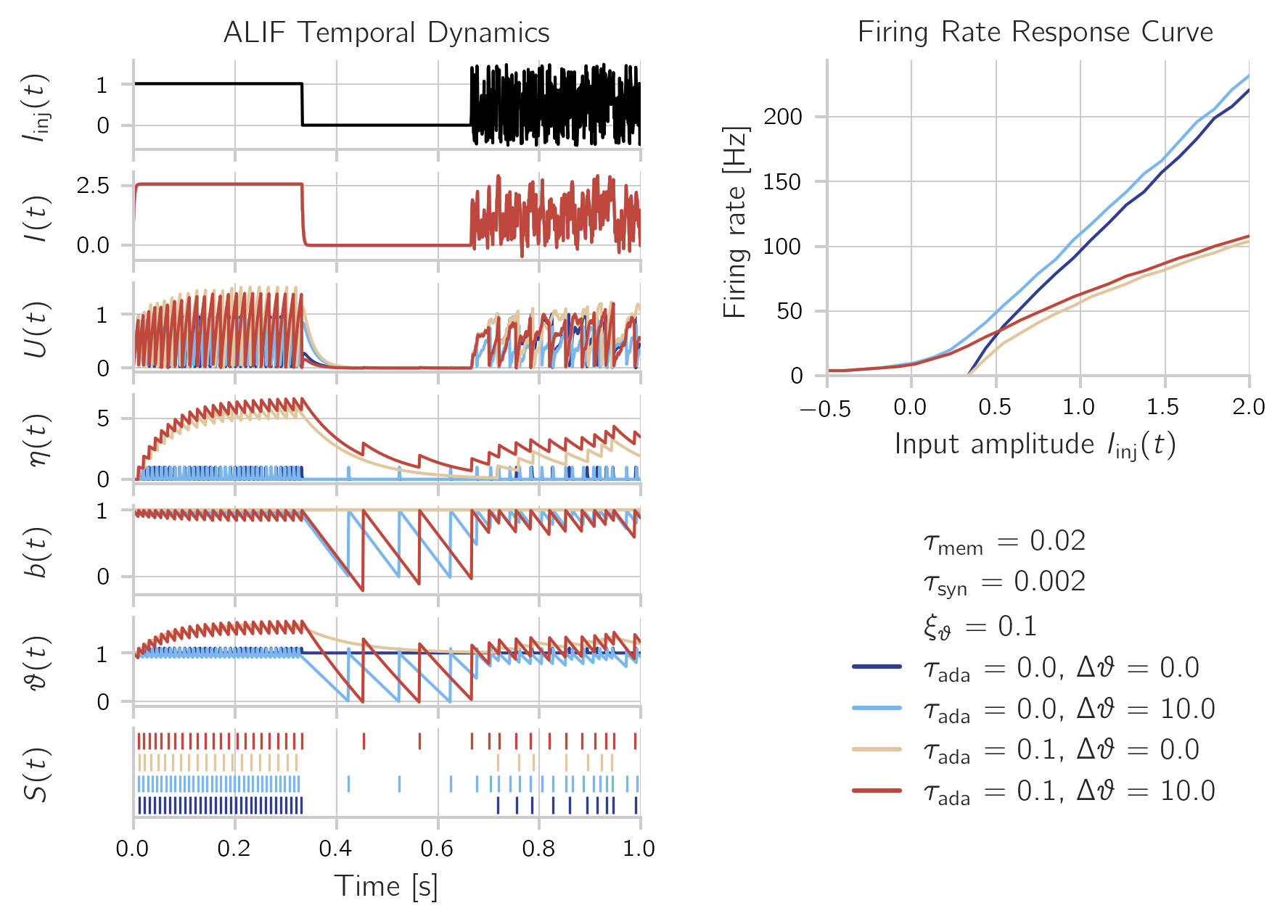}
    \caption{
    \textbf{Combined effect of threshold adaptation and baseline decay on single-neuron firing.}  
    The network simultaneously suppresses overactive neurons and reactivates silent ones, resulting in balanced, selective, and robust firing behavior across input amplitudes.
    }
    \label{fig:alif_combined}
\end{figure}

\autoref{fig:alif_combined} shows how this combination yields selective and robust firing responses under step and noisy current input.  
Spiking activity remains sparse, while silent neurons recover naturally without requiring external noise or activity penalties.

\begin{figure}[h!]
    \centering
    \includegraphics[width=0.9\linewidth]{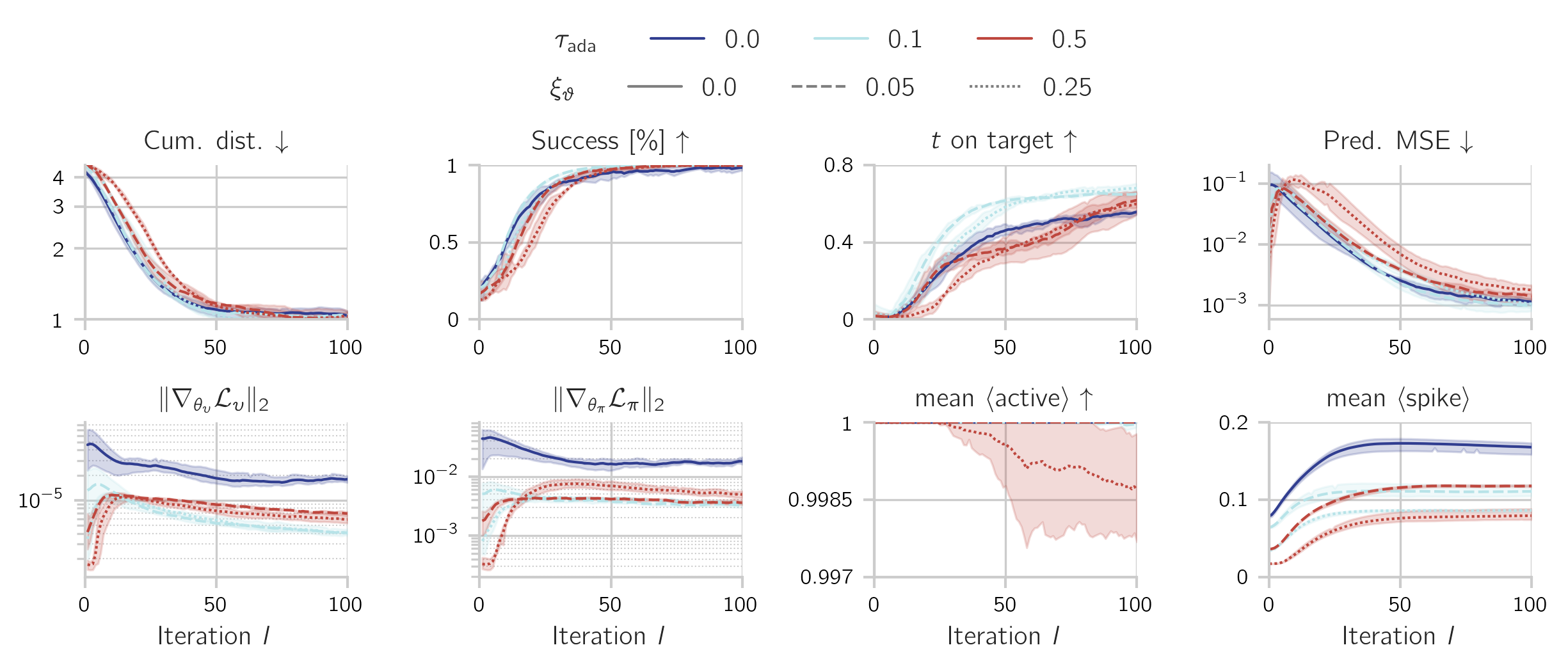}
    \caption{
    \textbf{Control task performance using combined ALIF dynamics.}  
    Baseline decay ($\Delta \vartheta = 10$) amplifies the benefits of spike-triggered adaptation.  
    All neurons remain active, while overall spiking is reduced and learning becomes more stable and efficient.
    }
    \label{fig:alif_combined_results}
\end{figure}

In the 2D control task, this combination proves highly effective.  
\autoref{fig:alif_combined_results} shows that all configurations yield strong task performance, with faster convergence and improved learning stability compared to using adaptation alone.  
Neurons remain active throughout training, and overall spiking is substantially reduced.  
While both $\xi_{\vartheta}$ settings perform similarly at convergence, smaller values slightly accelerate early learning.  
We also confirm that $\tau_{\text{ada}} = 0.1$ offers better performance than $\tau_{\text{ada}} = 0.5$ across metrics.  
Based on these results, we continue with the configuration $\Delta \vartheta = 10$, $\tau_{\text{ada}} = 0.1$, and $\xi_{\vartheta} = 0.1$ in subsequent experiments.

\FloatBarrier
\clearpage

\section{Weight Regularization}
\label{sec:weight_reg}

L2 weight decay is a widely used technique to prevent overfitting by penalizing large weights. 
In this experiment, we investigate whether applying L2 regularization to all learnable weights improves performance or stability in our spiking control setup. 
We test values of $\lambda_{\text{L2}} \in \{0.0,\; 10^{-4},\; 10^{-3},\; 10^{-2},\; 10^{-1}\}$, applied uniformly to both the prediction and policy networks.

\begin{figure}[h!]
    \centering
    \includegraphics[width=0.9\linewidth]{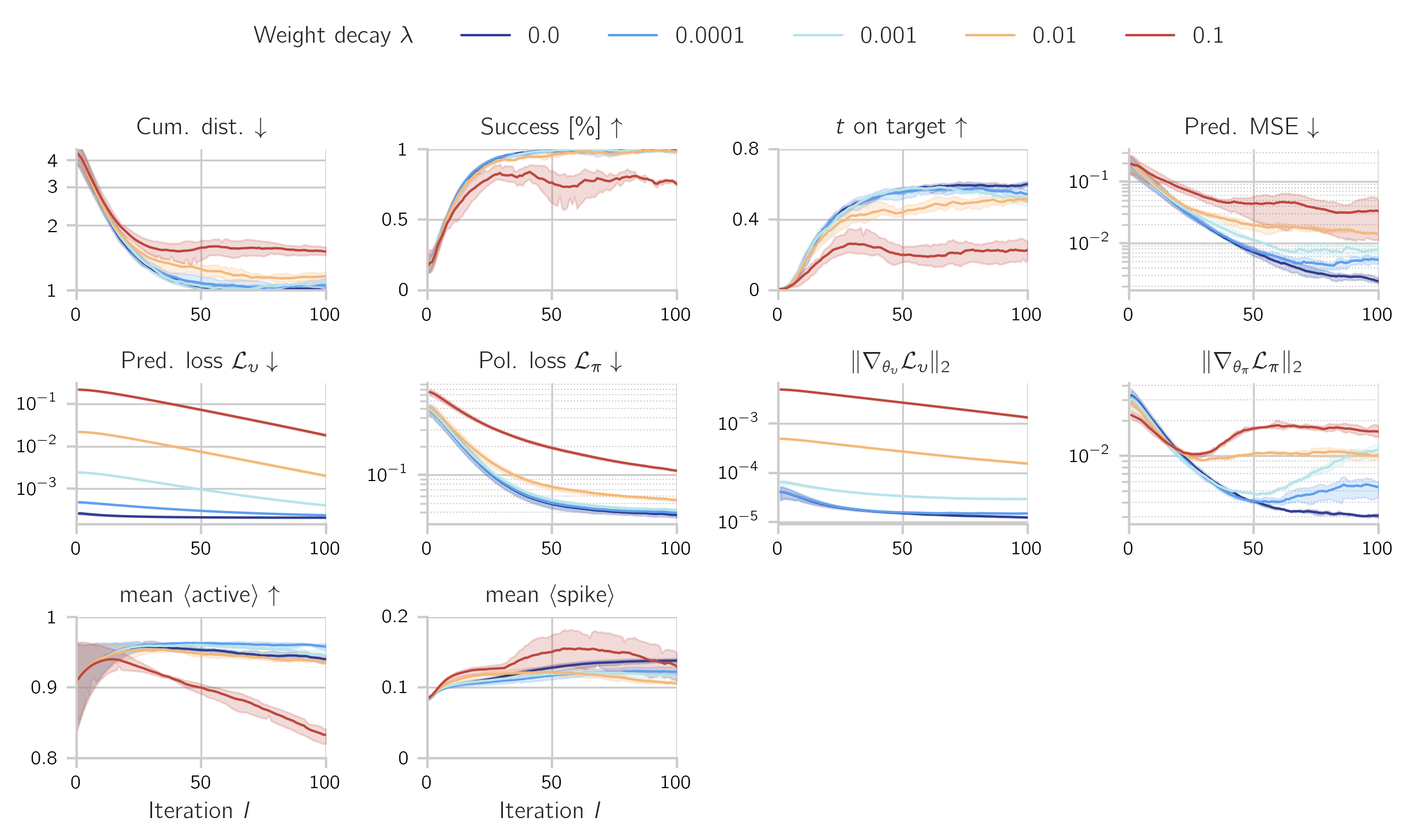}
    \caption{
    \textbf{Effect of L2 weight decay on performance, loss, and gradient magnitude.}
    Stronger weight regularization consistently increases losses and gradients while degrading task metrics. 
    Even mild values ($\lambda_{\text{L2}} = 0.001$) negatively impact time on target and success rate.
    Based on these results, no weight decay is used in subsequent experiments.
    }
    \label{fig:l2_results}
\end{figure}

As shown in \autoref{fig:l2_results}, increasing $\lambda_{\text{L2}}$ consistently degrades both task performance and loss minimization.
While all configurations still learn the task to some degree, higher regularization strengths lead to slower convergence, increased prediction and policy losses, and substantially higher gradient magnitudes—particularly in the prediction network. 
Even moderate regularization values (e.g., $\lambda_{\text{L2}} = 0.001$) reduce time on target and overall success.

Although some reduction in spike activity is observed at higher $\lambda_{\text{L2}}$, this sparsity does not translate into better generalization or performance. 
Instead, the results suggest that in this setting, where data is continuously refreshed and overfitting is not a central concern, weight decay acts more as a hindrance than a help.

We therefore set $\lambda_{\text{L2}} = 0$ in all subsequent experiments.

\FloatBarrier
\clearpage

\section{Activity Regularization}
\label{sec:activity_reg}

Spiking neural networks can exhibit unstable activity patterns during training, such as neurons that never spike, neurons that spike excessively, or entire layers becoming inactive.
These issues can impair learning dynamics and reduce model expressiveness.
To mitigate this, we introduce activity regularization losses that penalize extreme values of either membrane potential or spiking activity.

This method shares conceptual overlap with the adaptive threshold mechanism of ALIF neurons (\autoref{sec:adaptive_lif}), which implicitly discourages high-frequency firing.
In contrast, activity regularization enforces explicit constraints by applying loss terms that penalize deviations from user-defined bounds.
Both mechanisms aim to stabilize dynamics, but only regularization gives direct control over the desired operating regime.

Let $X_{i,l,t}$ denote the activity variable of neuron $i$ in layer $l$ at discrete time step $t$, 
where $X_{i,l,t} \in \{ U_{i,l,t}, S_{i,l,t} \}$ with $U_{i,l,t}$ the membrane voltage and $S_{i,l,t} \in \{0,1\}$ the spike output.
We define the per-neuron average activity over an unroll window of $T$ steps as
\begin{equation}
\bar{X}_{i,l} = \frac{1}{T}\sum_{t=1}^T X_{i,l,t}.
\end{equation}
The lower- and upper-bound activity regularization losses are then given by
\begin{equation}
L_{\text{low}}(X) = \sum_{i,l} \bigl[\min\bigl(0, \bar{X}_{i,l} - X_{\text{low}}\bigr)\bigr]^2,
\end{equation}
\begin{equation}
L_{\text{up}}(X) = \sum_{i,l} \bigl[\max\bigl(0, \bar{X}_{i,l} - X_{\text{up}}\bigr)\bigr]^2,
\end{equation}
where $X_{\text{low}}$ and $X_{\text{up}}$ are user-defined thresholds. 
These terms are weighted by $\lambda_{\text{low}}$ and $\lambda_{\text{up}}$ in the total network loss (see \autoref{eq:loss_pred} and \autoref{eq:loss_pol}) 
and can be applied independently or jointly to shape network dynamics.

\vspace{1em}
\subsubsection*{Lower-bound membrane potential regularization}

In the first experiment, we investigate whether penalizing low membrane potentials can prevent neuron inactivity and improve learning.
This is motivated by the observation that neurons with consistently subthreshold voltages may never spike, reducing their contribution to learning.
We apply the $L_{\text{low}}$ regularizer to the membrane potential $U_{i,l,t}$ and test three thresholds: $U_{\text{low}} \in \{-10, -5, 0\}$.
Each threshold is evaluated under regularization strengths $\lambda_{\text{low}} \in \{0.0, 0.001, 0.01, 0.1\}$.
A value of $\lambda_{\text{low}} = 0.0$ serves as the baseline with no regularization.

\begin{figure}[h!]
    \centering
    \includegraphics[width=0.9\linewidth]{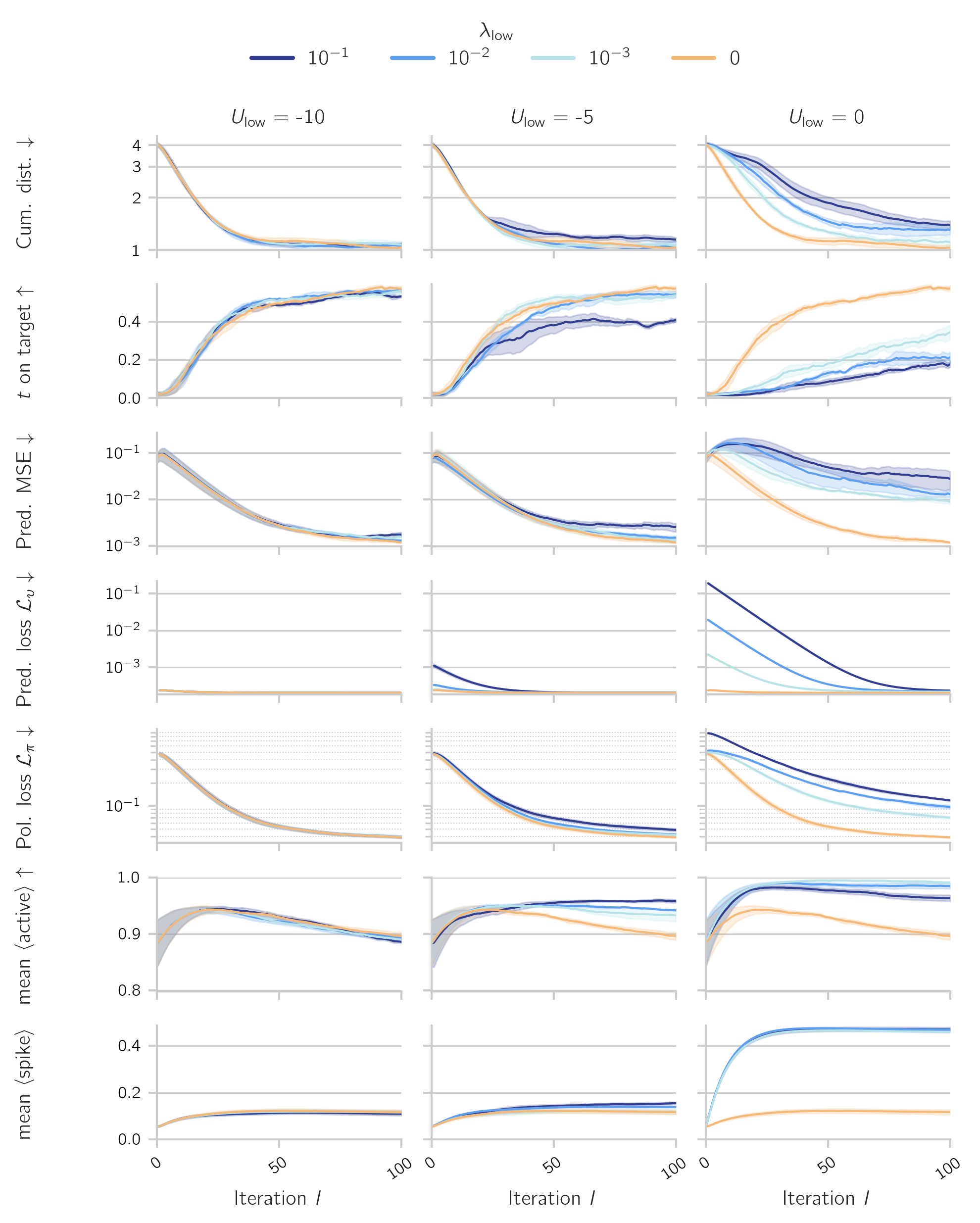}
    \caption{
    \textbf{Effect of lower-bound membrane potential regularization.}
    Each column shows results for a fixed threshold $U_{\text{low}} \in \{-10, -5, 0\}$, and each curve corresponds to a different regularization strength $\lambda_{\text{low}} \in \{0.001, 0.01, 0.1\}$.
    Increasing $\lambda_{\text{low}}$ elevates membrane potentials and increases spike rates (bottom rows), successfully activating silent neurons.
    However, this heightened activity does not translate to improved task performance.
    Both prediction and policy loss increase, and time on target is reduced, particularly for $U_{\text{low}} = 0$.
    These results indicate that naïvely enforcing activity may disrupt useful computation.
    All curves show mean and standard deviation across 3 seeds.
    }
    \label{fig:lower_bound_reg_results}
\end{figure}

\autoref{fig:lower_bound_reg_results} shows the effect of applying $L_{\text{low}}$ for different thresholds $U_{\text{low}}$ and regularization strengths $\lambda_{\text{low}}$.
As the regularization becomes stronger and the threshold increases, we observe a marked rise in spike rate and neuron utilization.
This confirms that the penalty effectively prevents neuron silence and enforces network-wide participation.

However, this comes at a cost: both prediction and policy losses increase, and downstream performance—as measured by time on target and cumulative distance—deteriorates notably, especially for $U_{\text{low}} = 0$.
This suggests that promoting activity without regard to task relevance can degrade the quality of learning signals.
Compared to adaptive thresholding mechanisms like ALIF (see \autoref{sec:adaptive_lif}), which modulate excitability based on recent spiking history, static lower-bound regularization lacks temporal nuance and may induce unnecessary spiking.

Given the unfavorable trade-off, we do not apply $L_{\text{low}}$ in later experiments.

\FloatBarrier
\clearpage

\vspace{1em}
\subsubsection*{Upper-bound spike activity regularization}

In the second experiment, we explore whether constraining spiking activity from above can reduce bursting and promote sparsity.  
We apply the $L_{\text{up}}$ regularizer to the spike output $\overline{S}_{i,l} = \tfrac{1}{T}\sum_{t=1}^T S_{i,l,t}$ of each neuron $i$ in layer $l$.  
The threshold values tested are $\overline{S}_{\text{up}} \in \{0.3, 0.2, 0.1\}$, representing the maximum allowed spike rate averaged over the unroll window.  
We evaluate regularization strengths $\lambda_{\text{up}} \in \{0.0, 0.001, 0.01\}$ to determine the trade-off between enforcing sparsity and maintaining task performance.  

All runs in this section use a fixed prediction unroll window of 40 steps, matching the policy model, and include learnable time constants.  
This provides a consistent basis for evaluating the isolated effect of spike activity regularization.  

\begin{figure}[h!]
    \centering
    \includegraphics[width=\linewidth]{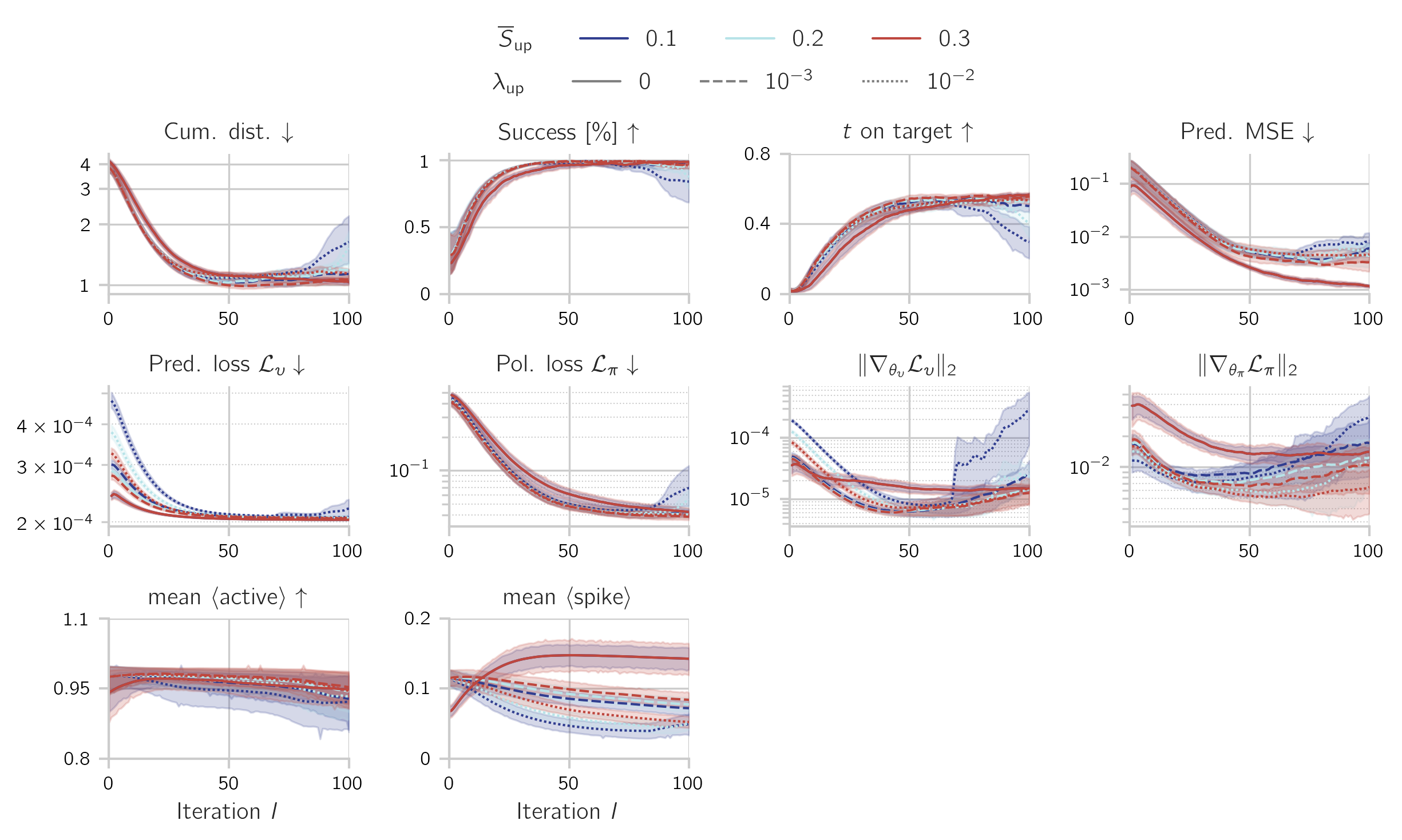}
    \caption{
    \textbf{Effect of upper-bound spiking activity regularization.}  
    Increasing $\lambda_{\text{up}}$ reduces mean spike activity (bottom right) while keeping most neurons active.
    This can accelerate early learning, as seen in higher success rates and time on target, but prediction errors plateau at higher levels and strong regularization impairs final performance and gradient stability.  
    Shaded regions indicate variance across three seeds, which increases for stronger regularization.  
    Overall, spike regularization provides only limited benefits compared to adaptive thresholds (ALIF), which better stabilize dynamics without degrading task accuracy. 
    All curves show mean and standard deviation across 3 seeds.
    }
    \label{fig:activity_reg_results}
\end{figure}

\autoref{fig:activity_reg_results} shows that mild spike regularization successfully suppresses overall spiking activity while keeping nearly all neurons active.  
This often accelerates early learning, as reflected by faster gains in success rate and time on target.  
However, prediction errors (MSE) plateau at higher values compared to unregularized baselines, and strong regularization ($\lambda_{\text{up}}=0.01$) further degrades final success and increases across-seed variance.  
These results suggest that while spike regularization can shape activity levels, it does not consistently improve control performance and can destabilize training when applied too strongly.  
For this reason, we did not adopt $L_{\text{up}}$ in the final model, where adaptive threshold mechanisms (ALIF) provided more robust performance benefits without sacrificing task accuracy.  

\FloatBarrier
\clearpage

\section{Action Regularization Loss}
\label{sec:action_reg_loss}

This experiment is conducted in the 3D reaching task, where we observed frequent overshooting and oscillatory behavior of the robot arm near the target.
While the primary training objective of the policy network is to minimize cumulative distance to the goal, there is no explicit incentive to stop acting once the target is reached.
As a result, the system can exhibit unnecessarily energetic or unstable behavior.

To address this, we introduce auxiliary loss terms that penalize the magnitude and temporal variation of the control signal.
These terms are inspired by classical optimal control, where energy and smoothness costs are frequently used to shape stable trajectories.
A typical control objective $J$ includes:

\begin{equation}
\label{eq:control_cost}
J = \sum_{t=1}^{T} \left[ c(x_t, x^*) + \lambda_u \|u_t\|^2 + \lambda_{u'} \|u_t - u_{t-1}\|^2 \right],
\end{equation}

where $c(x_t, x^*)$ is the task cost, $\lambda_u$ penalizes large control signals, and $\lambda_{u'}$ discourages abrupt changes over time.

We define the following regularization losses:
\begin{align}
\mathcal{L}_{\text{act}} &= \frac{1}{T} \sum_{t=1}^{T} \|u_t\|^2, \\
\mathcal{L}_{\text{smooth}} &= \frac{1}{T-1} \sum_{t=2}^{T} \|u_t - u_{t-1}\|^2.
\end{align}

Both losses are differentiable and applied only to the policy network $\bm{\pi}$, with weighting factors $\lambda_u$ and $\lambda_{u'}$.
We include a warmup phase during which these penalties are disabled to prevent interference with early learning dynamics.

We evaluate all combinations of:
\begin{itemize}
    \item Action magnitude penalty $\lambda_u \in \{0.0, 0.001, 0.01\}$
    \item Action smoothness penalty $\lambda_{u'} \in \{0.0, 0.001, 0.01\}$
\end{itemize}

\begin{figure}[h!]
    \centering
    \includegraphics[width=0.9\linewidth]{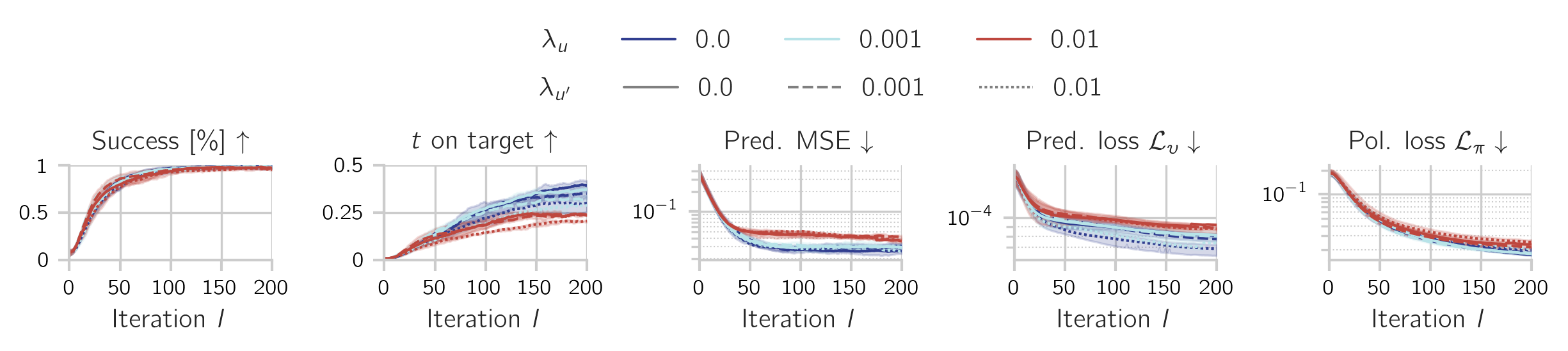}
    \caption{
    \textbf{Effect of action magnitude and smoothness regularization on 3D control task performance.}
    Although higher $\lambda_u$ and $\lambda_{u'}$ produce smoother outputs and slower actions, they do not eliminate oscillations and lead to a drop in task and model performance.
    }
    \label{fig:action_reg_loss_results}
\end{figure}

The results in \autoref{fig:action_reg_loss_results} show that applying either form of regularization slows learning and slightly reduces task performance.
Qualitatively, higher values of $\lambda_u$ and $\lambda_{u'}$ lead to visibly smoother policy outputs and slower arm movements (video not shown).
However, the oscillatory behavior is not fully suppressed, and the success rate and time-on-target metrics remain best when no regularization is applied.

Interestingly, we also observe a consistent degradation in prediction model performance as regularization increases—even though the loss terms affect only the policy.
This suggests an indirect interaction, potentially due to a distribution shift in the training data: slower, smoother actions might lead to less diverse or harder-to-predict state transitions early in training.

Given these findings, we opt not to include action regularization in subsequent experiments.

\FloatBarrier
\clearpage

\section{Action Noise During Training}
\label{sec:action_reg_noise}

Exploration is a key challenge in continuous control, especially in reinforcement learning settings where action distributions must be sampled efficiently. 
Although our control task is supervised, similar concerns can arise: early network biases may lead to overfitting or limited exploration of the state-action space, particularly during the collection of training episodes. 
To address this, we experiment with externally injected Gaussian noise added to the policy output $\bm{u}_t$ at each step during data collection.

The executed control signal $\tilde{\bm{u}}_t$ is sampled from:
\begin{equation}
\label{eq:noise_sample}
\tilde{\bm{u}}_t \sim \mathcal{N}(\bm{u}_t, \sigma_u^2 \mathbb{I}),
\end{equation}
where $\sigma_u$ is the standard deviation of the action noise. 
To balance early exploration with convergence, we optionally decay this noise exponentially over the course of training:
\begin{equation}
\label{eq:noise_decay}
\sigma_{u,t} = \sigma_0 \cdot \gamma_{\sigma}^{\,t},
\end{equation}
where $\sigma_0$ is the initial noise level and $\gamma_{\sigma} \in (0,1]$ is the decay factor. 
At test time, no noise is applied.

\begin{figure}[h!]
    \centering
    \includegraphics[width=0.9\linewidth]{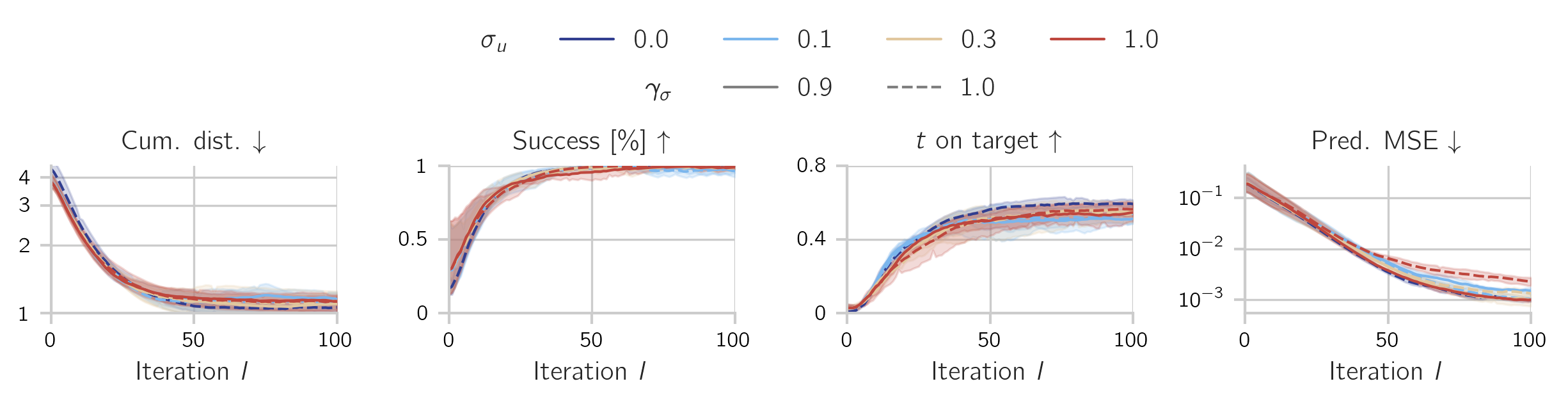}
    \caption{
    \textbf{Impact of Gaussian action noise during training.}  
    All models converge to near-optimal performance regardless of noise level.  
    However, omitting noise leads to faster and more stable convergence.  
    Each curve reflects evaluation on noise-free test episodes.}
    \label{fig:action_reg_noise}
\end{figure}

We compare fixed and decaying noise schedules using $\sigma_0 \in \{0.0, 0.1, 0.3, 1.0\}$ and $\gamma_{\sigma} \in \{1.0, 0.9\}$. 
As shown in \autoref{fig:action_reg_noise}, all configurations eventually converge to good performance on the control task. 
However, models trained without any action noise achieve faster convergence and slightly better final accuracy.
This suggests that in the low-dimensional, well-behaved dynamics of our setting, noise is unnecessary and may even hinder learning by introducing instability during early training.

While noise-driven exploration may prove beneficial in high-dimensional or reinforcement-based tasks, we find no clear advantage in the present case.
We therefore proceed without action noise in the remainder of our experiments.

\FloatBarrier
\clearpage

\section{Reducing Network Parameters}
\label{sec:reducing_parameters}

Spiking neural networks with large fully connected layers can quickly reach millions of trainable parameters, especially when multiple hidden populations are used. To address this, we introduce a simple low-rank factorization scheme that reduces model size without severely impacting task performance.

Specifically, instead of using a full weight matrix $W \in \mathbb{R}^{n \times m}$ to connect two spiking populations of size $n$ and $m$, we decompose $W$ into two smaller matrices: $W = AB$, with $A \in \mathbb{R}^{n \times d}$ and $B \in \mathbb{R}^{d \times m}$, where $d \ll \min(n, m)$. This structure is equivalent to a linear bottleneck of dimensionality $d$ placed between the two populations. We train both $A$ and $B$ end-to-end using backpropagation through time. Only the second linear stage includes a learnable bias term. No nonlinearity is applied between the two transformations.

This factorization is applied uniformly across both the prediction and policy networks, replacing all inter-layer spiking connections (including recurrent ones) with their low-rank equivalents. The latent dimension $d$ is shared across all layers of a network and selected from a fixed set $\{8, 16, 32, 64, \text{full}\}$.

\begin{figure}[h!]
    \centering
    \includegraphics[width=0.9\linewidth]{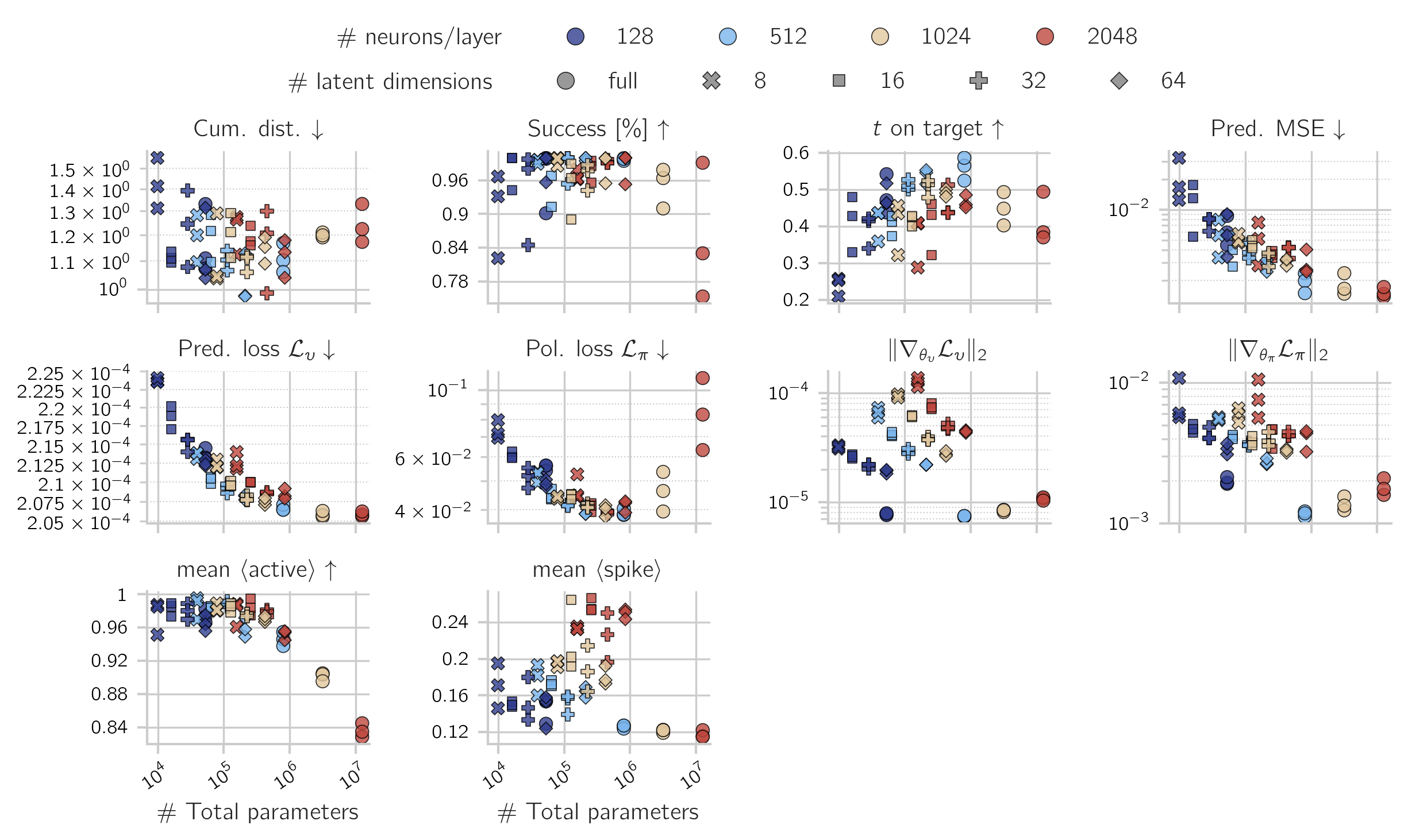}
    \caption{
    \textbf{Effect of low-rank weight factorization on performance and activity.}  
    Each point represents a single model trained on the 2D control task.  
    Marker color indicates the number of neurons per layer (128–2048), and marker shape denotes the number of latent dimensions ($d \in \{8, 16, 32, 64, \text{full}\}$).  
    Multiple metrics are plotted against total parameter count.  
    Low-rank architectures achieve strong performance with significantly fewer parameters than their full-rank counterparts.
    }
    \label{fig:low_rank_appendix}
\end{figure}

We evaluate the impact of this compression strategy on the 2D control task, sweeping over several values of neurons per layer (128, 512, 1024, 2048) and latent dimension $d$. \autoref{fig:low_rank_appendix} summarizes key performance metrics as a function of total parameter count. 

The results show that parameter count can be drastically reduced—by an order of magnitude or more—without a large drop in control performance. In particular, configurations using 512 or 1024 neurons per layer with latent dimension 16–64 consistently achieve good success rates and convergence behavior. Very large models (e.g., with 2048 neurons and full-rank matrices) offer little additional benefit and may lead to overfitting or excessive spiking activity. 

Overall, low-rank factorization provides a practical method for scaling SNN models while controlling memory and compute requirements. This approach also aligns with biological findings suggesting that population activity often resides on low-dimensional manifolds \citep{averbeck2006neural, churchland2012neural}.

\FloatBarrier
\clearpage

\section{SNN Ablation Study and Component Analysis}
\label{sec:snn_ablation}

To isolate which architectural components are responsible for the performance and stability gains of the proposed Pred-control SNN, we conduct a structured ablation study on the \textbf{3D reaching task}.
All variants are trained and evaluated under the same protocol as the full model described in the main text; only the network architecture is modified.

\paragraph{Ablated components and model variants.}
Starting from a minimal \emph{floor model}, we selectively remove or reintroduce three core components:
(i) latent-state compression (\autoref{sec:reducing_parameters}),
(ii) adaptive neuron dynamics via ALIF neurons (\autoref{sec:adaptive_lif}), and
(iii) learnable membrane and synaptic time constants (\autoref{sec:learnable_timescales}).
Intermediate variants disable exactly one component at a time, enabling attribution of performance changes to individual design choices.
Where applicable, we consider both a \emph{small} (S, 512 neurons per layer) and a \emph{large} (L, 2048 neurons per layer) configuration.
All other settings—including data collection, unroll length, optimization, and loss functions—are held fixed.

\paragraph{Quantitative comparison.}
\autoref{tab:snn_ablation_summary} reports final task performance and parameter counts for all ablation variants.
While removing compression yields slightly improved absolute performance, it does so at a dramatic cost in model size:
the uncompressed large model requires over \emph{14$\times$} more parameters than the compressed full model, for only marginal gains.
In contrast, disabling ALIF neurons or learnable time constants leads to consistent degradations in time-on-target, cumulative distance, and prediction accuracy, despite comparable parameter counts.
This indicates that adaptive neuron dynamics contribute more strongly to control quality than width alone.

\begin{figure}[h!]
    \centering
    \includegraphics[width=\linewidth]{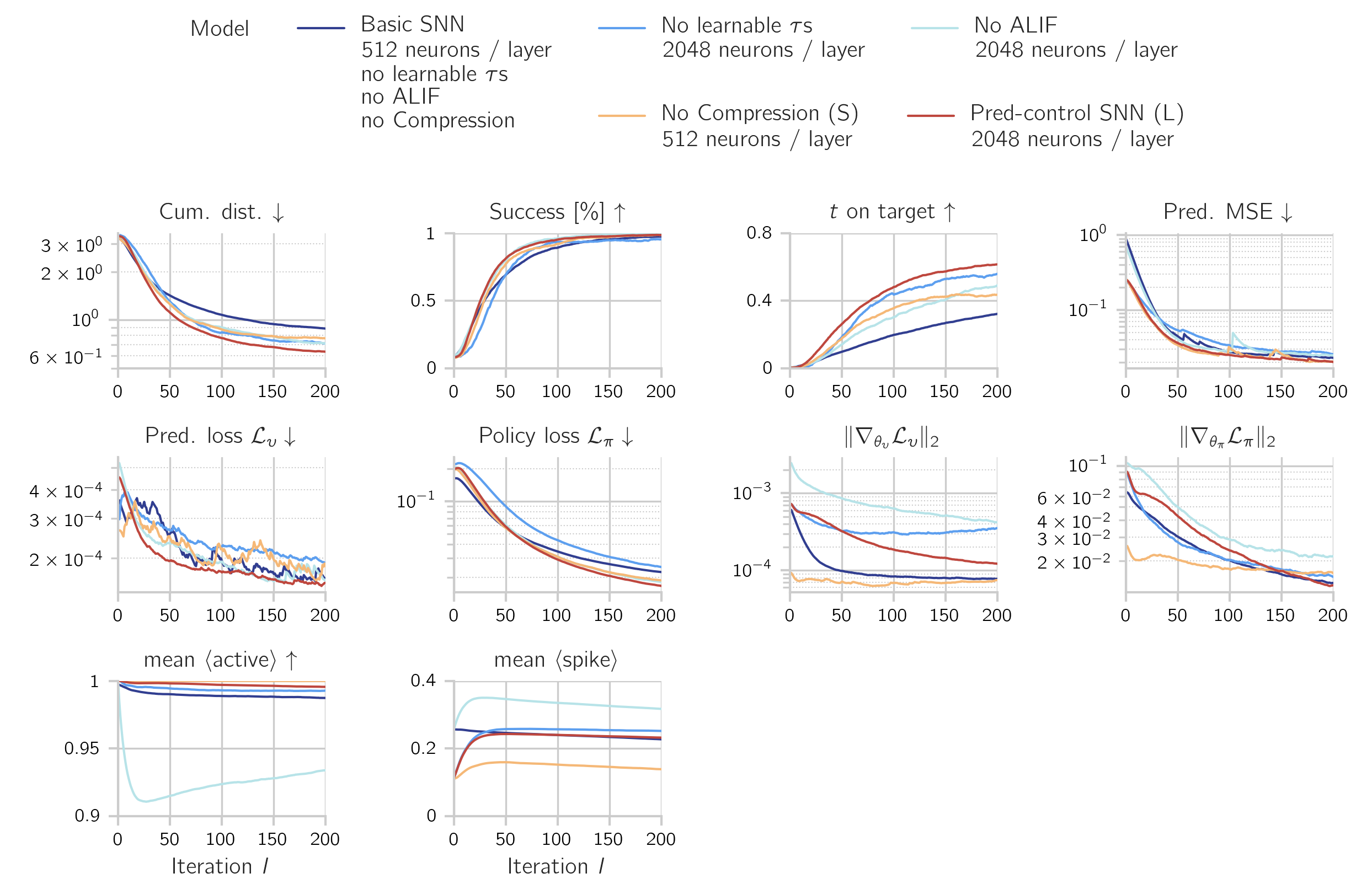}
    \caption{
    \textbf{SNN ablation study on the 3D reaching task.}
    Learning curves for Pred-control SNN variants in which individual architectural components are removed.
    Shown are a minimal \emph{basic} model (no learnable time constants, no ALIF neurons, no parameter compression; 512 neurons per layer),
    the full model (learnable time constants, ALIF neurons, and compression; 2048 neurons per layer),
    and intermediate variants where exactly one component is ablated.
    Top row: task performance metrics (cumulative distance, success rate, time on target, prediction MSE).
    Middle row: optimization losses and gradient norms for predictor and policy.
    Bottom row: population activity statistics.
    Curves show mean over 3 seeds (sd omitted for visual clarity).
    The full model consistently achieves faster convergence, higher time on target, and more stable gradients.
    Removing adaptive neuron dynamics (learnable time constants or ALIF neurons) degrades both learning efficiency and final control quality,
    while compression primarily affects performance through model capacity rather than training stability.
    }
    \label{fig:snn_ablation}
\end{figure}

\paragraph{Learning dynamics.}
Learning curves in \autoref{fig:snn_ablation} further reveal how individual components affect optimization.
The full model converges faster, achieves higher sustained time-on-target, and exhibits smaller, more stable gradient norms.
Removing ALIF neurons or learnable time constants leads to noisier optimization signals and reduced asymptotic control performance, whereas compression primarily influences performance through representational capacity rather than training stability.

\paragraph{Summary.}
Together, these results demonstrate that the gains of the proposed SNN controller do not arise from parameter count alone.
Adaptive neuron dynamics—via ALIF neurons and learnable time constants—play a central role in stabilizing training and enabling effective temporal representations, while compression provides a favorable trade-off between model size and performance.
The full Pred-control SNN therefore occupies a markedly better point on the performance–efficiency frontier than any ablated variant.

\begin{table}[h!]
\centering
\caption{\textbf{Ablation of key SNN components on the 3D reaching task.}
Models are evaluated at the final training iteration (means over 10 seeds; std omitted for clarity).
(L) denotes the large variant with 2048 neurons per layer, and (S) denotes the small variant with 512 neurons per layer.
The Basic SNN disables compression, learnable time constants, and ALIF neurons.}
\label{tab:snn_ablation_summary}
\setlength{\tabcolsep}{5pt}
\renewcommand{\arraystretch}{1.10}
{\fontsize{7.5}{9}\selectfont
\begin{tabular}{@{} l c c c c c c c @{}} \toprule
 & \textbf{Pred-control SNN (L)} & \textbf{Pred-control SNN (S)} & \textbf{No comp.\ (L)} & \textbf{No comp.\ (S)} & \textbf{No ALIF} & \textbf{No learnable $\tau$s} & \textbf{Basic SNN} \\
\midrule
Neurons / layer        & 2048 & 512 & 2048 & 512 & 2048 & 2048 & 512 \\
Policy parameters      & 315{,}422 & 78{,}878 & 4{,}247{,}582 & 275{,}486 & 311{,}326 & 303{,}122 & 272{,}402 \\
Predictor parameters   & 591{,}914 & 148{,}010 & 8{,}456{,}234 & 541{,}226 & 587{,}818 & 579{,}608 & 538{,}136 \\
\textbf{Total parameters} 
                       & \textbf{907{,}336} & \textbf{226{,}888} & \textbf{12{,}703{,}816} & \textbf{816{,}712} & \textbf{899{,}144} & \textbf{882{,}730} & \textbf{810{,}538} \\
\midrule
Cumulative distance $\downarrow$ 
                       & 0.634 & 0.833 & \textbf{0.607} & 0.769 & 0.712 & 0.714 & 0.884 \\
Success rate $\uparrow$
                       & 0.986 & 0.964 & \textbf{0.989} & 0.986 & 0.992 & 0.954 & 0.975 \\
Time on target $\uparrow$
                       & 0.615 & 0.364 & \textbf{0.652} & 0.435 & 0.492 & 0.559 & 0.322 \\
State MSE $\downarrow$
                       & 0.020 & 0.023 & \textbf{0.017} & 0.020 & 0.025 & 0.026 & 0.023 \\
\bottomrule
\end{tabular}
}
\end{table}

\FloatBarrier
\clearpage

\section{Robustness to target perturbations}
\label{sec:robustness}
\begin{figure}[h!]
    \centering
    \includegraphics[width=\linewidth]{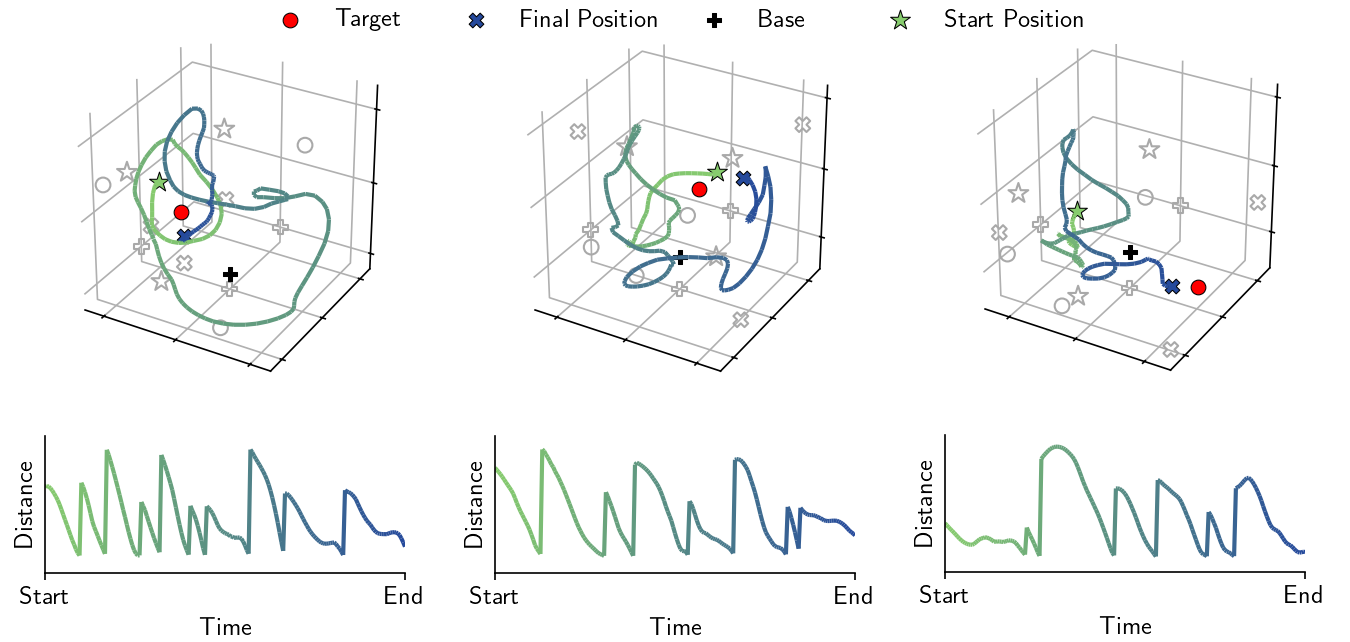}
    \caption{
    \textbf{Robustness to within-episode target shifts.}
    Evaluation of a pretrained Pred-Control SNN under a simple target-perturbation protocol.
    Whenever the end-effector enters the success region, the target position is shifted to a new location (multiple times within a single 200-step episode).
    \textbf{Top:} Example 3D end-effector trajectories under three representative episodes, illustrating repeated re-targeting behavior.
    \textbf{Bottom:} Corresponding distance-to-target traces; each increase reflects a target shift, followed by rapid error reduction as the controller re-acquires the new target.
    No online learning is performed during this test.
    }
    \label{fig:robustness_target_shift}
\end{figure}

To probe robustness beyond the nominal evaluation protocol, we performed a simple closed-loop perturbation test in which the target position is shifted whenever the end-effector enters the success radius.
As shown in \autoref{fig:robustness_target_shift}, the pretrained Pred-Control SNN rapidly reorients toward the updated target and typically re-acquires it within a few control cycles, producing repeated decreases in distance-to-target after each perturbation.
This behavior suggests that the learned policy does not merely memorize a fixed trajectory, but implements a reactive feedback strategy that can adapt online to changing goal states under the learned predictive–control loop.
Importantly, this adaptation occurs without any online weight updates, indicating that robustness arises from the learned internal dynamics and state-dependent control mapping.

\FloatBarrier
\clearpage

\section{Non-spiking RNN Baseline and Model Size Sweep}
\label{sec:rnn_baseline}

To contextualize the proposed spiking controller, we construct a non-spiking recurrent baseline for the \textbf{3D reaching task} that matches the SNN training pipeline as closely as possible.
We reuse the same data collection, episode buffer, unrolling/BPTT loop, optimizer settings, losses, and action scaling; only the network implementation differs (standard floating-point PyTorch layers instead of spiking modules).

\paragraph{Architecture.}
The baseline retains the same two-module structure (recurrent predictor + feedforward policy) and runs at the environment control rate (one recurrent step per control cycle) with the same unroll lengths as the SNN.
As in the spiking setup, the predictor outputs a state increment $\Delta \hat{\bm{s}}_t$ and forms the next-state estimate as
$\hat{\bm{s}}_{t+1} = \bm{s}_t + \Delta \hat{\bm{s}}_t$ under teacher forcing, or $\hat{\bm{s}}_{t+1} = \hat{\bm{s}}_t + \Delta \hat{\bm{s}}_t$ during autoregressive rollouts.
We use an Elman RNN with ReLU nonlinearities for the predictor and the same bounded-action transformation for the policy (including $\tanh$).
No parameter compression (\autoref{sec:reducing_parameters}) is applied.

\paragraph{Training objective and regularization.}
The predictor is trained with the same supervised one-step objective as in the SNN pipeline, i.e., predicting $\bm{s}_{t+1}$ from $(\bm{s}_t,\bm{u}_t)$ under teacher forcing.
The policy is trained via differentiable rollouts through the frozen predictor using the same multi-step objective; we also apply the same auxiliary penalty on the policy pre-$\tanh$ activation to reduce saturation-induced gradient collapse.
In the results below, we distinguish between the one-step prediction objective (\emph{Pred.\ loss}) and the multi-step, autoregressive rollout error (\emph{Pred.\ MSE}), which can diverge even when one-step prediction is accurate.

\begin{figure}[h!]
    \centering
    \includegraphics[width=0.9\linewidth]{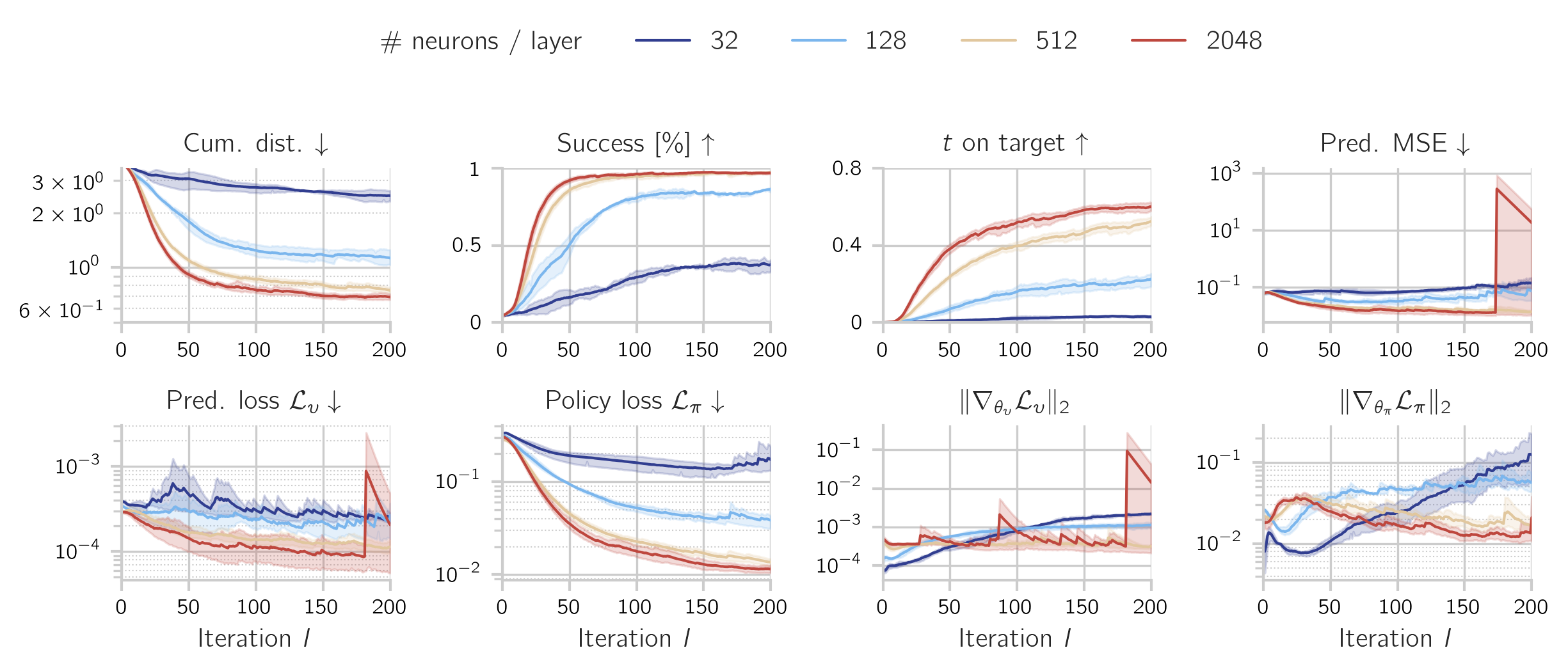}
    \caption{
    \textbf{Non-spiking RNN baseline (3D task): effect of model size.}
    Learning curves for neurons per layer $\in\{32,128,512,2048\}$ applied to all hidden layers in both predictor and policy (3 seeds; mean $\pm$ s.e.m.).
    Control performance improves with model size across runs (higher success and time on target, lower cumulative distance), with the 512--2048 models converging fastest.
    While the one-step prediction objective (Pred.\ loss) remains low across sizes, the largest model shows an outlier in long-horizon autoregressive prediction error (Pred.\ MSE), illustrating that multi-step rollouts can degrade despite strong one-step fits.}
    \label{fig:rnn_baseline_size_sweep}
\end{figure}

\paragraph{Model size sweep.}
We vary the number of neurons per layer $\in \{32,128,512,2048\}$ and set \emph{all} hidden layers in both predictor and policy to this width, keeping all other hyperparameters fixed.
\autoref{tab:rnn_baseline_summary} summarizes the resulting parameter counts and final 3D control performance (means over 3 seeds), while \autoref{fig:rnn_baseline_size_sweep} reports learning curves over training.

Overall, control performance improves strongly with model size: larger baselines converge faster, reach higher success rates, spend more time on target, and reduce cumulative distance more quickly.
The one-step prediction objective remains well-behaved across sizes (low Pred.\ loss), but the largest model exhibits an outlier in long-horizon autoregressive prediction accuracy: while it achieves the best control metrics, its Pred.\ MSE spikes late in training for an outlier run (see \autoref{fig:rnn_baseline_size_sweep}), which is surprising for the large model under constant teacher forcing.
Since the same training protocol (including constant teacher forcing during predictor training) is used for both baseline and SNN, we report this behaviour as-is.
Policy/task performance remains unaffected by this spike.

\begin{table}[h!]
\centering
\caption{\textbf{Non-spiking RNN baseline on the 3D reaching task.}
Models are evaluated at the final training iteration (means over 3 seeds; std omitted for clarity).
All hidden layers in both predictor and policy use the stated width (neurons / layer).
Note that \textbf{Pred.\ MSE} is a multi-step, autoregressive rollout error and can show outlier behaviour even when one-step prediction remains accurate (cf.\ \autoref{fig:rnn_baseline_size_sweep}).}
\label{tab:rnn_baseline_summary}
\setlength{\tabcolsep}{6pt}
\renewcommand{\arraystretch}{1.10}
{\fontsize{7.5}{9}\selectfont
\begin{tabular}{@{} l c c c c @{}} \toprule
 \textbf{Non-spiking baseline model} & & & & \\
\midrule
Neurons / layer         & 32        & 128       & 512        & 2048 \\
Policy parameters       & 1{,}868    & 19{,}724   & 275{,}468   & 4{,}247{,}564 \\
Predictor parameters    & 3{,}219    & 37{,}395   & 542{,}739   & 8{,}462{,}355 \\
\textbf{Total parameters}
                       & \textbf{5{,}087} & \textbf{57{,}119} & \textbf{818{,}207} & \textbf{12{,}709{,}919} \\
\midrule
Cumulative distance $\downarrow$ 
                       & 2.479     & 1.129     & 0.794     & \textbf{0.691} \\
Success rate $\uparrow$
                       & 0.377     & 0.864     & 0.969     & \textbf{0.970} \\
Time on target $\uparrow$
                       & 0.031     & 0.226     & 0.492     & \textbf{0.601} \\
Pred.\ MSE $\downarrow$
                       & 0.144     & 0.074     & \textbf{0.013} & 18.587 \\
\bottomrule
\end{tabular}
}
\end{table}

\paragraph{Note on policy recurrence.}
We also tested a recurrent policy variant under the same setup, but it did not improve performance and often reduced training stability; we therefore proceed with the feedforward policy baseline.

\FloatBarrier
\clearpage
    
\section{Hyperparameters}
\label{sec:hyperparameters}

Below is a summary of key hyperparameters and their values used in our final Pred-Control SNN model. 

\begin{table}[h]
\centering
\footnotesize
\caption{\textbf{Overview of key hyperparameters in the final Pred-Control SNN.} Values reflect the best-performing configuration after ablations.}
\label{tab:hyperparams}
\begin{tabular}{lll}
\toprule
\textbf{Hyperparameter} & \textbf{Symbol} & \textbf{Final Value} \\
\midrule
\multicolumn{3}{l}{\textit{Simulation and Training Setup}} \\
\midrule
Simulation step & $\Delta t$ & 0.02\,s (50\,Hz), 7 SNN sub-steps \\
Episode length & $T$ & 200 steps = 4\,s\\
Parallel environments & -- & 64 \\
Random seeds & -- & 3 \\
Success threshold & -- & 0.05 (2D), 0.123\,m (3D) \\
Replay buffer size & $M$ & 6400 (full memory) \\
Mini-batches / iteration & $n_{\upsilon}, n_{\pi}$ & 25 \\
Batch size & $N_{\upsilon}, N_{\pi}$ & 256 \\
Warmup steps & $T_{\text{warm}}$ & 10 \\
Unroll steps (prediction/policy) & $T_{\text{unroll}}^{\upsilon},T_{\text{unroll}}^{\pi}$ & 40 / 40 \\
Teacher forcing prob. & $p_{tf}$ & 1.0 \\
\midrule
\multicolumn{3}{l}{\textit{Optimization}} \\
\midrule
Learning rate (prediction / policy) & $\alpha_{\upsilon}, \alpha_{\pi}$ & $10^{-3}$ \\
Learning rate (time constants) & $\alpha_{\tau}$ & $0.01$ \\
Optimizer & -- & Adam \\
Learning rate schedule & -- & Constant \\
\midrule
\multicolumn{3}{l}{\textit{Neuron Model}} \\
\midrule
Membrane time constant & $\tau_{\text{mem}}$ & 10\,ms (learned) \\
Synaptic time constant & $\tau_{\text{syn}}$ & 5\,ms (learned) \\
Adaptation time constant & $\tau_{\text{ada}}$ & 100\,ms (learned) \\
Initialization rate & $\nu$ & 125\,Hz \\
Threshold baseline & $\vartheta_0$ & 1.0 \\
Resting potential & $U_{\text{rest}}$ & 0 \\
ALIF decay & $\Delta \vartheta$ & 10 \\
ALIF adaptation scale & $\xi_{\vartheta}$ & 0.1 \\
\midrule
\multicolumn{3}{l}{\textit{Surrogate Gradients}} \\
\midrule
Surrogate function & -- & Gaussian Spike \\
Steepness & $\beta$ & 16 \\
Scaling factor & $\gamma$ & 1.0 \\
\midrule
\multicolumn{3}{l}{\textit{Network Architecture}} \\
\midrule
Hidden layers & -- & 2 spiking + 1 readout \\
Recurrence & $\rho$ & Prediction model only, $\rho=0.9$ \\
Neurons per layer & -- & 512 (baseline), 2048 w/ compression \\
Latent dimension & $d$ & 64 (bottleneck) \\
\midrule
\multicolumn{3}{l}{\textit{Regularization (Final Model)}} \\
\midrule
Weight decay & $\lambda_{L2}$ & Not used \\
Activity regularization & $\lambda_{\text{low}}, \lambda_{\text{up}}$ & Not used \\
Action penalties & $\lambda_u, \lambda_{u'}$ & Not used \\
Action noise & $\sigma_u$ & Not used \\
\bottomrule
\end{tabular}
\end{table}